\documentclass{article}

     \PassOptionsToPackage{numbers, compress, sort}{natbib}

     \usepackage[final]{neurips_2021}

\usepackage{xcolor}
\definecolor{mydarkblue}{rgb}{0,0.08,0.45}

\usepackage[utf8]{inputenc} %
\usepackage[T1]{fontenc}    %
\usepackage[
    bookmarks=true,
    hyperindex=true,
    breaklinks=true,
    colorlinks=true,
    linkcolor=mydarkblue,
    citecolor=mydarkblue,
    filecolor=mydarkblue,
    urlcolor=mydarkblue]{hyperref}       %
\usepackage{bookmark}
\usepackage{url}            %
\usepackage{booktabs}       %
\usepackage{amsfonts}       %
\usepackage{nicefrac}       %
\usepackage{microtype}      %
\usepackage{amsmath}
\usepackage{amsthm}
\usepackage{mathtools}
\usepackage{verbatim}
\usepackage[capitalize,nameinlink]{cleveref}
\usepackage{booktabs} %
\usepackage[inline]{enumitem}
\usepackage{wrapfig}
\usepackage{multirow}
\usepackage{tablefootnote}
\usepackage{caption}
\usepackage{algorithm,algpseudocode}
\usepackage{changepage}
\usepackage{tabularx}
\usepackage{subcaption}
\usepackage{siunitx}
\robustify\bfseries  %

\usepackage{tcolorbox}
\newtcolorbox[auto counter,crefname={Box}{Boxes}]{pabox}[2][]{%
title=Box~\thetcbcounter $\quad$ #2, label={#1}}

\newtheorem{thm}{Theorem}[section]

\newtheorem{lem}[thm]{Lemma}

\theoremstyle{definition}
\newtheorem{defn}{Definition}[section]
\newtheorem{des}{Desiderata}[section]

\newcounter{daggerfootnote}
\newcommand*{\daggerfootnote}[1]{%
    \setcounter{daggerfootnote}{\value{footnote}}%
    \renewcommand*{\thefootnote}{\fnsymbol{footnote}}%
    \footnotetext[2]{#1}%
    \setcounter{footnote}{\value{daggerfootnote}}%
    \renewcommand*{\thefootnote}{\arabic{footnote}}%
    }

\newcommand{\highlight}[1]{\textcolor{purple}{#1}}
\newcommand{\fanin}{\mathrm{fan\_in}}
\newcommand{\fanout}{\mathrm{fan\_out}}
\global\long\def\onev{\mathbf{1}}%

\DeclareMathOperator*{\argmin}{arg\;min}

\DeclareMathOperator*{\EV}{\mathbb{E}}

\global\long\def\Gaus{\mathcal{N}}%
\global\long\def\defeq{\mathbin{\overset{\mathrm{def}}{=}}}%
\global\long\def\R{\mathbb{R}}%

\global\long\def\trsp{\top}%

\makeatletter
\let\orgdescriptionlabel\descriptionlabel
\newcommand*{\@restrictlabeltext}[1]{#1\protected@edef\@currentlabel{#1}}
\newcommand*{\nolabel}[1]{#1}%
\renewcommand*{\descriptionlabel}[1]{%
  \let\orglabel\label
  \let\label\@gobble
  \let\orig@hfil\hfil
  \def\hfil{}%
  \let\nolabel\@gobble
  \let\restrictlabeltext\@firstofone
  \phantomsection
  \protected@edef\@currentlabel{#1}%
  \let\hfil\orig@hfil
  \let\label\orglabel
  \let\restrictlabeltext\@restrictlabeltext
  \orgdescriptionlabel{#1}%
}
\makeatother

\crefformat{footnote}{#2\footnotemark[#1]#3}

\title{
Tensor Programs V:\\ Tuning Large Neural Networks via\\ Zero-Shot Hyperparameter Transfer
}

\author{%
Greg Yang\thanks{Equal contribution. Order is random.
  Correspondence to \texttt{\{gregyang, edwardhu\}@microsoft.com}} $\ ^\times$
  \quad Edward J. Hu\footnotemark[1]\  $\ ^\times$\footnotemark[2]
  \quad Igor Babuschkin$^\circ$
  \quad Szymon Sidor$^\circ$
  \quad Xiaodong Liu$^\times$ \\
  \quad \textbf{David Farhi}$^\circ$
  \quad \textbf{Nick Ryder}$^\circ$
  \quad \textbf{Jakub Pachocki}$^\circ$
  \quad \textbf{Weizhu Chen}$^\times$
  \quad \textbf{Jianfeng Gao}$^\times$\\
  $^\times$Microsoft Corporation  \qquad $^\circ$OpenAI
  }

\begin{document}
\daggerfootnote{Work done partly during \href{https://www.microsoft.com/en-us/research/academic-program/microsoft-ai-residency-program/}{Microsoft AI Residency Program}.}

\maketitle

\begin{abstract}
Hyperparameter (HP) tuning in deep learning is an expensive process, prohibitively so for neural networks (NNs) with billions of parameters.
We show that, in the recently discovered Maximal Update Parametrization ($\mu$P), many optimal HPs remain stable even as model size changes.
This leads to a new HP tuning paradigm we call \emph{$\mu$Transfer}: parametrize the target model in $\mu$P, tune the HP indirectly on a smaller model, and \textit{zero-shot transfer} them to the full-sized model, i.e., without directly tuning the latter at all.
We verify $\mu$Transfer on Transformer and ResNet.
For example, 1) by transferring pretraining HPs from a model of 13M parameters, we outperform published numbers of BERT-large (350M parameters), with a total tuning cost equivalent to pretraining BERT-large once; 2) by transferring from 40M parameters, we outperform published numbers of the 6.7B GPT-3 model, with tuning cost only 7\% of total pretraining cost.
A Pytorch implementation of our technique can be found at \url{github.com/microsoft/mup} and installable via \texttt{pip install mup}.%
\end{abstract}

\begin{wrapfigure}{r}{0.5\textwidth}
  \vspace{-4em}
    \centering
    \includegraphics[width=0.5\textwidth]{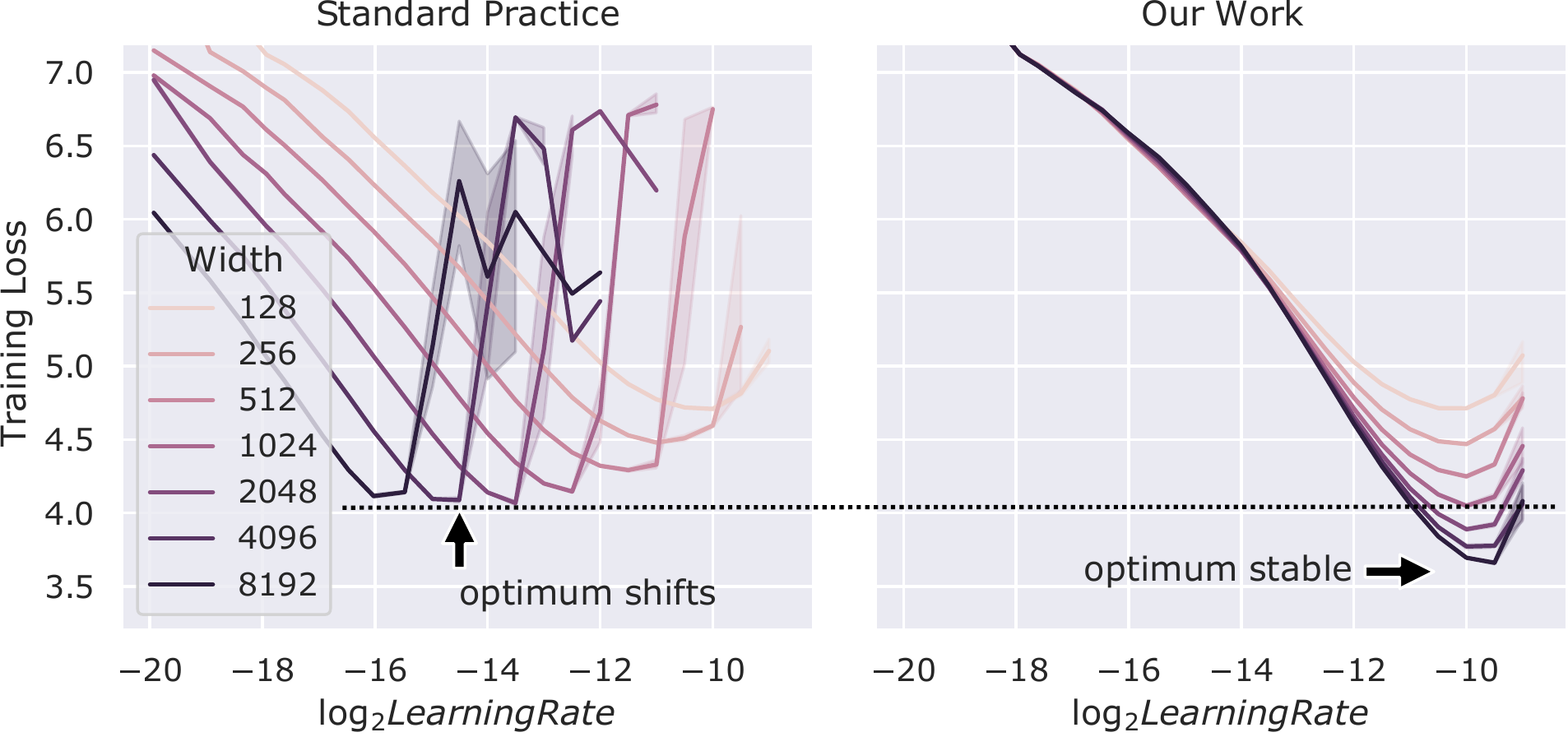}
    \caption[Training loss against learning rate on Transformers of varying $d_{model}$ trained with Adam]{
    Training loss against learning rate on Transformers of varying $d_{model}$ trained with Adam. Conventionally and in contrast with our technique, different widths do not share the same optimal hyperparameter; wider networks do not always perform better than narrower ones; in fact they underperform the same-width networks in our technique even after tuning learning rate (see dashed line).
    See \cref{sec:sp_fails,sec:mutransfer} for experimental setup.}
    \label{fig: figure1}
    \vspace{-1em}
\end{wrapfigure}

\section{Introduction}
\label{sec:intro}

Hyperparameter (HP) tuning is critical to deep learning.
Poorly chosen HPs result in subpar performance and training instability.
Many published baselines are hard to compare to one another due to varying degrees of HP tuning.
These issues are exacerbated when training extremely large deep learning models, since state-of-the-art networks with billions of parameters become prohibitively expensive to tune.

\begin{wrapfigure}{r}{0.4\textwidth}
    \centering
    \includegraphics[width=0.38\textwidth]{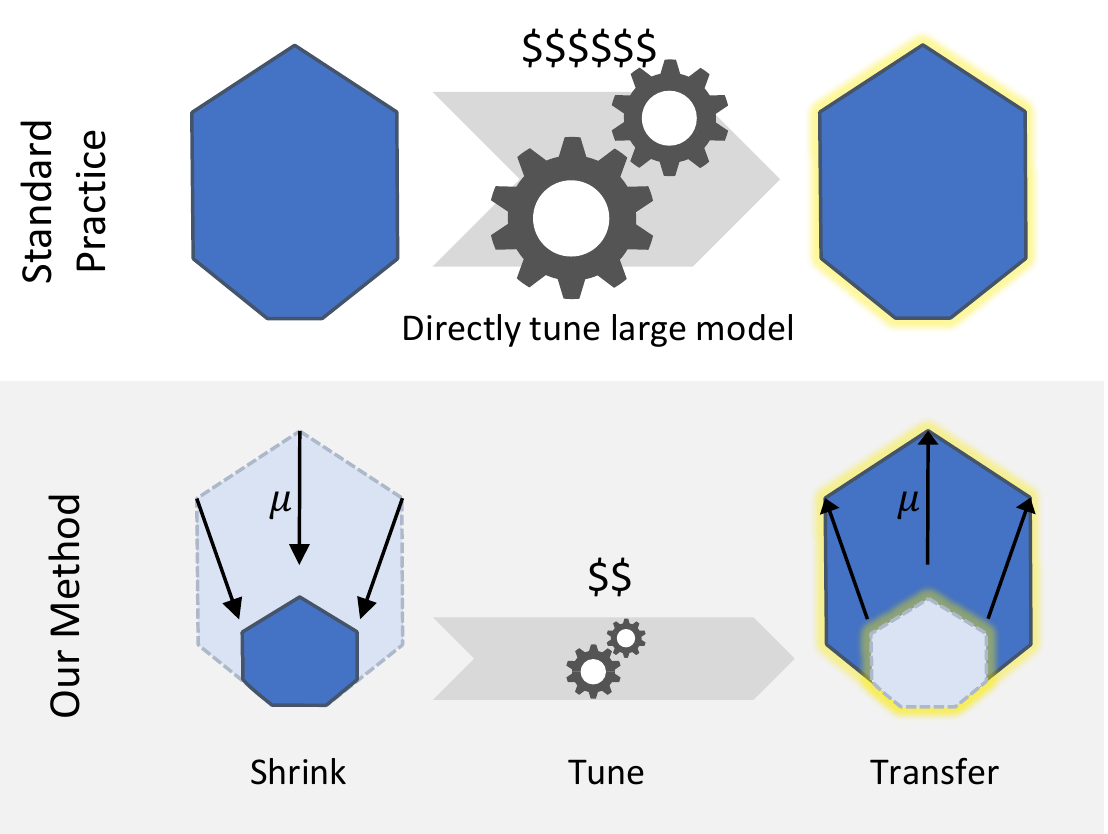}
    \caption{Illustration of $\mu$Transfer}
    \label{fig:mainfig}
\end{wrapfigure}
Recently, \citep{TP4} showed that different neural network parametrizations induce different infinite-width limits and proposed the \emph{Maximal Update Parametrization (abbreviated $\mu$P)} (summarized in \cref{tab:MUP}) that enables ``maximal'' feature learning in the limit.
Intuitively, it ensures that each layer is updated on the same order during training \emph{regardless of width}.%
\footnote{i.e., the updates' effect on activations becomes roughly independent of width in the large width limit.}
In contrast, while the standard parametrization (SP) ensures activations are of unit order at initialization, it actually causes them to blow up in wide models during training \citep{TP4} essentially due to an imbalance of per-layer learning rate (also see \cref{fig:blowupafter1step}).
We leverage $\mu$P to \emph{zero-shot transfer HPs from small models to large models} in this work -- that is, we obtain near optimal HPs on a large model without directly tuning it at all!
While practitioners have always guessed HPs of large models from those of small models, the results are hit-or-miss at best because of incorrect parametrization.
For example, as shown in \cref{fig: figure1}, in a Transformer, the optimal learning rate is stable with width in $\mu$P (right) but far from so in standard parametrization
(left).
In addition to width, we empirically verify that, with a few caveats, HPs can also be transferred across depth (in \cref{sec:scienceexperiments}) as well as batch size, language model sequence length, and training time (in \cref{app:bsz_seqlen_time}).
This reduces the tuning problem of an (arbitrarily) large model to that of a (fixed-sized) small model.
Our overall procedure, which we call \emph{$\mu$Transfer}, is summarized in \cref{alg:hptransfer} and \cref{fig:mainfig}, and the HPs we cover are summarized in \cref{tab:whichtransfer,tab:transferables}.

There are several benefits to our approach:
\begin{enumerate*}
    \item \textbf{Better Performance:} $\mu$Transfer is not just about predicting how the optimal learning rate scales in SP.
    In general, we expect the $\mu$Transferred model to outperform its SP counterpart with learning rate optimally tuned.
    For example, this is the case in \cref{fig: figure1} with the width-8192 Transformer.
    We discuss the reason for this in \cref{sec:SPdefect,sec:primermulti}.
    \item \textbf{Speedup:} It provides massive speedup to the tuning of large models.
    For example, we are able to outperform published numbers of (350M) BERT-large \citep{devlin_bert_2019}
    purely by zero-shot HP transfer, with tuning cost approximately equal to 1 BERT-large pretraining.
    Likewise, we outperform the published numbers of the 6.7B GPT-3 model \citep{brown2020language} with tuning cost being only 7\% of total pretraining cost.
    For models on this scale, HP tuning is not feasible at all without our approach.%
    \item \textbf{Tune Once for Whole Family:} For any fixed family of models with varying width and depth (such as the BERT family or the GPT-3 family), we only need to tune a single small model and can reuse its HPs for all models in the family.%
    \footnote{but possibly \emph{not} for different data and/or tasks.}
    For example, we will use this technique to tune BERT-base (110M parameters) and BERT-large (350M parameters) simultaneously by transferring from a 13M model.
    \item \textbf{Better Compute Utilization:} While large model training needs to be distributed across many GPUs, the small model tuning can happen on individual GPUs, greatly increasing the level of parallelism for tuning (and in the context of organizational compute clusters, better scheduling and utilization ratio).
    \item \textbf{Painless Transition from Exploration to Scaling Up:}
    Often, researchers explore new ideas on small models but, when scaling up, find their HPs optimized during exploration work poorly on large models.
    $\mu$Transfer would solve this problem.
\end{enumerate*}

\enlargethispage*{\baselineskip}

In addition to the HP stability property, we find that \emph{wider is better throughout training} in $\mu$P, in contrast to SP (\cref{sec:widerbetter}).
This increases the reliability of model scaling in deep learning.

In this work, we primarily focus on hyperparameter transfer with respect to training loss.
In settings where regularization is not the bottleneck to test performance, as in all of our experiments here, this also translates to efficacy in terms of test loss.
In other settings, such as finetuning of models on small datasets, $\mu$Transfer may not be sufficient, as we discuss in \cref{sec:scienceexperiments}.

\begin{algorithm}[t]
  \caption{Tuning a Large Target Model via $\mu$Transfer}
  \begin{algorithmic}[1]
  \State Parametrize target model in Maximal Update Parametrization ($\mu$P)
  \State Tune a smaller version (in width and/or depth) of target model
  \State Copy tuned hyperparameters to target model
  \end{algorithmic}
  \label{alg:hptransfer}
  \end{algorithm}

\begin{table}[t]
    \centering
    \caption[Hyperparameters That Can Be $\mu$Transferred, Not $\mu$Transferred, or $\mu$Transferred Across]{\textbf{Hyperparameters That Can Be $\mu$Transferred, Not $\mu$Transferred, or $\mu$Transferred Across,}
    with a few caveats discussed in \cref{sec:scienceexperiments}.
    * means \emph{empirically validated only} on Transformers, while all others additionally have theoretical justification.}
    \begin{tabular}{ccc}
    \toprule
    $\mu$Transferable                & Not $\mu$Transferable              & $\mu$Transferred \emph{Across}     \\
    \midrule
    optimization related, init,           & regularization                & width, depth*, batch size*,   \\
    parameter\ multipliers, etc    & (dropout, weight decay, etc)  & training time*, seq length*           \\
    \bottomrule                                       
    \end{tabular}
    \label{tab:whichtransfer}
\end{table}

\begin{table}[t]
    \centering
    \caption[Examples of $\mu$Transferable Hyperparameters]{\textbf{Examples of $\mu$Transferable Hyperparameters.}
    All of the below can also be specialized to per-layer hyperparameters.}
    \begin{tabular}{ccc}
    \toprule
    Optimizer Related                   & Initialization    & {Parameter Multipliers}            \\
    \midrule
    learning rate (LR), momentum,       & per-layer         & multiplicative constants after       \\
    Adam beta, LR schedule, etc              & init.\ variance    & weight/biases, etc  \\
    \bottomrule                                       
    \end{tabular}
    \label{tab:transferables}
\end{table}

\newpage

\paragraph{Our Contributions}
\begin{itemize}
    \item We demonstrate it is possible to zero-shot transfer near optimal HPs to a large model from a small version via the Maximal Update Parametrization ($\mu$P) from \citep{TP4}.
    \item While \citep{TP4} only covered SGD, here we derive $\mu$P for Adam as well (\cref{tab:MUP}).
    \item We propose a new HP tuning technique, \emph{$\mu$Transfer}, for large neural networks based on this observation that provides massive speedup over conventional methods and covers both SGD and Adam training;
    \item We thoroughly verify our method on machine translation and large language model pretraining (in~\cref{sec:BERT}) as well as image classification (in~\cref{sec:resnetexperiments});
    \item We release a PyTorch~\citep{pytorch} package for implementing $\mu$Transfer painlessly.
    A sketch of this package is given in \cref{sec:packageimpl}.
\end{itemize}

\paragraph{Terminologies}
Sometimes, to be less ambiguous, we often refer to the ``large model'' as the \emph{target model}, as it is the model we wish to ultimately tune, while we refer to the ``small model'' as the \emph{proxy model}, as it proxies the HP tuning process.
We follow standard notation $d_{model}, d_{head} = d_k, d_v, n_{head}, d_{ffn}$ regarding dimensions in a Transformer; one can see \cref{fig:transformerschematics} for a refresher.

\paragraph{\emph{Tensor Programs} Series}
This paper is the 5th installment of the \emph{Tensor Programs} series.
While it is self-contained with the target audience being practitioners and empirical researchers, this paper presents the first major \emph{practical} payoff of the \emph{theoretical} foundation built in previous works \citep{scaling,TP1,TP2,TP2b,TP3,TP4}.

\section{Parametrization Matters: A Primer}
\label{sec:primer}

In this section, we give a very basic primer on why the correct parametrization can allow HP transfer across width, but see \cref
{sec:GausVsTensorProducts,sec:DeriveMUP,sec:otherparamdontwork} for more (mathematical) details.

The Central Limit Theorem (CLT) says that, if $x_{1},\ldots,x_{n}$
are iid samples from a zero-mean, unit-variance distribution, then
$\frac{1}{\sqrt{n}}(x_{1}+\cdots+x_{n})$ converges to a standard
Gaussian $\Gaus(0,1)$ as $n\to\infty$. Therefore, we can say that
$\frac{1}{\sqrt{n}}$ is the right order of \emph{scaling factor}
$c_{n}$ such that $c_{n}(x_{1}+\cdots+x_{n})$ converges to something
nontrivial. In contrast, if we set $c_{n}=1/n$, then $c_{n}(x_{1}+\cdots+x_{n})\to0$;
or if $c_{n}=1$, then $c_{n}(x_{1}+\cdots+x_{n})$ blows up in variance
as $n\to\infty$.

Now suppose we would like to minimize the function 
\begin{equation}
F_{n}(c)\defeq\EV_{x_{1},\ldots,x_{n}}f(c(x_{1}+\cdots+x_{n}))
\label{eqn:FnExample}
\end{equation}
over $c\in\R$, for some bounded continuous function $f:\R\to\R$.
If we reparametrize $c=\alpha/\sqrt{n}$ for $\alpha \in \R$, then
by CLT, $G_{n}(\alpha)\defeq F_{n}(c)\to\EV f(\Gaus(0,\alpha^{2}))$
stabilizes into a function of $\alpha$ as $n\to\infty$. Then for
sufficiently large $n$, the optimal $\alpha_{n}^{*}\defeq\argmin_{\alpha}G_{n}(\alpha)$
should be close to $\alpha_{N}^{*}$ for any $N>n$, and indeed, for
$N=\infty$ --- this precisely means we can \emph{transfer} the optimal
$c_{n}^{*}$ or $\alpha_{n}^{*}$ for a smaller problem (say $F_{n}$)
to a larger problem (say $F_{N}$): $G_{N}$ is approximately minimized
by $\alpha_{n}^{*}$ and $F_{N}$ is approximately minimized by $c_{n}^{*}\sqrt{n/N}$.
Because the transfer algorithm is simply copying $\alpha$, we say
the parametrization $c=\alpha/\sqrt{n}$ is the \emph{correct parametrization}
for this problem.

In the scenario studied in this paper, $x_{1},\ldots,x_{n}$ are akin
to randomly initialized parameters of a width-$n$ neural network,
$c$ is akin to a HP such as learning rate, and $f$ is
the test-set performance of the network \emph{after training}, so
that $F_{n}$ gives its expectation over random initializations. Just
as in this example, if we parametrize the learning rate and other
HPs correctly, then we can directly copy the optimal HPs
for a narrower network into a wide network and expect approximately
optimal performance --- this is the \emph{(zero-shot) hyperparameter transfer}
we propose here. It turns out the Maximal Update Parametrization ($\mu$P)
introduced in \cite{TP4} is correct (akin to the parametrization
in $\alpha$ above), while the standard parametrization (SP) is incorrect
(akin to the parametrization in $c$).
We will review both parametrizations shortly.
Theoretically, a $\mu$P network has a well-defined infinite-width limit  --- akin to $(x_1 + \cdots + x_n)/\sqrt n$ having a $\Gaus(0, 1)$ limit by CLT --- while a SP network does not (the limit will blow up) \citep{TP4}.%
\footnote{The more theoretically astute reader may observe that SP with a $\Theta(1/width)$ learning rate induces a well-defined infinite-width limit exists as well.
Nevertheless, this does not allow HP transfer because this limit is in kernel regime as shown in \citep{TP4}.
See \cref{sec:otherparamdontwork} for more discussions.}
In fact, based on the theoretical foundation laid in \citep{TP4},  we argue in \cref{{sec:otherparamdontwork}} that $\mu$P should also be the \emph{unique} parametrization that allows HP transfer across width.
For a more formal discussion of the terminologies \emph{parametrization} and \emph{transfer}, see \cref{sec:paramterminology}

We emphasize that, to ensure transferability of any hyperparameter (such as learning rate), it's not sufficient to reparametrize \emph{only} that hyperparameter, but rather, we need to identify and correctly reparametrize \emph{all} hyperparameters in \cref{tab:transferables}.
For example, in \cref{fig: figure1}, the wide models in SP still underperform their counterparts in $\mu$P, even with learning rate tuned optimally.
This is precisely because SP does not scale parameter multipliers and input/output layer learning rates correctly in contrast to $\mu$P (see \cref{tab:MUP}).
See \cref{sec:primermulti} for more intuition via a continuation of our example here.
We shall also explain this more concretely in the context of neural networks in \cref{sec:SPdefect}.

\section{Hyperparameters Don't Transfer Conventionally}
\label{sec:sp_fails}

In the community there seem to be conflicting assumptions about HP stability.
\emph{A priori}, models of different sizes don't have any reason to share the optimal HPs.
Indeed, papers aiming for state-of-the-art results often tune them separately.
On the other hand, a nontrivial fraction of papers in deep learning fixes all HPs when comparing against baselines, which reflects an assumption that the optimal HPs should be stable --- not only among the same model of different sizes but also among models of different designs --- therefore, such comparisons are fair.
Here, we demonstrate HP \emph{instability}  across width explicitly in MLP and Transformers in the standard parametrization.
We will only look at training loss to exclude the effect of regularization.

\paragraph{MLP with Standard Parametrization}
We start with a 2-hidden-layer MLP with activation function $\phi$, using the standard parametrization%
\footnote{i.e.\ the default parametrization offered by common deep learning frameworks. See \cref{tab:MUP} for a review.}
with LeCun initialization%
\footnote{The key here is that the init.\ variance $\propto 1/\fanin$, so the same insights here apply with e.g.\ He initialization.}
 akin to the default in PyTorch:
\begin{equation}
    \begin{gathered}
        f(\xi) = W^{3\top} \phi(W^{2\top} \phi(W^{1\top}\xi + b^1) + b^2)\\
        \text{with init.}\quad 
        W^1\sim\mathcal{N}(0, \nicefrac{1}{d_{in}}),\ 
        W^{\{2,3\}}\sim\mathcal{N}(0, \nicefrac{1}{n}),\ 
        b^{\{1,2\}}=0,
    \end{gathered}
    \label{eqn:SPMLP}
\end{equation}
\begin{wrapfigure}{r}{0.6\textwidth}
    \centering
    \includegraphics[width=0.6\textwidth]{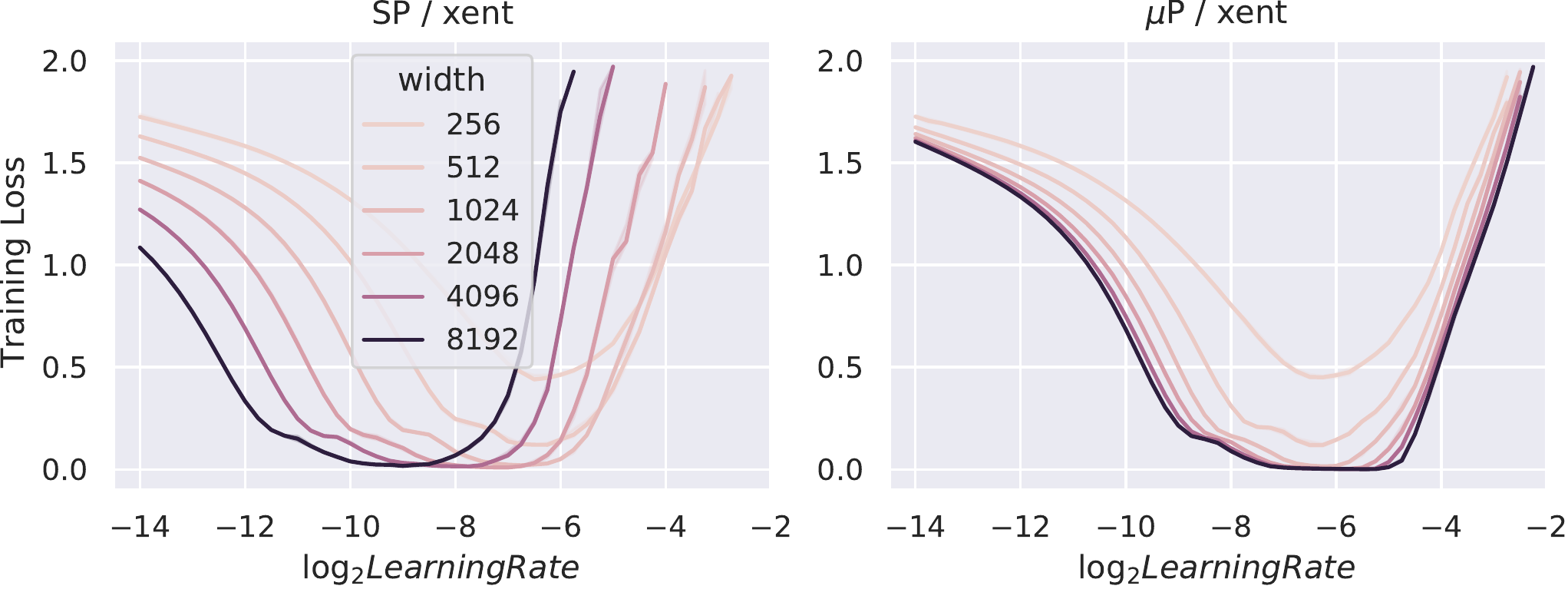}
    \caption[SP vs $\mu$P for MLPs on CIFAR10]{
        MLP width different hidden sizes trained for 20 epoch on CIFAR-10 using SGD. \textbf{Left} uses standard parametrization (SP); \textbf{right} uses maximal update parametrization ($\mu$P). $\mu$P networks exhibit better learning rate stability than their SP counterparts.}
    \label{fig: MLP_relu_xent}
\end{wrapfigure}
where $W^1\in\mathbb{R}^{d_{in}\times n}, b^1\in\mathbb{R}^{n}$,  $W^2\in\mathbb{R}^{n\times n}, b^2\in\mathbb{R}^{n}$, $W^3\in\mathbb{R}^{n\times d_{out}}$ and $d_{in}$, $n$, and $d_{out}$ are the input, hidden, and output dimensions.
The particular MLP we use has $\phi=ReLU$ and a cross-entropy (xent) loss function.
We define the width of MLP as the hidden size $n$, which is varied from 256 to 8192.
The models are trained on CIFAR-10 for 20 epochs, which is more than enough to ensure convergence.

As shown on the left in \cref{fig: MLP_relu_xent}, the optimal learning rate shifts by roughly an order of magnitude as the width increases from 256 to 8192; using the optimal learning of the smallest model on the largest model gives very bad performance, if not divergence.

\paragraph{Transformer with Standard Parametrization}
This perhaps unsurprising observation holds for more complex architectures such as Transformer as well, as shown in \cref{fig: figure1} (left).
We define width as $d_{model}$, with $d_{k}=d_{q}=d_{v}=\nicefrac{d_{model}}{n_{head}}$ and $d_{ffn}=4  d_{model}$.
The models are trained on wikitext-2 for 5 epochs.
In \cref{fig: wikitext2_SP_MUP} in the appendix we also show the instability of initialization scale and other HPs.

\section{Unlocking Zero-Shot Hyperparameter Transfer with \texorpdfstring{$\mu$P}{muP}}
\label{sec:mutransfer}

\begin{table}[t]
  \renewcommand{\hl}[1]{\textcolor{purple}{#1}}
  \renewcommand{\ll}[1]{\textcolor{gray}{#1}}
  \centering
  \caption[$\mu$P\citep{TP4} and SP for General Neural Networks]{
    \textbf{$\mu$P\citep{TP4} and SP for General Neural Networks.}
  Here, we emphasize the \emph{scaling with width ($\fanin$ or $\fanout$)}; in practice, we may insert tunable multipliers in front of $\fanin$ and $\fanout$ as in \cref{eqn:MUPMLP_basewidth}.
  The $\fanout$ of a bias vector is its dimension (whereas $\fanin$ is 1).
  \hl{Purple text} highlights key differences from standard parametrization (SP);
  \ll{Gray text} recalls the corresponding SP.
  \emph{SGD} (resp.\ \emph{Adam}) here can be replaced by variants such as SGD with momentum (resp.\ Adagrad, etc); see \cref{sec:otheroptimizers} for other optimizers.
  In general, the three columns here can be interpreted as linear layers that have \{finite, infinite, infinite\} input dimension and \{infinite, finite, infinite\} output dimension in an infinite-width network; this description generalizes more readily to other parameters such as those of layernorm.
  Transformer $\mu$P requires one more modification ($1/d$ attention instead of $1/\sqrt{d}$); see \cref{defn:trsfmrMUP}.
  This version of $\mu$P gets rid of parameter multipliers; for the version similar to that in \citep{TP4}, see \cref{tab:MUPorig}.
  Also see \cref{tab:MUPalt} for a $\mu$P formulation that is easier to implement (and compatible with input/output weight sharing).
  Further explanation of this table can be found in \cref{sec:furtherexplain}.
  Its derivation can be found in \cref{sec:intuitiveintro2muP}.}
  \begin{tabular}{lllllllll}
      \toprule
      &   \multicolumn{2}{c}{Input weights \& all biases} &  & \multicolumn{2}{c}{Output weights}  &   & \multicolumn{2}{c}{Hidden weights} 
      \\
      \midrule
      Init.\ Var.  & \multicolumn{2}{c}{$\nicefrac{1}{\fanin}$}   &    & \hl{$\nicefrac{1}{\fanin^2}$} & \ll{($\nicefrac{1}{\fanin}$)}          &       & \multicolumn{2}{c}{$\nicefrac{1}{\fanin}$}                
      \\
      SGD LR      & \hl{$\fanout$} & \ll{($1$)}  &  & \hl{$\nicefrac 1 \fanin$} & \ll{($1$)}  & & \multicolumn{2}{c}{$1$}               
      \\
      Adam LR             & \multicolumn{2}{c}{$1$}             &                 & \hl{$\nicefrac 1 \fanin$} & \ll{($1$)}           &                    & \hl{$\nicefrac 1 \fanin$} & \ll{($1$)}              
      \\
      \bottomrule
  \end{tabular}
  \label{tab:MUP}
\end{table}

We show that $\mu$P solves the problems we see in \cref{sec:sp_fails}.

\paragraph{MLP with $\mu$P}
For the MLP in \cref{sec:sp_fails}, to switch to $\mu$P, we just need to modify \cref{eqn:SPMLP}'s initialization of the last layer and its learning rates of the first and last layer as well as of the biases.
The \emph{basic form} is%
\footnote{
  While superficially different, this parametrization is equivalent to the $\mu$P defined in \citep{TP4}.
}
\begin{equation}
  \begin{gathered}
  \text{initialize}\quad
        W^1\sim\mathcal{N}(0, \nicefrac{1}{d_{in}}),\ 
        W^{2}\sim\mathcal{N}(0, \nicefrac{1}{n}),\ 
        W^3 \sim \Gaus(0, \nicefrac 1 {\highlight{n^2}}),\ 
        b^{\{1,2\}}=0
        \\
  \text{with SGD learning rates}\quad
        \eta_{W^1} = \eta_{b^1} = \eta_{b^2} = \eta \highlight{n},\ 
        \eta_{W^2} = \eta,\ 
        \eta_{W^3} = \eta \highlight{n^{-1}}.
  \end{gathered}
  \label{eqn:MUPMLP}
\end{equation}
Here, $\eta$ specifies the ``master'' learning rate, and we highlighted in \highlight{purple} the differences in the two parametrizations.
This basic form makes clear the \emph{scaling with width $n$} of the parametrization, but in practice we will often insert (possibly tune-able) multiplicative constants in front of each appearance of $n$.
For example, this is useful when we would like to be consistent with a SP MLP at a \emph{base width} $n_0$.
Then we may insert constants as follows: For  $\tilde n \defeq n / n_0$,

\begin{equation}
  \begin{gathered}
  \text{initialize}\quad
        W^1\sim\mathcal{N}(0, \nicefrac{1}{d_{in}}),\ 
        W^{2}\sim\mathcal{N}(0, \nicefrac{1}{n}),\ 
        W^3 \sim \Gaus(0, \nicefrac 1 {\highlight{n \cdot \tilde n}}),\ 
        b^{\{1,2\}}=0
        \\
  \text{with SGD learning rates}\quad
        \eta_{W^1} = \eta_{b^1} = \eta_{b^2} = \eta \highlight{\tilde n},\ 
        \eta_{W^2} = \eta,\ 
        \eta_{W^3} = \eta \highlight{\tilde n^{-1}}.
  \end{gathered}
  \label{eqn:MUPMLP_basewidth}
\end{equation}

Then at width $n=n_0$, all \highlight{purple} factors above are 1, and the parametrization is identical to SP (\cref{eqn:SPMLP}) at width $n_0$.
Of course, as $n$ increases from $n_0$, then \cref{eqn:MUPMLP_basewidth} quickly deviates from \cref{eqn:SPMLP}.
In other words, for a particular $n$, $\mu$P and SP can be identical up to the choice of some constants (in this case $n_0$), but $\mu$P determines a different ``set" of networks and optimization trajectory than SP as one varies $n$.
As we will see empirically in the next section, this deviation is crucial for HP transfer.

Indeed, in \cref{fig: MLP_relu_xent}(right), we plot the CIFAR10 performances, over various learning rates and widths, of $\mu$P MLPs with $n_0 = 128$.
In contrast to SP,  the optimal learning rate under $\mu$P is stable.
This means that, the best learning rate for a width-128 network is also best for a width-8192 network in $\mu$P --- i.e.\ HP transfer \emph{works} --- but not for SP.
In addition, we observe performance for a fixed learning rate always weakly improves with width in $\mu$P , but not in SP.

This MLP $\mu$P example can be generalized easily to general neural networks trained under SGD or Adam, as summarized in \cref{tab:MUP}, 
which is derived in \cref{sec:intuitiveintro2muP}.

\paragraph{Transformers with $\mu$P}
We repeat the experiments with base width $n_0=128$ for Transformers:
\begin{defn}\label{defn:trsfmrMUP}
The \emph{Maximal Update Parametrization ($\mu$P) for a Transformer} is given by \cref{tab:MUP} and $1/d$ attention instead of $1/\sqrt{d}$, i.e. the attention logit is calculated as $q^\trsp k/d$ instead of $q^\trsp k / \sqrt d$ where query $q$ and key $k$ have dimension $d$.%
\footnote{This is roughly because during training, $q$ and $k$ will be correlated so $q^\trsp k$ actually scales like $d$ due to Law of Large Numbers, in contrast to the original motivation that $q$, $k$ are uncorrelated at initialization so Central Limit applies instead.
See \cref{{sec:MUPDerivation}} for a more in-depth discussion.}
\end{defn}
The results are shown on the right in \cref{fig: figure1}, where the optimal learning rate is stable, and the performance improves monotonically with width.
See \cref{sec:furtherexplain} for further explanation of $\mu$P.

\begin{figure}
  \centering
  \includegraphics[width=0.85\textwidth]{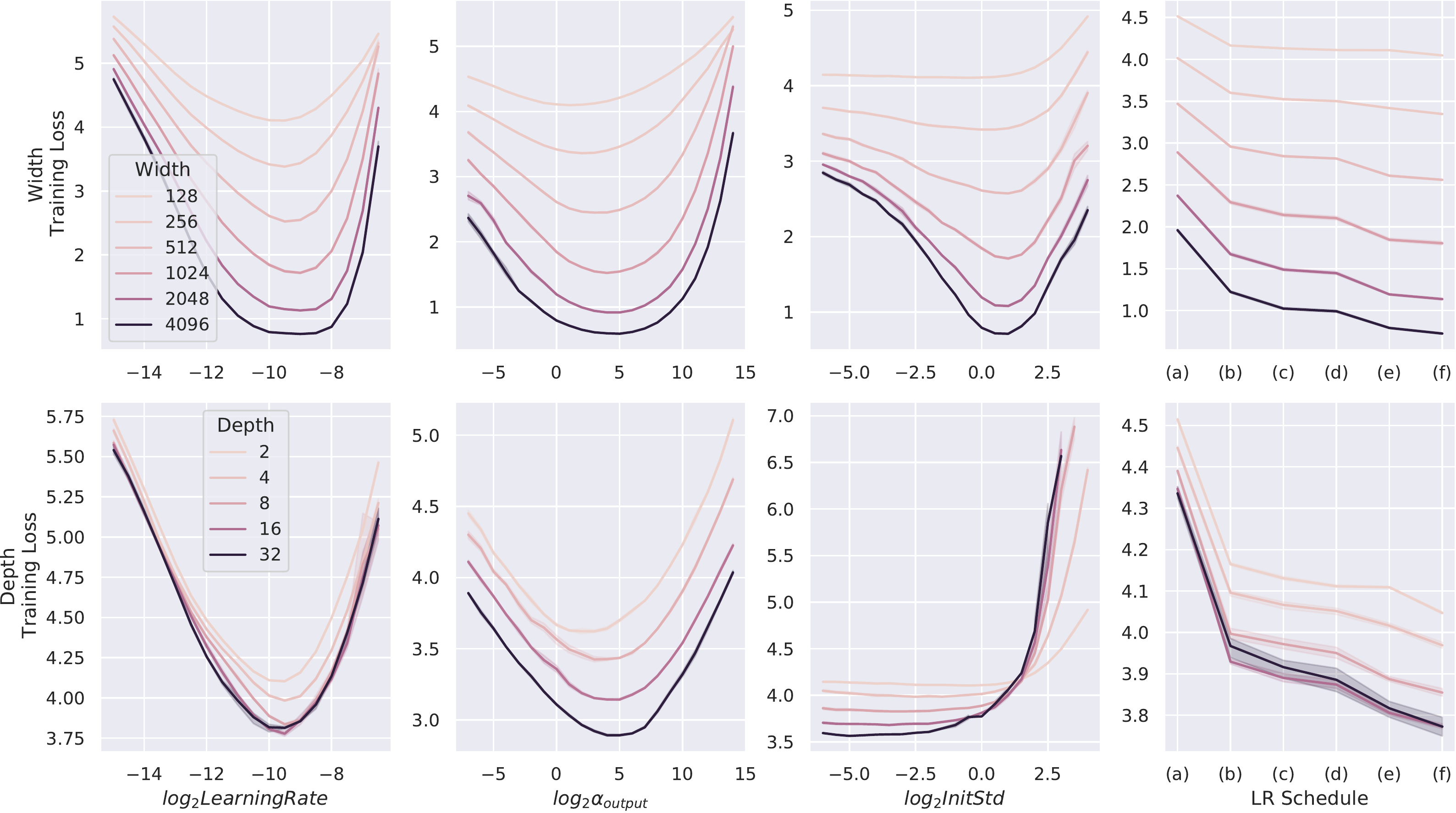}
  \caption[Empirical validation of the stability of four representative hyperparameters on pre-LN Transformers in $\mu$P]{\textbf{Empirical validation of the stability of four representative hyperparameters on pre-LN Transformers in $\mu$P}: learning rate, last layer weight multiplier $\alpha_{output}$, weight initialization standard deviation, and learning rate schedule.
  We use the following learning rate schedules: (a) linear decay; (b) StepLR @ [5k, 8k] with a decay factor of 0.1; (c) StepLR @ [4k, 7k] with a decay factor of 0.3; (d) cosine annealing; (e) constant; (f) inverse square-root decay.
  All models are trained on wikitext-2 for 10k steps.
  When not specified in the legend, the width used is 256, depth 2, batch size 20, sequence length 256, and LR schedule constant.
  We sweep a particular HP, corresponding to each column, while fixing all others constant.
  See \cref{sec:scienceexperiments} for discussion of these results.}
  \label{fig:wikitext2_mega}
\end{figure}

\section{The Defects of SP and How \texorpdfstring{$\mu$}{mu}P Fixes Them}
\label{sec:SPdefect}

The question of SP vs $\mu$P has already been studied at length in \citep{TP4}. 
Here we aim to recapitulate the key insights, with more explanations given in \cref{sec:otherparamdontwork}.

  \begin{figure}
    \centering
    \includegraphics[width=0.7\textwidth]{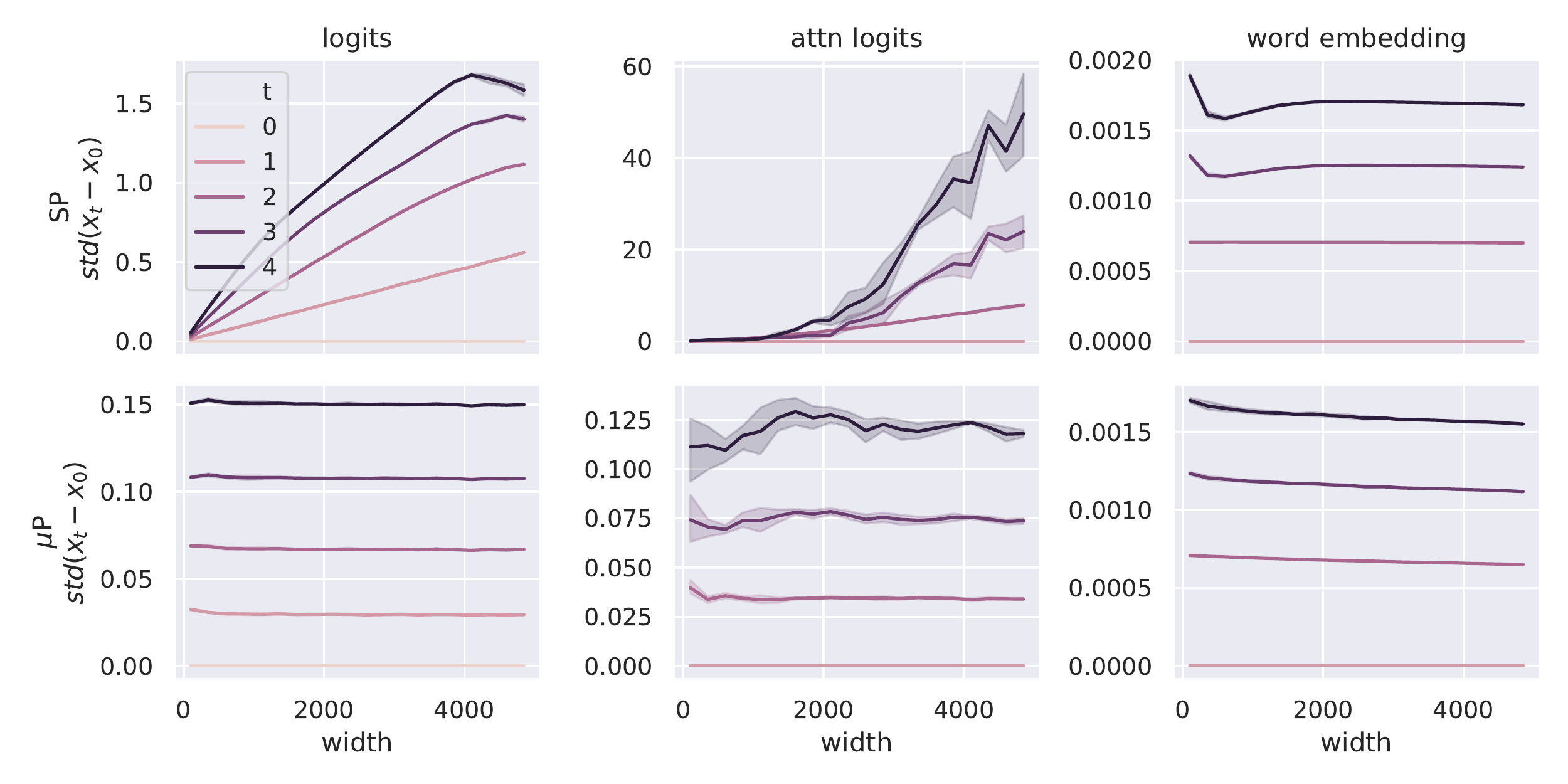}
    \caption[Activations blow up in SP but maintain a consistent scale in $\mu$P]{
      \textbf{Logits and attention logits, but not word embeddings, of a Transformer blow up with width in SP after 1 step of training.}
      In contrast, all three are well-behaved with width in $\mu$P.
      Here we measure how much different values change coordinatewise from initialization over 4 steps of Adam updates, as a function of width.
      Specifically, we plot the standard deviation of the coordinates of $x_t - x_0$, for $t = 0, \ldots, 4$, and $x \in \{\text{logits, attention logits, word embeddings}\}$, where $t=0$ indicates initialization.
      }

      \label{fig:blowupafter1step}
  \end{figure}
  
  \paragraph{An Instructive Example}
  As shown in \citep{TP4} and \cref{sec:otherparamdontwork}, in SP, the network output will blow up with width after 1 step of SGD.
  It's instructive to consider a 1-hidden-layer linear perceptron $f(x) = V^\trsp U x$ with scalar inputs and outputs, as well as weights $V, U \in \R^{n \times 1}$.
  In SP, $V_\alpha \sim \Gaus(0, 1/n)$ ad $U_\alpha \sim \Gaus(0, 1)$ for each $\alpha \in [n]$.
  This sampling ensures that $f(x) = \Theta(|x|)$ at initialization.
  After 1 step of SGD with learning rate 1, the new weights are $V' \gets V + \theta U, U' \gets U + \theta V$, where $\theta$ is some scalar of size $\Theta(1)$ depending on the inputs, labels, and loss function.
  But now
  \begin{equation}
  f(x) = V'{}^\trsp U' x = (V^\trsp U + \theta U^\trsp U + \theta V^\trsp V + \theta^2 U^\trsp V) x
  \label{eqn:lin1lp1step}
  \end{equation}
  blows up with width $n$ because $U^\trsp U = \Theta(n)$ by Law of Large Numbers.

  Now consider the same network in $\mu$P.
  According to \cref{tab:MUP}, we now have $V_\alpha \sim \Gaus(0, 1/n^2)$ in contrast to SP, but $U_\alpha \sim \Gaus(0, 1)$ as before, with learning rates $\eta_{V} = 1/n, \eta_U = n$.
  After 1 step of SGD, we now have
  \[
    f(x) = (V^\trsp U + \theta n^{-1} U^\trsp U + \theta n V^\trsp V + \theta^2 U^\trsp V) x,
  \]
  and one can verify this is $\Theta(1)$ and thus does not blow up with width.%
  \footnote{
  Note in this example, Glorot initialization \citep{glorot_understanding_2010} (i.e.\ with variance $1/(\fanin{} + \fanout{})$) would scale asymptotically the same as $\mu$P and thus is similarly well-behaved.
  However, if one adds layernorm or batchnorm, then Glorot will cause logit blowup like SP, but $\mu$P still will not.
  }
  
  \paragraph{Some Layers Update Too Fast, Others Too Slow}
  One can observe the same behavior in more advanced architectures like Transformers and optimizers like Adam;
  in fact, in SP, other hidden quantities like attention logits will also blow up with width after 1 step, but in $\mu$P still remain bounded, as shown in \cref{{fig:blowupafter1step}}(middle).

  One might think scaling down the learning rate with width can solve this problem in SP.
  However, other hidden activations like the word embedding (\cref{fig:blowupafter1step}(right)) in a Transformer update by a width-independent amount for each step of training, so scaling down the learning rate will effectively mean the word embeddings are not learned in large width models.
  Similar conclusions apply to other models like ResNet (in fact, one can observe in the SP linear MLP example above, the input layer is updated much more slowly than the output layer).
  On the other hand, $\mu$P is designed so that all hidden activations update with the same speed in terms of width (see \cref{sec:DeriveMUP} for why). 
  
  \paragraph{Performance Advantage of $\mu$P}
  This is why a wide model tuned with $\mu$Transfer should in general outperform its SP counterpart with (global) learning rate tuned.
  For example, this is the case for the width-8192 Transformer in \cref{fig: figure1}, where, in SP, the optimal learning rate needs to mollify the blow-up in quantities like logits and attention logits, but this implies others like word embeddings do not learn appreciably.
  This performance advantage means $\mu$Transfer does more than just predicting the optimal learning rate of wide SP models.
  Relatedly, we observe, for any fixed HP combination, training performance never decreases with width in $\mu$P, in contrast to SP (e.g., the $\mu$P curves in \cref{fig: figure1,{fig: MLP_relu_xent},fig: fast_cifar_SP_MUP} do not cross, but the SP curves do; see also \cref{{sec:widerbetter}}).

\section{Which Hyperparameters Can Be \texorpdfstring{$\mu$}{mu}Transferred?}
\label{sec:whichtransfer?}

In this section, we explore how common HPs fit into our framework.
In general, they can be divided into three kinds, summarized in \cref{tab:whichtransfer}: 
\begin{enumerate}
    \item those that can transfer from the small to the large model, such as learning rate (\cref{tab:transferables});
    \item those that primarily control regularization and don't work well with our technique; and
    \item those that define training \emph{scale}, such as width as discussed above as well as others like depth and batch size, across which we transfer other HPs.
\end{enumerate}

Those in the first category transfer across width, as theoretically justified above in \cref{sec:primer}.
To push the practicality and generality of our technique, we empirically explore the transfer across the other dimensions in the third category.
Note that $\mu$Transfer across width is quite general, e.g.\ it allows varying width ratio of different layers or number of attention heads in a Transformer; see \cref{sec:width}.
This will be very useful in practice.
For the second category, the amount of regularization (for the purpose of controlling overfitting) naturally depends on both the model size and data size, so we should not expect transfer to work if the parametrization only depends on model size.
We discuss these HPs in more detail in \cref{sec:hypdiscussions}.

\subsection{Empirical Validation and Limitations}
\label{sec:scienceexperiments}
Our empirical investigations focus on Transformers (here) and ResNet (in \cref{sec:resnetcifar10experiments}), the most popular backbones of deep learning models today.
We train a 2-layer pre-layernorm $\mu$P\footnote{``2 layers'' means the  model has 2 self-attention blocks. To compare with SP Transformer, see \cref{{fig: wikitext2_SP_MUP}}.} Transformer with 4 attention heads on Wikitext-2.
We sweep one of four HPs (learning rate, output weight multiplier, initialization standard deviation, and learning rate schedule) while fixing the others and sweeping along width and depth (with additional results in \cref{fig:wikitext2_bsz_seqlen_step} on transfer across batch size, sequence length, and training time).
\cref{fig:wikitext2_mega} shows the results averaged over 5 random seeds.

Empirically, we find that for language modeling on Transformers, HPs generally transfer across scale dimensions if some minimum width (e.g.\ 256), depth (e.g., 4), batch size (e.g., 32), sequence length (e.g., 128), and training steps (e.g., 5000) are met, and the target scale is within the ``reasonable range'' as in our experiments.
Now, there are some caveats.
While the exact optimum can shift slightly with increasing scale, this shift usually has very small impact on the loss, compared to SP (\cref{fig: figure1,fig: MLP_relu_xent}(left)).
However, there are some caveats.
For example, the best initialization standard deviation does not seem to transfer well across depth (2nd row, 3rd column), despite having a stabler optimum across width.
In addition, while our results on width, batch size, sequence length, and training time still hold for post-layernorm (\cref{fig:wikitext2_postln}),%
\footnote{in fact, post-layernorm Transformers are much more sensitive to HPs than pre-layernorm, so our technique is more crucial for them, especially for transfer across width.
\cref{fig: figure1} uses post-layernorm.} the transfer across depth only works for pre-layernorm Transformer.
Nevertheless, in practice (e.g.\ our results in \cref{sec:BERT}) we find that fixing initialization standard deviation while tuning other HPs works well when transferring across depth.

\section{Efficiency and Performance of \texorpdfstring{$\mu$}{mu}Transfer}
\label{sec:transfermainresults}

Now that the plausibility of $\mu$Transfer has been established in toy settings, we turn to more realistic scenarios to see if one can achieve tangible gains.
Specifically, we perform HP tuning only on a smaller proxy model, test the obtained HPs on the large target model directly, and compare against baselines tuned using the target model.
We seek to answer the question: Can $\mu$Transfer make HP tuning more efficient while achieving performance on par with traditional tuning? %
As we shall see by the end of the section, the answer is positive.
We focus on Transformers here, while experiments on ResNets on CIFAR10 and Imagenet can be found as well in \cref{sec:resnetexperiments}.
All of our experiments are run on V100 GPUs.

\subsection{Transformer on IWSLT14 De-En}

\paragraph{Setup}
IWSLT14 De-En is a well-known machine translation benchmark.
We use the default IWSLT (post-layernorm) Transformer implemented in \texttt{fairseq}~\citep{ott2019fairseq} with 40M parameters, which we denote as the \emph{1x model}.%
\footnote{\url{https://github.com/pytorch/fairseq/blob/master/examples/translation/README.md}.}
For $\mu$Transfer, we tune on a \emph{0.25x model} with $1/4$ of the width, amounting to 4M parameters.
For this experiment, we tune via random search the learning rate $\eta$, the output layer parameter multiplier $\alpha_{output}$, and the attention key-projection weight multiplier $\alpha_{attn}$.
See the grid and other experimental details in \cref{sec:IWSLTdetails}.

We compare transferring from the 0.25x model with tuning the 1x model while controlling the total tuning budget in FLOPs.%
\footnote{Ideally we would like to measure the wall clock time used for tuning.
However, smaller models such as the proxy Transformer used for IWSLT are not efficient on GPUs, so wall clock time would not reflect the speedup for larger models like GPT-3.
Thus, we measure in FLOPs, which is less dependent on hardware optimization.}
To improve the reproducibility of our result: 1) we repeat the entire HP search process (a \emph{trial}) 25 times for each setup, with number of samples as indicated in \cref{tab: iwslt14_transfer}, and report the 25th, 50th, 75th, and 100th percentiles in BLEU score; 2) we evaluate each selected HP combination using 5 random initializations and report the mean performance.\footnote{We do not report the standard deviation over random initializations to avoid confusion.}

We pick the HP combination that achieves the lowest validation loss\footnote{We find this provides more reliable result than selecting for the best BLEU score.} for each trial.
The reported best outcome is chosen according to the validation loss during tuning.
We compare against the default in \texttt{fairseq}, which is presumably heavily tuned.
The result is shown in \cref{tab: iwslt14_transfer}.

\begin{table}[t]
    \centering
    \small
    \caption[$\mu$Transfer results for Transformer on IWSLT14 De-En]{\textbf{Transformer on IWSLT14 De-En.}
    1x and 0.25x refers to scaling of width only. Compared to traditional tuning (``Tuning on 1x''), $\mu$Transfer from 0.25x provides better and more reliable outcome given fixed amount of compute.
    On the other hand, naive transfer (i.e.\ with SP instead of $\mu$P) fails completely.
    The percentiles are over independent trials, with each trial involving the entire tuning process with a new HP random search.
    }
    \begin{tabular}{ccccccc}
        \toprule
        \multicolumn{3}{c}{}                                     & \multicolumn{4}{c}{Val. BLEU Percentiles} \\
        Setup                    & Total Compute & \#Samples & 25 & 50             & 75 & 100            \\ \midrule
        \texttt{fairseq}\citep{ott2019fairseq} default & -             & -             & -   & -              &  -  & 35.40          \\ \midrule
        Tuning on 1x             & 1x            & 5             & 33.62   & 35.00    & 35.35   & 35.45          \\
        Naive transfer from 0.25x          & 1x            & 64            & \multicolumn{4}{c}{training diverged}\\
        $\mu$Transfer from 0.25x (Ours)         & 1x            & 64            & \textbf{35.27}   & \textbf{35.33} & \textbf{35.45}  & \textbf{35.53}
        \\ \bottomrule
        \end{tabular}
    \label{tab: iwslt14_transfer}
\end{table}

\begin{wrapfigure}{r}{0.5\textwidth}
  \centering
  \includegraphics[width=0.5\textwidth]{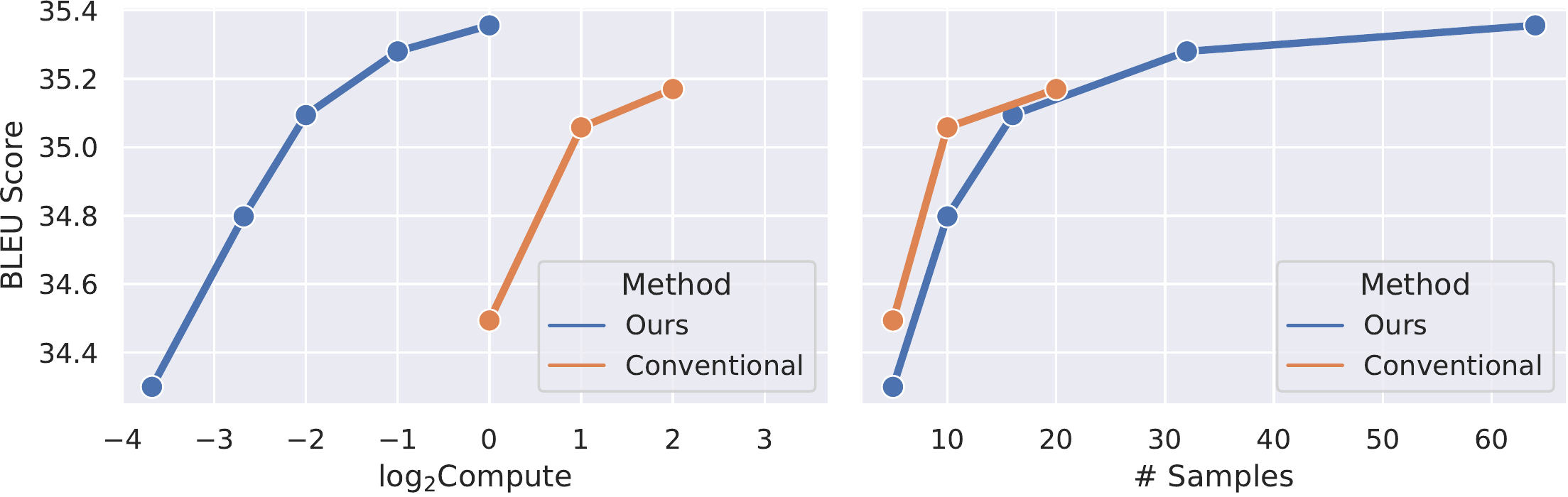}
  \caption[Efficiency-performance Pareto frontier of $\mu$Transfer ]{
      \textbf{Efficiency-performance Pareto frontier} of $\mu$Transfer compared to conventional tuning, on IWSLT Transformer, using random HP search as the base method.
  We plot the \emph{median} BLEU score over 25 trials (Left) against relative compute budget in log scale and (Right) against number of HP samples taken.
  While with the same number of samples, $\mu$Transfer slightly underperforms conventional tuning, this gap vanishes with more samples, and in terms of compute, our Pareto frontier strongly and consistently dominates that of conventional tuning.
  Note that, in larger models (e.g.\ BERT or GPT-3, not shown here), we believe our efficiency advantage will only widen as our small proxy model can stay the same size while the target model grows.}
  \label{fig:iwslt14_pareto}
  \vspace{-2em}
\end{wrapfigure}
\paragraph{Performance Pareto Frontier}
The result above only describes a particular compute budget.
Is $\mu$Transfer still preferable when we have a lot more (or less) compute?
To answer this question, we produce the compute-performance Pareto frontier in \cref{fig:iwslt14_pareto}(left), where we repeat the above experiment with different compute budgets.
Evidently, our approach completely dominates conventional tuning.

\paragraph{Sample Quality of Proxy Model vs Target Model}
The Pareto frontier in \cref{fig:iwslt14_pareto}(right) suggests that, given a fixed number of random \emph{samples} from the HP space, 1) tuning the target model directly yields slightly better results than tuning the proxy model (while taking much more compute of course),
but 2) this performance gap seems to vanish as more samples are taken.
This can be explained by the intuition that the narrower proxy model is a ``noisy estimator'' of the wide target model \citep{TP4}.%
With few samples, this noise can distort the random HP search, but with more samples, this noise is suppressed.%

\subsection{Transformer on WMT14 En-De}

We scale up to WMT14 En-De using the large (post-layernorm) Transformer from~\cite{DBLP:journals/corr/VaswaniSPUJGKP17} with 211M parameters.
We tune on a proxy model with 15M parameters by shrinking $d_{model}$, $d_{ffn}$, and $n_{head}$.
For this experiment, we tune via random search the learning rate $\eta$, the output layer parameter multiplier $\alpha_{output}$, and the attention key-projection weight multiplier $\alpha_{attn}$ following the grid in \cref{sec:WMTdetails}.
The result is shown in \cref{tab: wmt14_transfer}:
While random search with 3 HP samples far underperforms the \texttt{fairseq} default, we are able to match it via transfer using the same tuning budget.

\begin{table}[tph]
    \centering
    \small
    \caption[$\mu$Transfer results for Transformer on WMT14 En-De]{\textbf{Transformers on WMT14 En-De.}
     1x and 0.25x refers to scaling of width only.
     We report BLEU fluctuation over 3 independent trials, i.e., 3 independent random HP searches.}
    \begin{tabular}{cccccc}
        \toprule
        \multicolumn{3}{c}{}                                     & \multicolumn{3}{c}{Val. BLEU Percentiles}         \\
        Setup                    & Total Compute & \#Samples & Worst           & Median         & Best           \\ \midrule
        \texttt{fairseq}\citep{ott2019fairseq} default & -             & -             & -               & -              & 26.40          \\ \midrule
        Tuning on 1x             & 1x            & 3             & \multicolumn{2}{c}{training diverged}          & 25.69          \\
        Naive transfer from 0.25x          & 1x            & 64            & \multicolumn{3}{c}{training diverged} \\
        $\mu$Transfer from 0.25x (Ours)         & 1x            & 64            & \textbf{25.94}  & \textbf{26.34} & \textbf{26.42} \\ \bottomrule
                \end{tabular}
    \label{tab: wmt14_transfer}
\end{table}

\subsection{BERT}
\label{sec:BERT}

Finally, we consider large-scale language model pretraining where HP tuning is known to be challenging.
Using Megatron (pre-layernorm) BERT~\cite{megatron} as a baseline, we hope to recover the performance of the published HPs by only tuning a proxy model that has roughly 13M parameters, which we call \emph{BERT-prototype}.
While previous experiments scaled only width, here we will also scale depth, as discussed in \cref{sec:whichtransfer?} and validated in \cref{fig:wikitext2_mega}.
We use a batch size of 256 for all runs and follow the standard finetuning procedures.
For more details on BERT-prototype, what HPs we tune, and how we finetune the trained models, see \cref{app:bert}.

During HP tuning, we sample 256 combinations from the search space and train each combination on BERT-prototype for $10^5$ steps.
The total tuning cost measured in FLOPs is roughly the same as training 1 BERT-large for the full $10^6$ steps; the exact calculation is shown in \cref{app:bert_prototype}.
The results are shown in \cref{tab: large_scale_LM}.
Notice that on BERT-large, we obtain sizeable improvement over the well-tuned Megatron BERT-large baseline.

\begin{table}[tph]
    \centering
    \small
    \caption[$\mu$Transfer results for BERT pretraining]{\textbf{BERT pretraining.}
    HP transfer outperforms published baselines without tuning the full model directly at all.
    We tune BERT-base and BERT-large simultaneously via a single proxy model, \emph{BERT-prototype}.
    The total tuning cost = the cost of pretraining a single BERT-large.
    \emph{Model speedup} refers to the training speedup of BERT-prototype over BERT-base or BERT-large.
    \emph{Total speedup} in addition includes time saving from transferring across training steps.
    Both speedups can be interpreted either as real-time speedup on V100s or as FLOPs speedup (which turn out to be empirically very similar in this case).
    }
    \begin{tabular}{ccccccc} \toprule
         Model                     & Method            & Model Speedup   & Total Speedup & Test loss & MNLI (m/mm) & QQP \\ \midrule
         $\textsc{BERT}_{base}$    & Megatron Default  &  1x              &  1x              & 1.995     &     84.2/84.2    & 90.6    \\
         $\textsc{BERT}_{base}$    & Naive Transfer    &  4x              &  40x              & \multicolumn{3}{c}{training diverged}   \\
          $\textsc{BERT}_{base}$    & $\mu$Transfer (Ours)        &  4x              &  40x              & \textbf{1.970}    & \textbf{84.3/84.8} & \textbf{90.8}   \\ \midrule
         $\textsc{BERT}_{large}$   & Megatron Default  &  1x              &  1x              & 1.731     &         86.3/86.2    & 90.9\\
          $\textsc{BERT}_{large}$   & Naive Transfer   &  22x             &  220x              & \multicolumn{3}{c}{training diverged}\\ 
         $\textsc{BERT}_{large}$   & $\mu$Transfer (Ours)         &  22x             &  220x              & \textbf{1.683}      &   \textbf{87.0/86.5} & \textbf{91.4}\\ \bottomrule
    \end{tabular}
    \label{tab: large_scale_LM}
\end{table}

\subsection{GPT-3}

In order to further verify $\mu$Transfer at scale, we applied it to GPT-3 6.7B \cite{brown2020language} with relative attention.
This \emph{target model} consists of 32 residual blocks with width 4096.
We form the small \emph{proxy model} by shrinking width to 256, resulting in roughly 40 million trainable parameters, 168 times smaller than the target model.
HPs were then determined by a random search on the proxy model.
The total tuning cost was only 7\% of total pretraining cost.
Details of the HP sweep can be found in \cref{sec:gpt3-appendix}.

In order to exclude code difference as a possible confounder, we also re-trained GPT-3 6.7B from scratch using the original HPs from \cite{brown2020language}.
Unfortunately, after we have finished all experiments, we found this baseline mistakenly used absolute attention (like models in \cite{brown2020language}) when it was supposed to use relative attention like the target model.
In addition, during training of the $\mu$Transfer model we encountered numerical issues that lead to frequent divergences.
In order to avoid them, the model was trained using FP32 precision, even though the original 6.7B model and our re-run were trained using FP16.%
\footnote{
  While we are mainly focused on the efficacy of $\mu$Transfer regardless of precision, it would be interesting to ablate the effect of precision in our results, but we did not have enough resources to rerun the baseline in FP32}
\footnote{
  It is quite interesting that $\mu$Transfer identified a useful region of hyperparameters leading to much improved performance, which probably would be difficult to discover normally because 1) researchers usually change hyperparameters to accomodate precision and 2) there was no precise enough justification to go against this judgment until $\mu$Transfer.
}
The resulting $\mu$Transfer model outperforms the 6.7B from \cite{brown2020language}, and is in fact comparable to the twice-as-large 13B model across our evaluation suite (see \cref{{tab:gpt3-evals-13b-comparison}}).
Selected evaluation results can be found in \cref{tab:gpt3-main-evals} and further details are given in \cref{tab:gpt3-evals} and \cref{sec:gpt3-appendix}.

\begin{table}[]
    \centering
    \caption[$\mu$Transfer results for GPT-3 pretraining]{\textbf{GPT-3 6.7B Pretraining.} Selected evaluation results for the GPT-3 6.7B model tuned with $\mu$Transfer (transfered from a small proxy model of 40M parameters), compared to the results published in \cite{brown2020language} and a re-run with original HPs, as well as the 13B model in \cite{brown2020language} for reference.
    Note that the perplexities in this table are based on a custom tokenization and are not comparable to the literature. The validation loss refers to the loss achieved on a random held-out part of our dataset. \emph{Zero-shot}, \emph{One-Shot} and \emph{Few-Shot} refer to the number of additional query and answer pairs passed in the context when performing the sampling-based evaluations.
    See \cref{sec:gpt3-appendix} for full evaluation.}
    \begin{tabular}{llSSSS}
    \toprule
        Task & Metric & {6.7B+$\mu$P} & {6.7B re-run} & {6.7B \cite{brown2020language}} & {13B \cite{brown2020language}} \\
    \midrule
        Validation loss & cross-entropy & \bfseries 1.98 & 2.03 & {-} & {-} \\
        PTB & perplexity & \bfseries 11.39 & 13.00 & {-} & {-} \\
        WikiText-103 & perplexity & \bfseries 8.56 & 9.13 & {-} & {-} \\
        One Billion Words & perplexity & \bfseries 20.51 & 21.70 & {-} & {-} \\
        LAMBADA Zero-Shot & accuracy & \bfseries 73.4912 & 70.8131194114685 &  70.3 & 72.5 \\
        LAMBADA One-Shot & accuracy & \bfseries 69.8816 & 64.75839614868164 & 65.4 & 69.0 \\
        LAMBADA Few-Shot & accuracy & 74.71376061439514 & 77.12012529373169 & \bfseries 79.1000  & \bfseries 81.3\\
        HellaSwag Zero-Shot & accuracy & \bfseries 72.0 & 66.7 & 67.4 & 70.9 \\
        HellaSwag One-Shot & accuracy & \bfseries 71.1 & 65.9 & 66.5 & 70.0 \\
        HellaSwag Few-Shot & accuracy & \bfseries 72.4 & 66.4 & 67.3 & 71.3\\
    \bottomrule
    \end{tabular}
    \label{tab:gpt3-main-evals}
\end{table}

\begin{figure}
  \centering
  \includegraphics[width=0.9\textwidth]{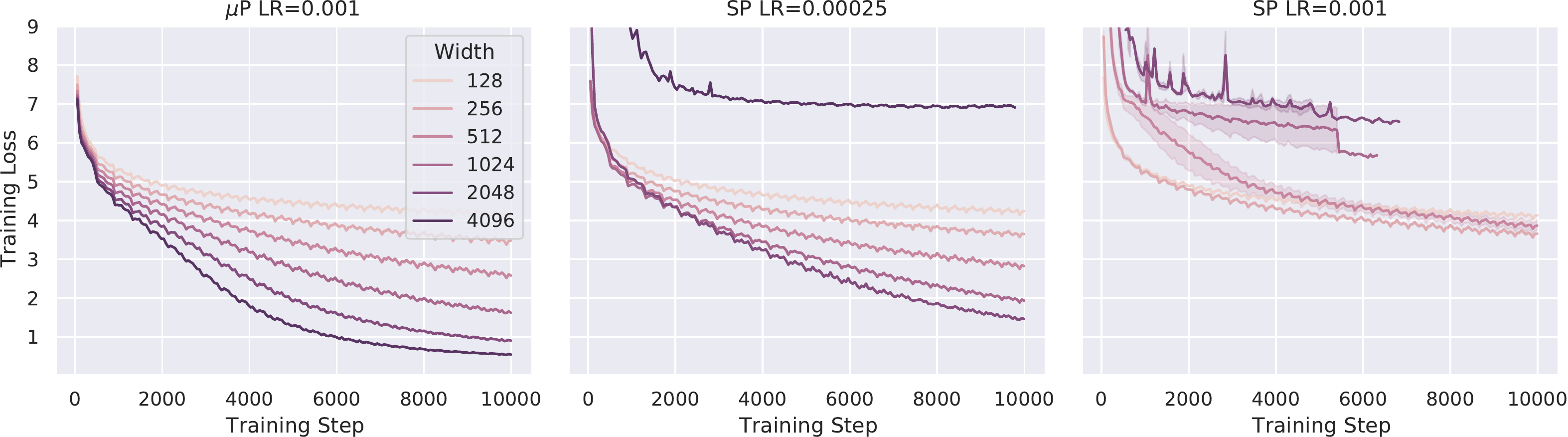}
  \caption[Wider is always better in training loss under $\mu$P, but not in SP, given the same HP]{\textbf{Wider is always better in training loss under $\mu$P, but not in SP, given the same HP.} Learning curves for $\mu$P and SP with different learning rates, aggregated over 5 seeds.
  \textbf{(Left)} Wider $\mu$P models always achieve better training loss at any time in training.
  \textbf{(Middle)} If using a small learning rate, SP models can appear to do so up to some large width, at which point the pattern fails (at width 2048 in our plot).
  \textbf{(Right)} If using a large learning rate, SP model can strictly do \emph{worse} with width; here the SP model is identical to the $\mu$P model in (Left) at width 128.}
  \label{fig:wikitext2_wider_better}
\end{figure}

\section{Wider is Better in \texorpdfstring{$\mu$}{mu}P \emph{Throughout Training}}
\label{sec:widerbetter}
~
\begin{wrapfigure}{r}{0.51\textwidth}
  \vspace{-2em}
  \center
  \includegraphics[width=0.4\textwidth]{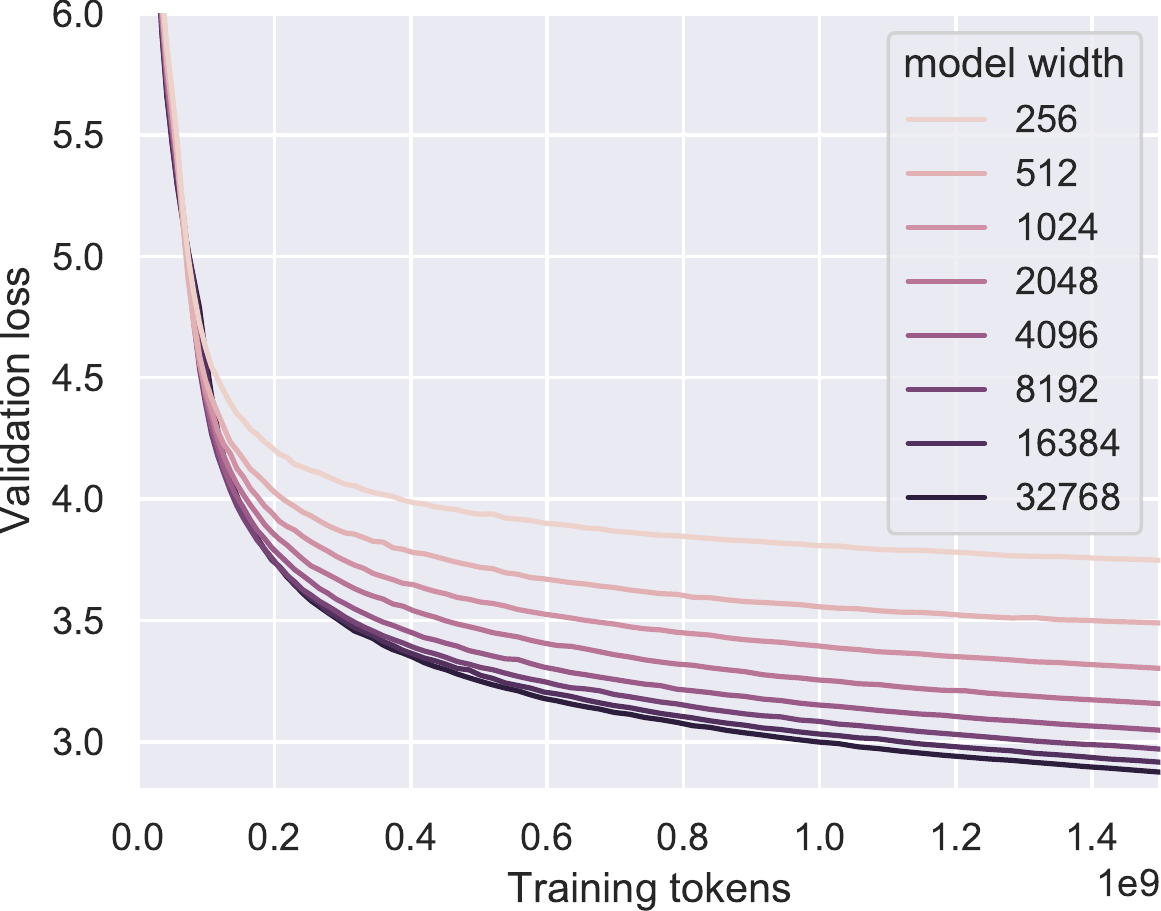}
  \caption[Stress-testing ``wider-is-better'' in $\mu$P]{
      \textbf{Stress-testing ``wider-is-better'' in $\mu$P.}
      Here we trained a GPT-3 transformer with 4 layers and widths from 256 to 32,768.
      Modulo a brief period around 1e8 training tokens, wider is better throughout training.
  }
  \vspace{-1em}
  \label{fig:oai_width_result}
\end{wrapfigure}
In earlier plots like \cref{{fig: figure1},{fig: MLP_relu_xent}}, we saw that at the end of training, wider is always better in $\mu$P but not in SP.
In fact, we find this to be true \emph{throughout training}, as seen in \cref{fig:wikitext2_wider_better}, modulo noise from random initialization and/or data ordering, and assuming the output layer is zero-initialized (which has no impact on performance as discussed in \cref{sec:outputzeroinit}).
We then stress-tested this on a $\mu$P GPT-3 Transformer (on the GPT-3 training data) by scaling width from 256 to 32,768 using a fixed set of HPs (\cref{fig:oai_width_result}).
Wider models consistently match or outperform narrower models at each point in training (except a brief period around 1e8 training tokens, likely due to noise because we ran only 1 seed due to computational cost).
Our observation suggests that wider models are strictly more data-efficient if scaled appropriately.
By checking ``wider-is-better'' early in training, one can also cheaply debug a $\mu$P implementation.

\section{\emph{Useful} Hyperparameter Transfer: A Theoretical Puzzle}

We want to tune HPs on a small model with width $N$ such that its HP landscape looks like that of a large model with width $\gg N$. 
Our intuition in \cref{{sec:primer},{sec:primermulti},{sec:intuitiveintro2muP}} leads us to $\mu$P.
However, for this to be useful, we \emph{do not want} the small model (as a function) after training to be close to that of the large model --- otherwise there is no point in training the large model to begin with. So $N$ 1) must be large enough so that the HP optimum converges, but 2) cannot be so large that the functional dynamics (and the loss) converges. 
The fact that such $N$ exists, as demonstrated by our experiments, shows that:
In some sense, the HP optimum is a ``macroscopic'' or ``coarse'' variable which converges quickly with width, while the neural network function (and its loss) is a very ``microscopic'' or ``fine'' detail that converges much more slowly with width.
However, theoretically, it is unclear why this should happen, and where else we should expect such \emph{useful} HP transfer.
We leave an explanation to future work.

\section{Related Works}

\label{sec:relatedworks}

\subsection{Hyperparameter Tuning}
\label{sec:relatedworks_hyptune_appendix}

Many have sought to speedup HP tuning beyond the simple grid or random search.
\citet{snoek2012practical} treated HP tuning as an optimization process and used Bayesian optimization by treating the performance of each HP combination as a sample from a Gaussian process (GP).
\citet{snoek2015scalable} further improved the runtime by swapping the GP with a neural network.
Another thread of work investigated how massively parallel infrasture can be used for efficient tuning under the multi-arm bandit problem~\cite{jamieson2015nonstochastic, li2020massively}.
There are also dedicated tools such as Optuna~\cite{akiba2019optuna} and Talos~\cite{talos} which integrate with existing deep learning frameworks and provide an easy way to apply more advanced tuning techniques.

Our approach is distinct from all of the above in that it does not work on the HP optimization process itself.
Instead, it decouples the size of the target model from the tuning cost, which was not feasible prior to this work.
This means that \textbf{no matter how large the target model is, we can always use a fixed-sized proxy model to probe its HP landscape}.
Nevertheless, our method is complementary, as the above approaches can naturally be applied to the tuning of the proxy model;
it is only for scientific reasons that we use either grid search or random search throughout this work.

\subsection{Hyperparameter Transfer}
\label{sec:hptransfer_related_works}
Many previous works explored transfer learning of HP tuning (e.g.\ \citep{yogatama_efficient_2014,perrone_scalable,stoll2021hyperparameter,hptransferadaptive}).
However, to the best of our knowledge, our work is the first to explore \emph{zero-shot} HP transfer.
In addition, we focus on transferring across model scale rather than between different tasks or datasets.
Some algorithms like Hyperband \citep{li_hyperband} can leverage cheap estimates of HP evaluations (like using a small model to proxy a large model) but they are not zero-shot algorithms, so would still be very expensive to apply to large model training.
Nevertheless, all of the above methods are complementary to ours as they can be applied to the tuning of our proxy model.

\subsection{Previously Proposed Scaling Rules of Hyperparameters}
\label{sec:previousscalings_appendix}
\paragraph{(Learning Rate, Batch Size) Scaling}
\citep{smith_dont_2017} proposed to scale learning rate with batch size while fixing the total epochs of training; \citep{hoffer_train_2017} proposed to scale learning rate as $\sqrt{batchsize}$ while fixing the total number of steps of training.
However, \citep{shallue_measuring_2018} showed that there's no consistent (learning rate, batch size) scaling law across a range of dataset and models.
Later, \citep{mccandlish_empirical_2018} studied the trade-off of training steps vs computation as a result of changing batch size.
They proposed an equation of $a/(1 + b / batchsize)$, where $a$ and $b$ are task- and model-specific constants, for the optimal learning rate (see their fig 3 and fig 5).
This law suggests that for sufficiently large batch size, the optimal learning rate is roughly constant.%
\footnote{while the optimal learning is roughly linear in batch size when the latter is small}
This supports our results here as well as the empirical results in \citep[fig 8]{shallue_measuring_2018}.

\paragraph{Learning Rate Scaling with Width}

Assuming that the optimal learning rate should scale with batch size following \citep{smith_dont_2017}, \citep{park_effect_2019}
empirically investigated how the optimal ``noise ratio'' $LR/batchsize$ scales with width for MLP and CNNs in NTK parametrization (NTP) or standard parametrization (SP) trained with SGD.
They in particular focus on test loss in the regime of small batch size and training to convergence.
In this regime, they claimed that in networks without batch normalization, the optimal noise ratio is constant in SP but scales like $1/width$ for NTP.
However, they found this law breaks down for networks with normalization.

In contrast, here we focus on training loss, without training to convergence and with a range of batch sizes from small to very large (as is typical in large scale pretraining).
Additionally, our work applies universally to 1) networks with normalization, along with 2) Adam and other adaptive optimizers; furthermore 3) we empirically validate transfer across depth and sequence length, and 4) explicitly validate tuning via $\mu$Transfer on large models like BERT-large and GPT-3.

Finally, as argued in \citep{TP4} and \cref{{sec:otherparamdontwork}}, SP and NTP lead to bad infinite-width limits in contrast to $\mu$P and hence are suboptimal for wide neural networks.
For example, sufficiently wide neural networks in SP and NTP would lose the ability to learn features, as concretely demonstrated on word2vec in \citep{TP4}.

\paragraph{Input Layer Parametrization}
The original formulation of $\mu$P in \citep{TP4} (see \cref{tab:MUPorig}, which is equivalent to \cref{tab:MUP}) uses a fan-out initialization for the input layer.
This is atypical in vision models, but in language models where the input and output layers are shared (corresponding to word embeddings), it can actually be more natural to use a fan-out initialization (corresponding to fan-in initialization of the output layer).
In fact, we found that \texttt{fairseq} \citep{ott2019fairseq} by default actually implements both the fan-out initialization and the $\sqrt{\fanout}$ multiplier.%

\paragraph{Other Scaling Rules}
Many previous works proposed different initialization or parametrizations with favorable properties, such as better stability for training deep neural networks \cite{glorot_understanding_2010,schoenholz_deep_2016,yang_mean_2017,yang_layerwise_2018,zhang_residual_2019,bachlechner_rezero_2020,huang_improving_nodate,liu_understanding_2020}.
Our work differs from these in that we focus on the transferability of optimal HPs from small models to large models in the same parametrization.

\subsection{Infinite-Width Neural Networks: From Theory to Practice and Back}

\citep{TP4} introduced $\mu$P as the unique parametrization that enables all layers of a neural network to learn features in the infinite-width limit, especially in contrast to the NTK parametrization \citep{jacot_neural_2018} (which gives rise to the NTK limit) that does not learn features in the limit.
Based on this theoretical insight, in \cref{{sec:otherparamdontwork}}, we argue that $\mu$P should also be the \emph{unique} parametrization (in the sense of \citep{TP4}) that allows HP transfer across width;
in short this is because it both 1) preserves feature learning, so that performance on feature learning tasks (such as language model pretraining) does not become trivial in the limit, and 2) ensures each parameter tensor is not stuck at initialization in the large width limit, so that its learning rate does not become meaningless.
At the same time, our results here suggest that $\mu$P is indeed the \emph{correct} parametrization for wide neural networks and thus provide empirical motivation for the theoretical study of the infinite-width $\mu$P limit.
Note, \emph{parametrization} here refers to a rule to scale hyperparameters with width (``how should my initialization and learning rate change when my width doubles?''), which is coarser than a prescription for setting hyperparameters at any particular width (``how should I set my initialization and learning rate at width 1024?'').

\section{Conclusion}
\label{sec:conclusion}
Leveraging the discovery of a feature learning neural network infinite-width limit, we hypothesized and verified that the HP landscape across NNs of different width is reasonably stable if parametrized according to Maximal Update Parametrization ($\mu$P).
We further empirically showed that it's possible to transfer across depth, batch size, sequence length, and training time, with a few caveats.
This allowed us to indirectly tune a very large network by tuning its smaller counterparts and transferring the HPs to the full model.
Our results raise an interesting new theoretical question of how \emph{useful HP transfer} is possible in neural networks in the first place.

\paragraph{Venues of Improvement}
Nevertheless, our method has plenty of room to improve.
For example, initialization does not transfer well across depth, and depth transfer generally still does not work for post-layernorm Transformers.
This begs the question whether a more principled parametrization in depth could solve these problems.
Additionally, \cref{fig:wikitext2_mega} shows that the optimal HP still shifts slightly for smaller models.
Perhaps by considering finite-width corrections to $\mu$P one can fix this shift.
Finally, it will be interesting to study if there's a way to transfer regularization HPs as a function of both the model size and data size, especially in the context of finetuning of pretrained models.

\newpage

\paragraph{Acknowledgements}

In alphabetical order, we thank
Arthur Jacot,
Arturs Backurs,
Colin Raffel,
Denny Wu,
Di He,
Huishuai Zhang,
Ilya Sutskever,
James Martens,
Janardhan Kulkarni,
Jascha Sohl-Dickstein,
Jeremy Bernstein,
Lenaic Chizat,
Luke Metz,
Mark Chen,
Michael Santacroce,
Muhammad ElNokrashy,
Pengchuan Zhang,
Sam Schoenholz,
Sanjeev Arora,
Taco Cohen,
Yiping Lu,
Yisong Yue, and
Yoshua Bengio
for discussion and help during our research.

\bibliography{main}

\begin{thebibliography}{66}
\providecommand{\natexlab}[1]{#1}
\providecommand{\url}[1]{\texttt{#1}}
\expandafter\ifx\csname urlstyle\endcsname\relax
  \providecommand{\doi}[1]{doi: #1}\else
  \providecommand{\doi}{doi: \begingroup \urlstyle{rm}\Url}\fi

\bibitem[nvi()]{nvidia_resnet50}
{NVIDIA}/{DeepLearningExamples}, apache v2 license.
\newblock URL \url{https://github.com/NVIDIA/DeepLearningExamples}.

\bibitem[dav(2019)]{davidnet}
Davidnet, mit license, 2019.
\newblock URL \url{https://github.com/davidcpage/cifar10-fast}.

\bibitem[tal(2019)]{talos}
Autonomio talos, mit license, 2019.
\newblock URL \url{http://github.com/autonomio/talos}.

\bibitem[Akiba et~al.(2019)Akiba, Sano, Yanase, Ohta, and
  Koyama]{akiba2019optuna}
Takuya Akiba, Shotaro Sano, Toshihiko Yanase, Takeru Ohta, and Masanori Koyama.
\newblock Optuna: A next-generation hyperparameter optimization framework,
  2019.

\bibitem[Bachlechner et~al.(2020)Bachlechner, Majumder, Mao, Cottrell, and
  McAuley]{bachlechner_rezero_2020}
Thomas Bachlechner, Bodhisattwa~Prasad Majumder, Huanru~Henry Mao, Garrison~W.
  Cottrell, and Julian McAuley.
\newblock {ReZero} is {All} {You} {Need}: {Fast} {Convergence} at {Large}
  {Depth}.
\newblock \emph{arXiv:2003.04887 [cs, stat]}, June 2020.
\newblock URL \url{http://arxiv.org/abs/2003.04887}.

\bibitem[Bernstein et~al.(2021)Bernstein, Vahdat, Yue, and
  Liu]{bernstein_distance_2021}
Jeremy Bernstein, Arash Vahdat, Yisong Yue, and Ming-Yu Liu.
\newblock On the distance between two neural networks and the stability of
  learning.
\newblock \emph{arXiv:2002.03432 [cs, math, stat]}, January 2021.
\newblock URL \url{http://arxiv.org/abs/2002.03432}.

\bibitem[Brown et~al.(2020)Brown, Mann, Ryder, Subbiah, Kaplan, Dhariwal,
  Neelakantan, Shyam, Sastry, Askell, Agarwal, Herbert-Voss, Krueger, Henighan,
  Child, Ramesh, Ziegler, Wu, Winter, Hesse, Chen, Sigler, Litwin, Gray, Chess,
  Clark, Berner, McCandlish, Radford, Sutskever, and Amodei]{brown2020language}
Tom~B. Brown, Benjamin Mann, Nick Ryder, Melanie Subbiah, Jared Kaplan,
  Prafulla Dhariwal, Arvind Neelakantan, Pranav Shyam, Girish Sastry, Amanda
  Askell, Sandhini Agarwal, Ariel Herbert-Voss, Gretchen Krueger, Tom Henighan,
  Rewon Child, Aditya Ramesh, Daniel~M. Ziegler, Jeffrey Wu, Clemens Winter,
  Christopher Hesse, Mark Chen, Eric Sigler, Mateusz Litwin, Scott Gray,
  Benjamin Chess, Jack Clark, Christopher Berner, Sam McCandlish, Alec Radford,
  Ilya Sutskever, and Dario Amodei.
\newblock Language models are few-shot learners, 2020.

\bibitem[Carbonnelle and De~Vleeschouwer(2019)]{carbonnelle_layer_2019}
Simon Carbonnelle and Christophe De~Vleeschouwer.
\newblock Layer rotation: a surprisingly powerful indicator of generalization
  in deep networks?
\newblock \emph{arXiv:1806.01603 [cs, stat]}, July 2019.
\newblock URL \url{http://arxiv.org/abs/1806.01603}.

\bibitem[Chelba et~al.(2014)Chelba, Mikolov, Schuster, Ge, Brants, Koehn, and
  Robinson]{1billionword}
Ciprian Chelba, Tomas Mikolov, Mike Schuster, Qi~Ge, Thorsten Brants, Phillipp
  Koehn, and Tony Robinson.
\newblock One billion word benchmark for measuring progress in statistical
  language modeling, 2014.

\bibitem[Dai et~al.(2019)Dai, Yang, Yang, Carbonell, Le, and
  Salakhutdinov]{dai2019transformerxl}
Zihang Dai, Zhilin Yang, Yiming Yang, Jaime Carbonell, Quoc~V. Le, and Ruslan
  Salakhutdinov.
\newblock Transformer-xl: Attentive language models beyond a fixed-length
  context, 2019.

\bibitem[Devlin et~al.(2019)Devlin, Chang, Lee, and
  Toutanova]{devlin_bert_2019}
Jacob Devlin, Ming-Wei Chang, Kenton Lee, and Kristina Toutanova.
\newblock {BERT}: {Pre}-training of {Deep} {Bidirectional} {Transformers} for
  {Language} {Understanding}.
\newblock \emph{arXiv:1810.04805 [cs]}, May 2019.
\newblock URL \url{http://arxiv.org/abs/1810.04805}.

\bibitem[Ding et~al.(2021)Ding, Xia, Zhang, Chu, Han, and
  Ding]{ding_repmlp_2021}
Xiaohan Ding, Chunlong Xia, Xiangyu Zhang, Xiaojie Chu, Jungong Han, and
  Guiguang Ding.
\newblock {RepMLP}: {Re}-parameterizing {Convolutions} into {Fully}-connected
  {Layers} for {Image} {Recognition}.
\newblock \emph{arXiv:2105.01883 [cs]}, August 2021.
\newblock URL \url{http://arxiv.org/abs/2105.01883}.

\bibitem[Glorot and Bengio(2010)]{glorot_understanding_2010}
Xavier Glorot and Yoshua Bengio.
\newblock Understanding the difficulty of training deep feedforward neural
  networks.
\newblock In Yee~Whye Teh and Mike Titterington, editors, \emph{Proceedings of
  the {Thirteenth} {International} {Conference} on {Artificial} {Intelligence}
  and {Statistics}}, volume~9 of \emph{Proceedings of {Machine} {Learning}
  {Research}}, pages 249--256, Chia Laguna Resort, Sardinia, Italy, May 2010.
  PMLR.
\newblock URL \url{http://proceedings.mlr.press/v9/glorot10a.html}.

\bibitem[Hoffer et~al.(2017)Hoffer, Hubara, and Soudry]{hoffer_train_2017}
Elad Hoffer, Itay Hubara, and Daniel Soudry.
\newblock Train longer, generalize better: closing the generalization gap in
  large batch training of neural networks.
\newblock \emph{arXiv:1705.08741 [cs, stat]}, May 2017.
\newblock URL \url{http://arxiv.org/abs/1705.08741}.

\bibitem[Horv{\'{a}}th et~al.(2021)Horv{\'{a}}th, Klein, Richt{\'{a}}rik, and
  Archambeau]{hptransferadaptive}
Samuel Horv{\'{a}}th, Aaron Klein, Peter Richt{\'{a}}rik, and C{\'{e}}dric
  Archambeau.
\newblock Hyperparameter transfer learning with adaptive complexity.
\newblock \emph{CoRR}, abs/2102.12810, 2021.
\newblock URL \url{https://arxiv.org/abs/2102.12810}.

\bibitem[Huang and Pérez()]{huang_improving_nodate}
Xiao~Shi Huang and Felipe Pérez.
\newblock Improving {Transformer} {Optimization} {Through} {Better}
  {Initialization}.
\newblock page~9.

\bibitem[Jacot et~al.(2018)Jacot, Gabriel, and Hongler]{jacot_neural_2018}
Arthur Jacot, Franck Gabriel, and Clément Hongler.
\newblock Neural {Tangent} {Kernel}: {Convergence} and {Generalization} in
  {Neural} {Networks}.
\newblock \emph{arXiv:1806.07572 [cs, math, stat]}, June 2018.
\newblock URL \url{http://arxiv.org/abs/1806.07572}.

\bibitem[Jamieson and Talwalkar(2015)]{jamieson2015nonstochastic}
Kevin Jamieson and Ameet Talwalkar.
\newblock Non-stochastic best arm identification and hyperparameter
  optimization, 2015.

\bibitem[Kaplan et~al.(2020)Kaplan, McCandlish, Henighan, Brown, Chess, Child,
  Gray, Radford, Wu, and Amodei]{kaplan_scaling_2020}
Jared Kaplan, Sam McCandlish, Tom Henighan, Tom~B. Brown, Benjamin Chess, Rewon
  Child, Scott Gray, Alec Radford, Jeffrey Wu, and Dario Amodei.
\newblock Scaling {Laws} for {Neural} {Language} {Models}.
\newblock \emph{arXiv:2001.08361 [cs, stat]}, January 2020.
\newblock URL \url{http://arxiv.org/abs/2001.08361}.

\bibitem[Kingma and Ba(2014)]{kingma2014adam}
Diederik Kingma and Jimmy Ba.
\newblock Adam: A method for stochastic optimization.
\newblock \emph{arXiv preprint arXiv:1412.6980}, 2014.

\bibitem[Lee et~al.(2018)Lee, Bahri, Novak, Schoenholz, Pennington, and
  Sohl-dickstein]{lee_deep_2018}
Jaehoon Lee, Yasaman Bahri, Roman Novak, Sam Schoenholz, Jeffrey Pennington,
  and Jascha Sohl-dickstein.
\newblock Deep {Neural} {Networks} as {Gaussian} {Processes}.
\newblock In \emph{International {Conference} on {Learning} {Representations}},
  2018.
\newblock URL \url{https://openreview.net/forum?id=B1EA-M-0Z}.

\bibitem[Li et~al.(2020)Li, Jamieson, Rostamizadeh, Gonina, Hardt, Recht, and
  Talwalkar]{li2020massively}
Liam Li, Kevin Jamieson, Afshin Rostamizadeh, Ekaterina Gonina, Moritz Hardt,
  Benjamin Recht, and Ameet Talwalkar.
\newblock A system for massively parallel hyperparameter tuning, 2020.

\bibitem[Li et~al.()Li, Jamieson, DeSalvo, Rostamizadeh, and
  Talwalkar]{li_hyperband}
Lisha Li, Kevin Jamieson, Giulia DeSalvo, Afshin Rostamizadeh, and Ameet
  Talwalkar.
\newblock Hyperband: {A} {Novel} {Bandit}-{Based} {Approach} to
  {Hyperparameter} {Optimization}.
\newblock \emph{JMLR 18}, page~52.

\bibitem[Liu et~al.(2021{\natexlab{a}})Liu, Dai, So, and Le]{liu_pay_2021}
Hanxiao Liu, Zihang Dai, David~R. So, and Quoc~V. Le.
\newblock Pay {Attention} to {MLPs}.
\newblock \emph{arXiv:2105.08050 [cs]}, June 2021{\natexlab{a}}.
\newblock URL \url{http://arxiv.org/abs/2105.08050}.

\bibitem[Liu et~al.(2020{\natexlab{a}})Liu, Liu, Gao, Chen, and
  Han]{liu-etal-2020-understanding}
Liyuan Liu, Xiaodong Liu, Jianfeng Gao, Weizhu Chen, and Jiawei Han.
\newblock Understanding the difficulty of training transformers.
\newblock In \emph{Proceedings of the 2020 Conference on Empirical Methods in
  Natural Language Processing (EMNLP)}, pages 5747--5763, Online, November
  2020{\natexlab{a}}. Association for Computational Linguistics.
\newblock \doi{10.18653/v1/2020.emnlp-main.463}.
\newblock URL \url{https://www.aclweb.org/anthology/2020.emnlp-main.463}.

\bibitem[Liu et~al.(2020{\natexlab{b}})Liu, Liu, Gao, Chen, and
  Han]{liu_understanding_2020}
Liyuan Liu, Xiaodong Liu, Jianfeng Gao, Weizhu Chen, and Jiawei Han.
\newblock Understanding the {Difficulty} of {Training} {Transformers}.
\newblock \emph{arXiv:2004.08249 [cs, stat]}, September 2020{\natexlab{b}}.
\newblock URL \url{http://arxiv.org/abs/2004.08249}.

\bibitem[Liu et~al.(2019)Liu, He, Chen, and Gao]{liu2019mt-dnn}
Xiaodong Liu, Pengcheng He, Weizhu Chen, and Jianfeng Gao.
\newblock Multi-task deep neural networks for natural language understanding.
\newblock In \emph{Proceedings of the 57th Annual Meeting of the Association
  for Computational Linguistics}, pages 4487--4496, Florence, Italy, July 2019.
  Association for Computational Linguistics.
\newblock URL \url{https://www.aclweb.org/anthology/P19-1441}.

\bibitem[Liu et~al.(2021{\natexlab{b}})Liu, Bernstein, Meister, and
  Yue]{liu_learning_2021}
Yang Liu, Jeremy Bernstein, Markus Meister, and Yisong Yue.
\newblock Learning by {Turning}: {Neural} {Architecture} {Aware}
  {Optimisation}.
\newblock \emph{arXiv:2102.07227 [cs]}, September 2021{\natexlab{b}}.
\newblock URL \url{http://arxiv.org/abs/2102.07227}.

\bibitem[Matthews et~al.(2018)Matthews, Rowland, Hron, Turner, and
  Ghahramani]{matthews_gaussian_2018}
Alexander G. de~G. Matthews, Mark Rowland, Jiri Hron, Richard~E. Turner, and
  Zoubin Ghahramani.
\newblock Gaussian {Process} {Behaviour} in {Wide} {Deep} {Neural} {Networks}.
\newblock \emph{arXiv:1804.11271 [cs, stat]}, April 2018.
\newblock URL \url{http://arxiv.org/abs/1804.11271}.
\newblock arXiv: 1804.11271.

\bibitem[McCandlish et~al.(2018)McCandlish, Kaplan, Amodei, and
  Team]{mccandlish_empirical_2018}
Sam McCandlish, Jared Kaplan, Dario Amodei, and OpenAI~Dota Team.
\newblock An {Empirical} {Model} of {Large}-{Batch} {Training}.
\newblock \emph{arXiv:1812.06162 [cs, stat]}, December 2018.
\newblock URL \url{http://arxiv.org/abs/1812.06162}.

\bibitem[Melas-Kyriazi(2021)]{melas-kyriazi_you_2021}
Luke Melas-Kyriazi.
\newblock Do {You} {Even} {Need} {Attention}? {A} {Stack} of {Feed}-{Forward}
  {Layers} {Does} {Surprisingly} {Well} on {ImageNet}.
\newblock \emph{arXiv:2105.02723 [cs]}, May 2021.
\newblock URL \url{http://arxiv.org/abs/2105.02723}.

\bibitem[Merity et~al.(2016)Merity, Xiong, Bradbury, and Socher]{wikitext103}
Stephen Merity, Caiming Xiong, James Bradbury, and Richard Socher.
\newblock Pointer sentinel mixture models, 2016.

\bibitem[Ott et~al.(2019)Ott, Edunov, Baevski, Fan, Gross, Ng, Grangier, and
  Auli]{ott2019fairseq}
Myle Ott, Sergey Edunov, Alexei Baevski, Angela Fan, Sam Gross, Nathan Ng,
  David Grangier, and Michael Auli.
\newblock fairseq: A fast, extensible toolkit for sequence modeling, mit
  license.
\newblock In \emph{Proceedings of NAACL-HLT 2019: Demonstrations}, 2019.

\bibitem[Park et~al.(2019)Park, Sohl-Dickstein, Le, and
  Smith]{park_effect_2019}
Daniel~S. Park, Jascha Sohl-Dickstein, Quoc~V. Le, and Samuel~L. Smith.
\newblock The {Effect} of {Network} {Width} on {Stochastic} {Gradient}
  {Descent} and {Generalization}: an {Empirical} {Study}.
\newblock May 2019.
\newblock URL \url{https://arxiv.org/abs/1905.03776v1}.

\bibitem[Paszke et~al.(2019)Paszke, Gross, Massa, Lerer, Bradbury, Chanan,
  Killeen, Lin, Gimelshein, Antiga, Desmaison, Kopf, Yang, DeVito, Raison,
  Tejani, Chilamkurthy, Steiner, Fang, Bai, and Chintala]{pytorch}
Adam Paszke, Sam Gross, Francisco Massa, Adam Lerer, James Bradbury, Gregory
  Chanan, Trevor Killeen, Zeming Lin, Natalia Gimelshein, Luca Antiga, Alban
  Desmaison, Andreas Kopf, Edward Yang, Zachary DeVito, Martin Raison, Alykhan
  Tejani, Sasank Chilamkurthy, Benoit Steiner, Lu~Fang, Junjie Bai, and Soumith
  Chintala.
\newblock Pytorch: An imperative style, high-performance deep learning library,
  bsd-style license.
\newblock In H.~Wallach, H.~Larochelle, A.~Beygelzimer, F.~d\'Alch\'{e} Buc,
  E.~Fox, and R.~Garnett, editors, \emph{Advances in Neural Information
  Processing Systems 32}, pages 8024--8035. Curran Associates, Inc., 2019.
\newblock URL
  \url{http://papers.neurips.cc/paper/9015-pytorch-an-imperative-style-high-performance-deep-learning-library.pdf}.

\bibitem[Perrone et~al.()Perrone, Jenatton, Seeger, and
  Archambeau]{perrone_scalable}
Valerio Perrone, Rodolphe Jenatton, Matthias~W Seeger, and Cedric Archambeau.
\newblock Scalable {Hyperparameter} {Transfer} {Learning}.
\newblock \emph{NeurIPS 2018}, page~11.

\bibitem[Popel and Bojar(2018)]{popel_training_2018}
Martin Popel and Ondřej Bojar.
\newblock Training {Tips} for the {Transformer} {Model}.
\newblock \emph{The Prague Bulletin of Mathematical Linguistics}, 110\penalty0
  (1):\penalty0 43--70, April 2018.
\newblock ISSN 1804-0462.
\newblock \doi{10.2478/pralin-2018-0002}.
\newblock URL
  \url{http://content.sciendo.com/view/journals/pralin/110/1/article-p43.xml}.

\bibitem[Raffel et~al.(2020)Raffel, Shazeer, Roberts, Lee, Narang, Matena,
  Zhou, Li, and Liu]{raffel_exploring_2020}
Colin Raffel, Noam Shazeer, Adam Roberts, Katherine Lee, Sharan Narang, Michael
  Matena, Yanqi Zhou, Wei Li, and Peter~J. Liu.
\newblock Exploring the {Limits} of {Transfer} {Learning} with a {Unified}
  {Text}-to-{Text} {Transformer}.
\newblock \emph{arXiv:1910.10683 [cs, stat]}, July 2020.
\newblock URL \url{http://arxiv.org/abs/1910.10683}.

\bibitem[Rosenfeld et~al.(2019)Rosenfeld, Rosenfeld, Belinkov, and
  Shavit]{rosenfeld_constructive_2019}
Jonathan~S. Rosenfeld, Amir Rosenfeld, Yonatan Belinkov, and Nir Shavit.
\newblock A {Constructive} {Prediction} of the {Generalization} {Error}
  {Across} {Scales}.
\newblock \emph{arXiv:1909.12673 [cs, stat]}, December 2019.
\newblock URL \url{http://arxiv.org/abs/1909.12673}.

\bibitem[Schoenholz et~al.(2016)Schoenholz, Gilmer, Ganguli, and
  Sohl-Dickstein]{schoenholz_deep_2016}
Samuel~S. Schoenholz, Justin Gilmer, Surya Ganguli, and Jascha Sohl-Dickstein.
\newblock Deep {Information} {Propagation}.
\newblock \emph{arXiv:1611.01232 [cs, stat]}, November 2016.
\newblock URL \url{http://arxiv.org/abs/1611.01232}.

\bibitem[Shallue et~al.(2018)Shallue, Lee, Antognini, Sohl-Dickstein, Frostig,
  and Dahl]{shallue_measuring_2018}
Christopher~J. Shallue, Jaehoon Lee, Joseph Antognini, Jascha Sohl-Dickstein,
  Roy Frostig, and George~E. Dahl.
\newblock Measuring the {Effects} of {Data} {Parallelism} on {Neural} {Network}
  {Training}.
\newblock \emph{arXiv:1811.03600 [cs, stat]}, November 2018.
\newblock URL \url{http://arxiv.org/abs/1811.03600}.

\bibitem[Shazeer and Stern(2018)]{shazeer_adafactor_2018}
Noam Shazeer and Mitchell Stern.
\newblock Adafactor: {Adaptive} {Learning} {Rates} with {Sublinear} {Memory}
  {Cost}.
\newblock April 2018.
\newblock URL \url{https://arxiv.org/abs/1804.04235v1}.

\bibitem[Shoeybi et~al.(2019)Shoeybi, Patwary, Puri, LeGresley, Casper, and
  Catanzaro]{megatron}
Mohammad Shoeybi, Mostofa Patwary, Raul Puri, Patrick LeGresley, Jared Casper,
  and Bryan Catanzaro.
\newblock Megatron-lm: Training multi-billion parameter language models using
  model parallelism.
\newblock \emph{CoRR}, abs/1909.08053, 2019.
\newblock URL \url{http://arxiv.org/abs/1909.08053}.

\bibitem[Smith et~al.(2017)Smith, Kindermans, and Le]{smith_dont_2017}
Samuel~L. Smith, Pieter-Jan Kindermans, and Quoc~V. Le.
\newblock Don't {Decay} the {Learning} {Rate}, {Increase} the {Batch} {Size}.
\newblock \emph{arXiv:1711.00489 [cs, stat]}, November 2017.
\newblock URL \url{http://arxiv.org/abs/1711.00489}.

\bibitem[Snoek et~al.(2012)Snoek, Larochelle, and Adams]{snoek2012practical}
Jasper Snoek, Hugo Larochelle, and Ryan~P. Adams.
\newblock Practical bayesian optimization of machine learning algorithms, 2012.

\bibitem[Snoek et~al.(2015)Snoek, Rippel, Swersky, Kiros, Satish, Sundaram,
  Patwary, Prabhat, and Adams]{snoek2015scalable}
Jasper Snoek, Oren Rippel, Kevin Swersky, Ryan Kiros, Nadathur Satish,
  Narayanan Sundaram, Md. Mostofa~Ali Patwary, Prabhat, and Ryan~P. Adams.
\newblock Scalable bayesian optimization using deep neural networks, 2015.

\bibitem[Stoll et~al.(2021)Stoll, Franke, Wagner, Selg, and
  Hutter]{stoll2021hyperparameter}
Danny Stoll, J{\"o}rg~K.H. Franke, Diane Wagner, Simon Selg, and Frank Hutter.
\newblock Hyperparameter transfer across developer adjustments, 2021.
\newblock URL \url{https://openreview.net/forum?id=WPO0vDYLXem}.

\bibitem[Tolstikhin et~al.(2021)Tolstikhin, Houlsby, Kolesnikov, Beyer, Zhai,
  Unterthiner, Yung, Steiner, Keysers, Uszkoreit, Lucic, and
  Dosovitskiy]{tolstikhin_mlp-mixer_2021}
Ilya Tolstikhin, Neil Houlsby, Alexander Kolesnikov, Lucas Beyer, Xiaohua Zhai,
  Thomas Unterthiner, Jessica Yung, Andreas Steiner, Daniel Keysers, Jakob
  Uszkoreit, Mario Lucic, and Alexey Dosovitskiy.
\newblock {MLP}-{Mixer}: {An} all-{MLP} {Architecture} for {Vision}.
\newblock \emph{arXiv:2105.01601 [cs]}, June 2021.
\newblock URL \url{http://arxiv.org/abs/2105.01601}.

\bibitem[Touvron et~al.(2021)Touvron, Bojanowski, Caron, Cord, El-Nouby, Grave,
  Izacard, Joulin, Synnaeve, Verbeek, and Jégou]{touvron_resmlp_2021}
Hugo Touvron, Piotr Bojanowski, Mathilde Caron, Matthieu Cord, Alaaeldin
  El-Nouby, Edouard Grave, Gautier Izacard, Armand Joulin, Gabriel Synnaeve,
  Jakob Verbeek, and Hervé Jégou.
\newblock {ResMLP}: {Feedforward} networks for image classification with
  data-efficient training.
\newblock \emph{arXiv:2105.03404 [cs]}, June 2021.
\newblock URL \url{http://arxiv.org/abs/2105.03404}.

\bibitem[Vaswani et~al.(2017)Vaswani, Shazeer, Parmar, Uszkoreit, Jones, Gomez,
  Kaiser, and Polosukhin]{DBLP:journals/corr/VaswaniSPUJGKP17}
Ashish Vaswani, Noam Shazeer, Niki Parmar, Jakob Uszkoreit, Llion Jones,
  Aidan~N. Gomez, Lukasz Kaiser, and Illia Polosukhin.
\newblock Attention is all you need.
\newblock \emph{CoRR}, abs/1706.03762, 2017.
\newblock URL \url{http://arxiv.org/abs/1706.03762}.

\bibitem[Wang et~al.(2018)Wang, Singh, Michael, Hill, Levy, and
  Bowman]{wang2018glue}
Alex Wang, Amanpreet Singh, Julian Michael, Felix Hill, Omer Levy, and Samuel~R
  Bowman.
\newblock Glue: A multi-task benchmark and analysis platform for natural
  language understanding.
\newblock \emph{EMNLP 2018}, page 353, 2018.

\bibitem[Williams et~al.(2018)Williams, Nangia, and Bowman]{mnli2018}
Adina Williams, Nikita Nangia, and Samuel Bowman.
\newblock A broad-coverage challenge corpus for sentence understanding through
  inference.
\newblock In \emph{Proceedings of the 2018 Conference of the North American
  Chapter of the Association for Computational Linguistics: Human Language
  Technologies, Volume 1 (Long Papers)}, pages 1112--1122. Association for
  Computational Linguistics, 2018.
\newblock URL \url{http://aclweb.org/anthology/N18-1101}.

\bibitem[Yang(2019{\natexlab{a}})]{TP1}
Greg Yang.
\newblock Tensor {Programs} {I}: {Wide} {Feedforward} or {Recurrent} {Neural}
  {Networks} of {Any} {Architecture} are {Gaussian} {Processes}.
\newblock \emph{arXiv:1910.12478 [cond-mat, physics:math-ph]}, December
  2019{\natexlab{a}}.
\newblock URL \url{http://arxiv.org/abs/1910.12478}.

\bibitem[Yang(2019{\natexlab{b}})]{scaling}
Greg Yang.
\newblock Scaling {Limits} of {Wide} {Neural} {Networks} with {Weight}
  {Sharing}: {Gaussian} {Process} {Behavior}, {Gradient} {Independence}, and
  {Neural} {Tangent} {Kernel} {Derivation}.
\newblock \emph{arXiv:1902.04760 [cond-mat, physics:math-ph, stat]}, February
  2019{\natexlab{b}}.
\newblock URL \url{http://arxiv.org/abs/1902.04760}.

\bibitem[Yang(2020{\natexlab{a}})]{TP2}
Greg Yang.
\newblock Tensor {Programs} {II}: {Neural} {Tangent} {Kernel} for {Any}
  {Architecture}.
\newblock \emph{arXiv:2006.14548 [cond-mat, stat]}, August 2020{\natexlab{a}}.
\newblock URL \url{http://arxiv.org/abs/2006.14548}.

\bibitem[Yang(2020{\natexlab{b}})]{TP3}
Greg Yang.
\newblock Tensor {Programs} {III}: {Neural} {Matrix} {Laws}.
\newblock \emph{arXiv:2009.10685 [cs, math]}, September 2020{\natexlab{b}}.
\newblock URL \url{http://arxiv.org/abs/2009.10685}.

\bibitem[Yang and Hu(2020)]{TP4}
Greg Yang and Edward~J. Hu.
\newblock Feature learning in infinite-width neural networks.
\newblock \emph{arXiv}, 2020.

\bibitem[Yang and Littwin(2021)]{TP2b}
Greg Yang and Etai Littwin.
\newblock Tensor {Programs} {IIb}: {Architectural} {Universality} of {Neural}
  {Tangent} {Kernel} {Training} {Dynamics}.
\newblock \emph{arXiv:2105.03703 [cs, math]}, May 2021.
\newblock URL \url{http://arxiv.org/abs/2105.03703}.

\bibitem[Yang and Schoenholz(2018)]{yang_layerwise_2018}
Greg Yang and Sam~S. Schoenholz.
\newblock Deep {Mean} {Field} {Theory}: {Layerwise} {Variance} and {Width}
  {Variation} as {Methods} to {Control} {Gradient} {Explosion}.
\newblock February 2018.
\newblock URL \url{https://openreview.net/forum?id=rJGY8GbR-}.

\bibitem[Yang and Schoenholz(2017)]{yang_mean_2017}
Greg Yang and Samuel~S. Schoenholz.
\newblock Mean {Field} {Residual} {Networks}: {On} the {Edge} of {Chaos}.
\newblock \emph{arXiv:1712.08969 [cond-mat, physics:nlin]}, December 2017.
\newblock URL \url{http://arxiv.org/abs/1712.08969}.

\bibitem[Yang et~al.(2022)Yang, Santacroce, and Hu]{yang2022efficient}
Greg Yang, Michael Santacroce, and Edward~J Hu.
\newblock Efficient computation of deep nonlinear infinite-width neural
  networks that learn features.
\newblock In \emph{International Conference on Learning Representations}, 2022.
\newblock URL \url{https://openreview.net/forum?id=tUMr0Iox8XW}.

\bibitem[Yogatama and Mann(2014)]{yogatama_efficient_2014}
Dani Yogatama and Gideon Mann.
\newblock Efficient {Transfer} {Learning} {Method} for {Automatic}
  {Hyperparameter} {Tuning}.
\newblock In \emph{Artificial {Intelligence} and {Statistics}}, pages
  1077--1085. PMLR, April 2014.
\newblock URL \url{http://proceedings.mlr.press/v33/yogatama14.html}.

\bibitem[You et~al.(2017)You, Gitman, and Ginsburg]{you_large_2017}
Yang You, Igor Gitman, and Boris Ginsburg.
\newblock Large {Batch} {Training} of {Convolutional} {Networks}.
\newblock \emph{arXiv:1708.03888 [cs]}, September 2017.
\newblock URL \url{http://arxiv.org/abs/1708.03888}.

\bibitem[You et~al.(2020)You, Li, Reddi, Hseu, Kumar, Bhojanapalli, Song,
  Demmel, Keutzer, and Hsieh]{you_large_2020}
Yang You, Jing Li, Sashank Reddi, Jonathan Hseu, Sanjiv Kumar, Srinadh
  Bhojanapalli, Xiaodan Song, James Demmel, Kurt Keutzer, and Cho-Jui Hsieh.
\newblock Large {Batch} {Optimization} for {Deep} {Learning}: {Training} {BERT}
  in 76 minutes.
\newblock \emph{arXiv:1904.00962 [cs, stat]}, January 2020.
\newblock URL \url{http://arxiv.org/abs/1904.00962}.

\bibitem[Zagoruyko and Komodakis(2017)]{zagoruyko2017wide}
Sergey Zagoruyko and Nikos Komodakis.
\newblock Wide residual networks, 2017.

\bibitem[Zhang et~al.(2019)Zhang, Dauphin, and Ma]{zhang_residual_2019}
Hongyi Zhang, Yann~N. Dauphin, and Tengyu Ma.
\newblock Residual {Learning} {Without} {Normalization} via {Better}
  {Initialization}.
\newblock In \emph{International {Conference} on {Learning} {Representations}},
  2019.
\newblock URL \url{https://openreview.net/forum?id=H1gsz30cKX}.

\end{thebibliography}
\bibliographystyle{plainnat}

\newpage
\appendix

\pdfbookmark[section]{Table of Contents}{Contents}
\tableofcontents

\pdfbookmark[section]{List of Figures}{Figures}
\listoffigures

\pdfbookmark[section]{List of Tables}{Tables}
\listoftables

\newpage

\section{Parametrization Terminologies}
\label{sec:paramterminology}

This section seeks to make formal and clarify some of the notions regarding parametrization discussed informally in the main text.

\begin{defn}[Multiplier and Parameter Multiplier]
  In a neural network, one may insert a ``multiply by $c$'' operation anywhere, where $c$ is a non-learnable scalar hyperparameter.
  If $c=1$, then this operation is a no-op.
  This $c$ is called a \emph{multiplier}.

  Relatedly, for any parameter tensor $W$ in a neural network, we may replace $W$ with $cW$ for some non-learnable scalar hyperparameter $c$.
  When $c=1$, we recover the original formulation.
  This $c$ is referred to as a \emph{parameter multiplier}.
\end{defn}
For example, in the attention logit calculation $\langle k, q\rangle/\sqrt{d_{head}}$ where $q = W x$, the $1/\sqrt{d_{head}}$ factor is a multiplier.
It may also be thought of as the parameter multiplier of $W$ if we rewrite the attention logit as $ \langle k, (W/\sqrt{d_{head}}) x\rangle$.

Note parameter multipliers cannot be absorbed into the initialization in general, since they affect backpropagation.
Nevertheless, after training is done, parameter multipliers can always be absorbed into the weight.

\begin{defn}[Parametrization]
  In this work, a \emph{parametrization} is a rule for how to \emph{change} hyperparameters when the widths of a neural network \emph{change}, but note that it does not necessarily prescribes how to set the hyperparameters for any specific width.
  In particular, for any neural network, an \emph{abc-parametrization} is a rule for how to scale a) the parameter multiplier, b) the initialization, and c) the learning rate individually for each parameter tensor as the widths of the network change, as well as any other multiplier in the network; all other hyperparameters are kept fixed with width.
\end{defn}
For example, SP and $\mu$P are both abc-parametrizations.
Again, we note that, in this sense, a parametrization does not prescribe, for example, that the initialization variance be $1/\fanin$, but rather that it be halved when $\fanin$ doubles.

\begin{defn}[Zero-Shot Hyperparameter Transfer]
  In this work, we say a parametrization admits \emph{zero-shot transfer of a set of hyperparameters $\mathcal H$ w.r.t.\ a metric $\mathcal L$} if the optimal combination of values of $\mathcal H$ w.r.t.\ $\mathcal L$ converges as width goes to infinity, i.e. it stays approximately optimal w.r.t.\ $\mathcal L$ under this parametrization as width increases.
\end{defn}
Throughout this paper, we take $\mathcal L$ to be the training loss, but because regularization is not the bottleneck in our experiments (especially large scale pretraining with BERT and GPT-3), we nevertheless see high quality test performance in all of our results.
We also remark that empirically, using training loss as the metric can be more robust to random seed compared to validation loss and especially BLEU score.
See \cref{tab:whichtransfer}(left) for $\mathcal H$.
By our arguments in \cref{{sec:otherparamdontwork}} and our empirical results, $\mu$P is the unique abc-parametrization admitting zero-shot transfer for such $\mathcal H$ and $\mathcal L$ in this sense.

More generally, one may define a \emph{$K$-shot transfer algorithm of a set of hyperparameters $\mathcal H$ w.r.t.\ a metric $\mathcal L$} as one that 1) takes width values $n$ and $n'$ and an approximately optimal combination of values of $\mathcal H$ w.r.t.\ $\mathcal L$ at a width $n$ and 2) returns an approximately optimal combination of values of $\mathcal H$ w.r.t.\ $\mathcal L$ at width $n'$, given 3) a budget of $K$ evaluations of candidate hyperparameter combinations on models of width $n'$.
However, we will have no use for this definition in this paper.

\section{Further Explanations of the \texorpdfstring{$\mu$P}{muP} Tables}
\label{sec:furtherexplain}

\begin{table}[t]
  \renewcommand{\hl}[1]{\textcolor{purple}{#1}}
  \renewcommand{\ll}[1]{\textcolor{gray}{#1}}
  \centering
  \caption[Alternative (Equivalent) $\mu$P Formulation for Easier Implementation]{
    \textbf{Alternative (Equivalent) $\mu$P Formulation for Easier Implementation.}
  Same format as in \cref{tab:MUP}.
  In contrast to the formulation in \cref{tab:MUP}, here all ``vector-like'' parameters (i.e.\ those that have only one dimension tending to infinity), including input and output weights and biases, have the same width scaling for initialization variance and SGD/Adam LR (note the $1/\fanin{}$ for input weight/bias init.\ var.\ is $\Theta(1)$ in width).
  This has two benefits in practice: 1) implementation is unified and simplified for all ``vector-like'' parameters; 2) input and output weights can now be tied, in contrast to \cref{tab:MUP}, which is a common design feature of Transformer models.
  Note that in this table, for biases, the $\fanin$ is 1 (compare to PyTorch \texttt{nn.Linear} default initialization of biases, where $\fanin$ refers to $\fanin$ of the layer.)
  This table can be derived from \cref{tab:MUP} via \cref{lem:scalingEquivalences}.
  See \cref{sec:furtherexplain} for further explanations.}
  \begin{tabular}{lllllllll}
      \toprule
      &   \multicolumn{2}{c}{Input weights \& all biases} &  & \multicolumn{2}{c}{Output weights}  &   & \multicolumn{2}{c}{Hidden weights} 
      \\
      \midrule
      Init.\ Var.  & \multicolumn{2}{c}{$\nicefrac{1}{\fanin}$}   &    & \hl{1} & \ll{($\nicefrac{1}{\fanin}$)}          &       & \multicolumn{2}{c}{$\nicefrac{1}{\fanin}$}                
      \\
      Multiplier & \multicolumn{2}{c}{1} & & \hl{$\nicefrac 1 {\fanin}$}  & \ll{($1$)}&   & \multicolumn{2}{c}{1}    %
      \\
      SGD LR      & \hl{$\fanout$} & \ll{($1$)}  &  & \hl{$\fanin$} & \ll{($1$)}   & & \multicolumn{2}{c}{$1$}               
      \\
      Adam LR             & \multicolumn{2}{c}{$1$}             &                 & \multicolumn{2}{c}{$1$}           &                    & \hl{$\nicefrac 1\fanin$} & \ll{($1$)}              
      \\
      \bottomrule
  \end{tabular}
  \label{tab:MUPalt}
\end{table}

\begin{table}[t]
  \renewcommand{\hl}[1]{\textcolor{purple}{#1}}
  \renewcommand{\ll}[1]{\textcolor{gray}{#1}}
  \centering
  \caption[$\mu$P Formulation in the Style of \citep{TP4}]{\textbf{$\mu$P Formulation in the Style of \citep{TP4}.}
  This table can be derived from \cref{tab:MUP} via \cref{lem:scalingEquivalences}.}
  \begin{tabular}{lllllllll}
      \toprule
      &   \multicolumn{2}{c}{Input weights \& all biases} &  & \multicolumn{2}{c}{Output weights}  &   & \multicolumn{2}{c}{Hidden weights} 
      \\
      \midrule
      Init.\ Var.  & \hl{$\nicefrac{1}{\fanout}$} & \ll{$(\nicefrac{1}{\fanin})$}   &    & \multicolumn{2}{c}{$\nicefrac{1}{\fanin}$}          &       & \multicolumn{2}{c}{$\nicefrac{1}{\fanin}$}                
      \\
      Multiplier & \hl{$\sqrt{\fanout}$} & \ll{($1$)} & & \hl{$\nicefrac 1 {\sqrt{\fanin}}$}  & \ll{($1$)}&   & \multicolumn{2}{c}{1}    %
      \\
      SGD LR      & \multicolumn{2}{c}{$1$}  & & \multicolumn{2}{c}{$1$}   & & \multicolumn{2}{c}{$1$}               
      \\
      Adam LR             & \hl{$\nicefrac 1 {\sqrt{\fanout}}$} &  \ll{($1$)}             &                 & \hl{$\nicefrac 1 {\sqrt{\fanin}}$} & \ll{($1$)}           &                    & \hl{$\nicefrac 1 \fanin$} & \ll{($1$)}              
      \\
      \bottomrule
  \end{tabular}
  \label{tab:MUPorig}
\end{table}

In addition to \cref{tab:MUP}, we provide \cref{tab:MUPalt} as an equivalent $\mu$P formulation that is easier to implement, as well as \cref{tab:MUPorig} for those more familiar with the original $\mu$P formulation in \citep{TP4}.
Below, we provide some commentary on corner cases not well specified by the tables.
Ultimately, by understanding \cref{sec:intuitiveintro2muP}, one can derive $\mu$P for any architecture, new or old.

\paragraph{Matrix-Like, Vector-Like, Scalar-Like Parameters}
We can classify any dimension in a neural network as ``infinite'' if it scales with width, or ``finite'' otherwise.
For example, in a Transformer, $d_{model}, d_{ffn}, d_{head}, n_{head}$ are all infinite, but vocab size and context size are finite.
Then we can categorize parameter tensors by how many infinite dimensions they have.
If there are two such dimensions, then we say the parameter is \emph{matrix-like}; if there is only one, then we say it is \emph{vector-like}; if there is none, we say it is \emph{scalar-like}.
Then in \cref{tab:MUP,tab:MUPalt,tab:MUPorig}, ``input weights \& all biases'' and ``output weights'' are all vector-like parameters, while hidden weights are matrix-like parameters.
An advantage of \cref{tab:MUPalt} is that it gives a uniform scaling rule of initialization and learning rate for all vector-like parameters.
The multiplier rule in \cref{tab:MUPalt} can be more interpreted more generally as the following: a multiplier of order $1/\fanin$ should accompany any weight that maps an infinite dimension to a finite one.
This interpretation then nicely covers both the output logits and the attention logits (i.e.\ $1/d$ attention).

Scalar-like parameters are not as common as matrix-like and vector-like ones, but we will mention a few examples in \cref{sec:otherparams}.
The scaling rule for their initialization, learning rate (for both SGD and Adam), and multiplier is very simple: hold them constant with width.

\paragraph{Initialization Mean}
We did not specify the initialization mean in the tables, since most commonly the mean is just set to 0,
but it can be nonzero for vector-like parameters (e.g., layernorm weights) and scalar-like parameters but must be 0 for matrix-like parameters.

\paragraph{Zero Initialization Variance}
The initialization scaling rules in our tables can all be trivially satisfied if the initialization variance is set to 0.
This can be useful in some settings (e.g., \cref{sec:outputzeroinit}) but detrimental in other settings (e.g., hidden weights).

\paragraph{What Are Considered Input Weights? Output Weights?}
Here, input weights very specifically refer to weights that map from an infinite dimension to a finite dimension.
As a counterexample, in some architectures, the first layer can actually map from a finite dimension to another finite dimension, e.g., a PCA layer.
Then this is not an ``input weight''; if the next layer maps into an infinite dimension, then that's the input weight.
A similar, symmetric discussion applies to output weights.

\paragraph{What Counts As a ``Model''? Does the MLP in a Transformer Count As a ``Model''?}
For our tables, a model is specifically a function that maps a finite dimension to another finite dimension, consistent with the discussion above.
For example, for an image model on CIFAR10, it maps from $3 \times 32 \times 32 = 3072$ dimensions to 10 dimensions, and these numbers are fixed regardless of the width of the model.
Likewise, for an autoregressive Transformer model, the input and output dimension are both the vocab size, which is independent of the width.
In contrast, an MLP inside a Transformer is not a ``model'' in this sense because its input and output dimension are both equal to the width of the Transformer.

\subsection{Walkthrough of \texorpdfstring{$\mu$P}{muP} Implementation in a Transformer}

To ground the abstract description in \cref{tab:MUP,tab:MUPalt,tab:MUPorig}, we walk through the parameters of a typical Transformer and discuss concretely how to parametrize each.

We assume that the user wants to replicate SP when the model widths are equal to some base widths, for example, when $d_{model} = d_{model,0} = 128, d_{ffn} = d_{ffn,0} = 512$, etc, as in the MLP example in \cref{sec:mutransfer}.
For this purpose, it's useful to define $\tilde d_{model} = d_{model} / d_{model,0},\ \tilde d_{ffn} = d_{ffn} / d_{ffn,0}$, and so on.
One can always take $d_{model,0} = d_{ffn,0} = \cdots = 1$ for a ``pure'' $\mu$P.

Below, we introduce hyperparameters $\sigma_\bullet, \eta_\bullet$ for each parameter tensor, as well as a few multipliers $\alpha_\bullet$.
One may always tie $\sigma_\bullet$ (resp.\ $\eta_\bullet$) across all parameter tensors, but in our experiments, we found it beneficial to at least distinguish the input and output layer initialization and learning rates.

\paragraph{Input Word Embeddings}
The input word embedding matrix $W^{wordemb}$ has size $d_{model} \times vocabsize$, where $vocabsize$ is the fan-in and $d_{model}$ is the fan-out.
Follow the ``input weight \& all biases'' column in \cref{tab:MUP,tab:MUPalt,tab:MUPorig}.
For example, for \cref{tab:MUP,tab:MUPalt},
\[
  W^{wordemb} \sim \Gaus(0, \sigma^2_{wordemb}),\quad\text{with Adam LR }\eta_{wordemb}\]
Note here, because fan-in ($vocabsize$) here is independent of width ($d_{model}$), the ``$1/\fanin$'' for the initialization variance in these tables is equivalent to ``$1$'', i.e. the initialization variance can be anything fixed with width.
In this case of the word embedding, setting the variance to 1, for example, is more natural than setting the variance to $1/\fanin$, because the embedding is one-hot ($1/\fanin$ would be more natural for image inputs).

\paragraph{Positional Embeddings}
The (absolute or relative) positional embedding matrix $W^{posemb}$ has size $d_{model} \times contextsize$, where $contextsize$ is the fan-in and $d_{model}$ is the fan-out.
With the same discussion as above for input word embeddings, follow the ``input weight \& all biases'' column in \cref{tab:MUP,tab:MUPalt,tab:MUPorig}.
For example, for \cref{tab:MUP,tab:MUPalt},
\[
  W^{posemb} \sim \Gaus(0, \sigma^2_{posemb}),\quad\text{with Adam LR }\eta_{posemb}\]

\paragraph{Layernorm Weights and Biases}
Layernorm weights $w^{LN}$ and biases $b^{LN}$ both have shape $d_{model}$ and can be thought of ``input weights'' to the scalar input of 1.
Hence one should follow the ``input weight \& all biases'' column in \cref{tab:MUP,tab:MUPalt,tab:MUPorig}.
In particular, the usual initialization of layernorm weights as all 1s and biases as all 0s suffice (where the initialization variance is 0).
For example, for \cref{tab:MUP,tab:MUPalt},
\[
  w^{LN} \gets 1,\quad\text{with Adam LR }\eta_{LNw},\qquad \text{and} \qquad
  b^{LN} \gets 0,\quad\text{with Adam LR }\eta_{LNb}\]

\paragraph{Self-Attention}
There are 4 matrices, $W^q, W^k \in \R^{(d_k n_{head}) \times d_{model}}$, $W^v \in \R^{(d_v n_{head}) \times d_{model}}$, and $W^o \in\R^{d_{model} \times (d_v n_{head})}$ (where the shapes are $\R^{\fanout \times \fanin}$).
Since $d_{model}$, $(d_k n_{head})$, and $(d_v n_{head})$ all scale with width (where the latter two are commonly just set to $d_{model}$), all 4 matrices should be parametrized according to the ``hidden weights'' column in \cref{tab:MUP,tab:MUPalt,tab:MUPorig}.
For example, for \cref{tab:MUP,tab:MUPalt},
\begin{align*}
  W^q &\sim \Gaus(0, \sigma_q^2/d_{model}),&&\text{with Adam LR }\eta_{q}/\tilde d_{model}\\
  W^k &\sim \Gaus(0, \sigma_k^2/d_{model}),&&\text{with Adam LR }\eta_{k}/\tilde d_{model}\\
  W^v &\sim \Gaus(0, \sigma_v^2/d_{model}),&&\text{with Adam LR }\eta_{v}/\tilde d_{model}\\
  W^o &\sim \Gaus(0, \sigma_o^2/(d_v n_{head})),&&\text{with Adam LR }\eta_{o}/(\tilde d_v \tilde n_{head}).
\end{align*}

\paragraph{Attention Logit Scaling}

We use $1/d$ instead of $1/\sqrt{d}$ attention.
To be compatible with $1/\sqrt{d}$ attention when at a particular base $d_{head} = d_{head,0}$, we set
\[AttnLogit = \alpha_{attn}\frac{\sqrt{d_{head,0}}}{d_{head}} q^\trsp k,\]
where $\alpha_{attn}$ is a tunable multiplier.

\paragraph{MLP}
There are 2 matrices, $W^1 \in \R^{d_{ffn}\times d_{model}}, W^2 \in \R^{d_{model} \times d_{ffn}}$ (where the shapes are $\R^{\fanout \times \fanin}$), where $d_{ffn}$ is commonly set to $4 d_{model}$.
Since both $d_{model}, d_{ffn}$ scale with width, both matrices are considered ``hidden weights.''
For example, for \cref{tab:MUP,tab:MUPalt},
\begin{align*}
  W^1 &\sim \Gaus(0, \sigma_q^2/d_{model}),&&\text{with Adam LR }\eta_{q}/\tilde d_{model}\\
  W^2 &\sim \Gaus(0, \sigma_k^2/d_{ffn}),&&\text{with Adam LR }\eta_{k}/\tilde d_{ffn}
\end{align*}

\paragraph{Word Unembeddings}
Symmetric to the discussion on input word embeddings, the output word unembeddings should be parametrized according to the ``output weights'' column of \cref{tab:MUP,tab:MUPalt,tab:MUPorig}.
Often, the unembeddings are tied with the embeddings, and \cref{tab:MUPalt,tab:MUPorig} allow for this as their initialization schemes are symmetric between input and output weights.

For example, for \cref{tab:MUP}, we'd set
\[
  W^{unemb} \sim \Gaus(0, \sigma^2_{unemb}/(d_{model} \tilde d_{model})),\quad\text{with Adam LR }\eta_{unemb}/\tilde d_{model}.
\]

For \cref{tab:MUPalt}, we would instead have
\[
  W^{unemb} \sim \Gaus(0, \sigma^2_{unemb} / d_{model,0}),\quad\text{with Adam LR }\eta_{unemb},
\]
(note $d_{model,0}$ here is the base width and therefore is a constant)
and the output is computed as
\[
  logits = \frac{\alpha_{output}}{\tilde d_{model}} W^{unemb} z\]
where $z$ is the final layer embedding of a token, and $\alpha_{output}$ is a tunable multiplier.

\subsection{Other Parameters}
\label{sec:otherparams}

\paragraph{Learnable scalar multipliers}
For learnable scalar multipliers (e.g., softmax inverse temperature), one can initialize them to 1 and use a constant (in width) learning rate for both SGD and Adam.
This is compatible with \cref{tab:MUP,tab:MUPalt,tab:MUPorig}.

\paragraph{Positional Bias}
Some Transformers use positional bias (of size $contextsize \times contextsize$, which are added to the attention logits).
They are considered ``scalar-like'' in that it has no width dimension.
One can initialize them to 0 and use a constant (in width) learning rate for both SGD and Adam.
This is compatible with \cref{tab:MUP,tab:MUPalt,tab:MUPorig}.

\paragraph{Spatial MLPs}
Recent works \citep{liu_pay_2021,tolstikhin_mlp-mixer_2021,melas-kyriazi_you_2021,touvron_resmlp_2021,ding_repmlp_2021} on MLP-only architectures in NLP and CV replace the self-attention layer in Transformers with MLPs across tokens or spatial locations.
In our language here, such MLPs have finite input and output dimensions (the context size) and infinite hidden dimensions, so their input, output, and hidden weights should be parametrized via the corresponding columns in \cref{tab:MUP,tab:MUPalt,tab:MUPorig}.

\subsection{Optimizer Variants and Hyperparameters}

\label{sec:otheroptimizers}
\paragraph{AdamW}
Exactly the same as Adam in all of our tables, with the added benefit that weight decay is automatically scaled correctly in AdamW (but is incompatible with $\mu$P Adam).
For this reason, we recommend using AdamW when weight decay is desired (which is consistent with current standard practice).

\paragraph{Frobenius Normalization}
LARS \citep{you_large_2017}, Adafactor \citep{shazeer_adafactor_2018}, Lamb \citep{you_large_2020}, Layca \citep{carbonnelle_layer_2019}, Fromage \citep{bernstein_distance_2021}, Nero \citep{liu_learning_2021}
all involve a normalization step in which the update $g$ (which may be obtained from SGD, Adam, or other optimzers) is normalized to have Frobenius norm equal to that of the parameter $w$: $g \gets \frac{\|w\|_F}{\|g\|_F} g$.
They can be made compatible with $\mu$P in \cref{tab:MUPalt} by scaling their learning rate for hidden weights like $1/\sqrt{\fanin}$ (for \cref{tab:MUP}, the output weight learning rate should be likewise scaled).
The intuitive reasoning (which can be formalized straightforwardly using Tensor Programs) is as follows.

This normalization implicitly encodes a width scaling:
If one initializes a weight matrix with variance $1/\fanin{}$, then an $n \times n$ matrix (e.g., a hidden weight matrix) has Frobenius norm $\sqrt{n}$ at initialization.
Thus, in the first step and, by induction, in any step $t$, the normalized update to this $n \times n$ weight also has Frobenius norm $\Theta(\sqrt n)$ (for any fixed $t$, as $n \to \infty$).
Heuristically, this means each entry of $g$ is approximately of size $\Theta(1/\sqrt n)$.
But, by the derivation of \cref{{sec:intuitiveintro2muP}}, we want $\Theta(1/n)$ and this is $\Theta(\sqrt n)$ too large!
Thus, in wide enough networks, one should see a network blowup after one update, like demonstrated in \cref{fig:blowupafter1step}.

However, note that the $\Theta(1/\sqrt n)$ coordinate size induced by the normalization here is closer to the right size $\Theta(1/n)$ than Adam, whose update have coordinate size $\Theta(1)$.
This may partially explain the apparent benefit of these optimizers.
In particular, this may explain the observation that T5 \citep{raffel_exploring_2020}, using Adafactor, was able to train its entire range of models from 220 million to 11 billion parameters with a fixed set of hyperparameters, while GPT-3 \citep{brown2020language}, using Adam, needed to decrease its learning rate with model size.

\paragraph{RAdam}
RAdam \citep{liu-etal-2020-understanding} is a variant of Adam that uses SGD with momentum in an initial stage with learning rate warmup, followed by a second stage of Adam with a particular setting of learning rate with time.
Thus, one can adapt RAdam to $\mu$P by individually scaling the learning rates of the initial SGD stage and the final Adam stage according to \cref{tab:MUP}, \cref{tab:MUPalt}, or \cref{tab:MUPorig}.

\paragraph{Adagrad and RMSProp}
Exactly the same as Adam in all of our tables.

\paragraph{$\epsilon$ in Adam and Its Variants}
All of our derivations here assume $\epsilon$ is negligible in Adam.
If it is set to a non-negligible number, then it needs to be scaled, for all parameters, like $1/\fanin^2$ if it is added before the square root, or like $1/\fanin$ if it is added after the square root.

\paragraph{Gradient Clipping}
Gradient ($\ell_2$-norm-wise) clipping is compatible with \cref{tab:MUP} (as well as \cref{tab:MUPalt,tab:MUPorig}), for either SGD or Adam, if the clip value is held constant with respect to width.

\paragraph{Weight Decay}
Weight decay should be scaled independently of width in SGD and AdamW, for all of our tables.
However, note it's not compatible with $\mu$P Adam.

\paragraph{Momentum}
Momentum should be scaled independently of width for all of our tables.

\section{Parametrization Matters: A Primer for Multiple Hyperparameters}
\label{sec:primermulti}

Here we give more intuition why we need to reparametrize \emph{all} hyperparameters.
In practice, neural networks have multitudes of hyperparameters all interacting together.
In our example of \cref{sec:primer}, hyperparameter optimization would be akin to minimizing the function%
\footnote{
Here, for simplicity of the example, we model the interaction between ``hyperparameters'' $c^1, \ldots, c^k$ as additive, but in real neural networks such interactions are usually much more complicated.
}
\[
F_{n}(c^1, \ldots, c^k)\defeq\EV_{x_{1},\ldots,x_{n}}f((c^1 + \cdots + c^k)(x_{1}+\cdots+x_{n})).
\]
where $x_1, \ldots, x_n$ are as in \cref{eqn:FnExample} and $c^1, \ldots, c^k$ are analogous to $k$ hyperparameters.
For the same reasoning in \cref{sec:primer}, the \emph{correct parametrization} is in $(\alpha^1, \ldots, \alpha^k)$ where $\alpha^i = c^i \sqrt{n}$.

While this is straightforward, in practice, researchers often fix some hyperparameters (e.g., they tune only learning rate but neglects to scale parameter multipliers or initialization correctly).
For example, if we only partially reparametrize and optimize in $\alpha^1$ while fixing $c^2, \ldots, c^k$, then the optimal $\alpha^1$ is $(\alpha^1)^* = \alpha^* - (c^1 + \ldots + c^k) \sqrt n$ where $\alpha^*$ is the optimal $\alpha$ for \cref{eqn:FnExample}.
Thus, as $n \to \infty$, $(\alpha^1)^*$ still blows up even though we parametrized $\alpha^1$ correctly.
More generally, the incorrect parametrization of some hyperparameters forces other hyperparameters to increasingly compensate for it as width grows, distorting their optima, even if the latter are correctly parametrized.

\section{Practical Considerations}
\label{sec:practical}

In this section, we outline several useful tips and tricks that can improve the quality of hyperparameter transfer in practice.

\subsection{Verifying \texorpdfstring{$\mu$P}{muP} Implementation via \emph{Coordinate Checking}}

Even though $\mu$P is neatly encapsulated by \cref{tab:MUP}, implementing it correctly can in practice be error-prone, just like how implementing autograd by hand can be error-prone even though the math behind is just chain-rule.
In the case of autograd, gradient checking is a simple way of verifying implementation correctness; similarly, we propose \emph{coordinate checking} to verify the correctness of $\mu$P implementation:
Exemplified by \cref{fig:blowupafter1step}, one calculates the average coordinate size of every (pre)activation vector in the network over a few steps of training, as width is varied over a large range.
An incorrect implementation will see some activation vector blow up or shrink to zero with width (like in the top row of \cref{fig:blowupafter1step}).
In the \texttt{mup} package we release with this paper, we include an easy-to-use method for coordinate checking.

\subsection{Zero Initialization for Output Layers and Query Layers in Attention}
\label{sec:outputzeroinit}

We find that the optimal hyperparameters of small and large width models match more closely when we initialize output layers at 0 (i.e.\ with variance $\sigma^2/\fanin{}$ where $\sigma = 0$ instead of positive $\sigma$).
This is because the neural network in $\mu$P is approximately a Gaussian process (GP) at initialization with variance on the order $\Theta(\sigma^2/width)$
(contrast this with SP networks, which approximates a GP with $\Theta(\sigma^2)$ variance)
\citep{TP1,TP4,lee_deep_2018,matthews_gaussian_2018}.
Of course, when width is large, this variance vanishes, but this can be far from so in the small proxy model.
This discrepancy in the initial GP can cause the training trajectory of the proxy model to be very different from the trajectory of the large target model, causing a mismatch in the optimal hyperparameters.
By initializing the output layer at 0, we remove this mismatch in the initial GP.
Empirically we do not find this modification to be detrimental to performance.

A similar consideration applies to the query layer in self-attention:
At initialization, the attention logit $q^\trsp k / {d_{head}}$ looks like a Gaussian with variance $\Theta(1/d_{head})$ because $q$ and $k$ are almost independent and zero-mean.
In the limit $d_{head} \to\infty$, the logit is exactly 0, which can be a large discrepancy compared to when $d_{head}$ is small in the small proxy model we want to tune.
By initializing the query projection matrix $W^q$ to 0, $q$ will also be 0, and hence the attention logit is always 0 at initialization regardless of width (but will generally become nonzero after a gradient step), resolving this discrepancy.

More generally, any layer or computation that goes from an ``infinite'' dimension (i.e.\ width) to a ``finite'' dimension (e.g.\ output dimension or sequence length) can exhibit this kind of discrepancy due to the initial GP.
When $d_{head} \to \infty$ and $n_{head}$ is fixed, attention logit calculation can be viewed in the same vein as a function $\R^{seqlen \times d_{model}} \to \R^{n_{head} \times seqlen \times seqlen}$, which ``reduces to'' $\R^{\infty} \to \R^1$.

\subsection{Activation Functions}

\begin{figure}[H]
    \centering
    \includegraphics[width=\textwidth]{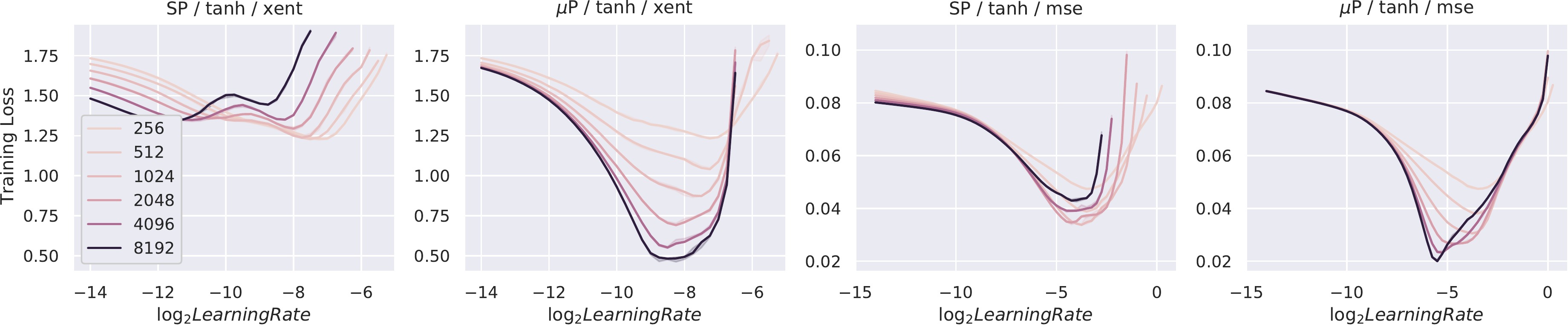}
    \caption[Squashing activation functions reduce transfer quality.]{\textbf{Squashing activation functions reduce transfer quality.}
      MLP of different hidden sizes with \texttt{tanh} activation trained for 20 epoch on CIFAR-10 using SGD. Left uses cross-entropy as loss function; right  uses mean squared error; columns alternate between standard parametrization (SP) and maximal update parametrization ($\mu$P). Compared to \texttt{ReLU}, \texttt{tanh} exhibits slower convergence for $\mu$P, yet it still outperforms SP when width is increased}
    \label{fig: MLP_tanh}
\end{figure}

When the network is narrow, its approximation to the infinite-width behavior becomes crude, which is manifested as large fluctuations in preactivation coordinates.
When using a squashing activation functions like \texttt{softmax} or \texttt{tanh}, this causes narrower networks to saturate the activation more than wider ones, which results in a systematic bias toward small gradients and therefore distorting the hyperparameter landscape.
This can be seen in \cref{fig: MLP_tanh}, where we use \texttt{tanh} as the network activation function.

Therefore, we recommend replacing non-essential squashing activation functions with \texttt{ReLU}, whose derivative depends only on the sign of the pre-activation.
A similar reasoning can be applied to superlinear activation functions, where the distribution of activation values can have heavy tails, leading to slow convergence to the infinite-width limit.
However, such activations are rarely used in practice.

\subsection{Enlarge \texorpdfstring{$d_k$}{Head Dimension}}

\begin{figure}[t]
    \centering
    \includegraphics[width=0.85\textwidth]{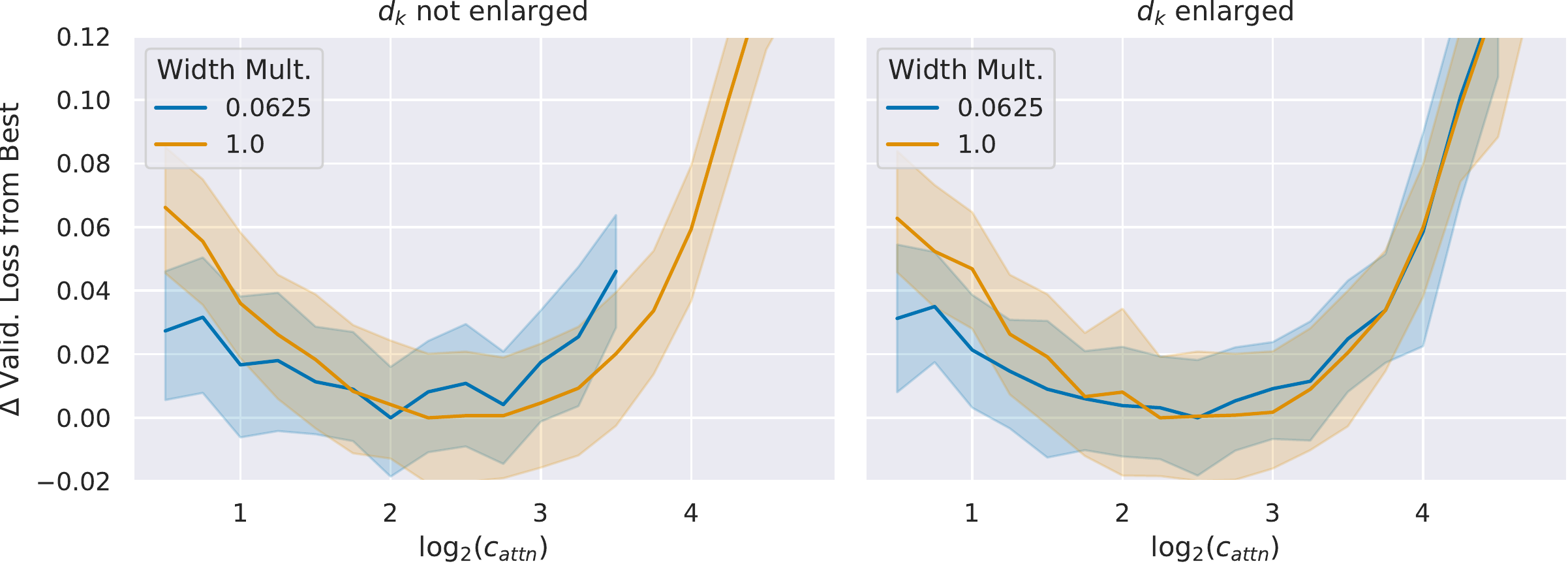}
    \caption[Enlarging $d_{k}$ makes $\mu$Transfer more precise in Transformers]{\textbf{Enlarging $d_{k}$ makes $\mu$Transfer more precise.}
    Here we plot all curves \emph{after subtracting their minima} for easier visual comparison.
      Transformer on IWSLT 14 similar to the setup in~\cref{sec:IWSLTdetails} where the $d_{model}=512$ for a width multiplier of 1, $n_{head}=4$, and $d_q=d_k$.
      \textbf{(Left)} We leave $d_q=d_k=\nicefrac{d_{model}}{n_{head}}$, so $d_k=8$ for width-multiplier 0.0625.
      The optimum for the attention logit multiplier $c_{attn}$ is noisy and does not accurately transfer across width.
      \textbf{(Right)} We enlarge $d_q=d_k$ to a minimum of 128.
      The HP landscape is much smoother than in (Left), and the optima align between narrow and wide models.}
    \label{fig:Widen_QK}
\end{figure}

We find that small $d_{head} = d_k$ can lead to a highly noisy HP landscape, as shown in \cref{{fig:Widen_QK}}.
This can significiantly decrease the quality of random HP search on the small proxy model.
To solve this, we find it useful to decouple $d_k$ from $d_{model}$ (so that $d_{model} \ne d_k \cdot n_{head}$) and maintain a relatively large $d_k$ even as $d_{model}$ is shrunk in the proxy model.
For example, pegging $d_{k} = 32$ is generally effective.
Training or inference speed are not usually affected much by the larger $d_k$ because of CUDA optimizations.
By \cref{sec:width}, this decoupling of $d_k$ from $d_{model}$ is theoretically justified, and as shown in \cref{{fig:Widen_QK}}, it significantly denoises the HP landscape.

\subsection{Non-Gaussian vs Gaussian Initialization}

We find non-Gaussian (e.g.\ uniform) initialization can sometimes cause wider models to perform worse than narrower models, whereas we do not find this behavior for Gaussian initialization.
This is consistent with theory, since in the large width limit, one should expect non-Gaussian initialization to behave like Gaussian initializations anyway (essentially due to Central Limit Theorem, or more precisely, universality), but the non-Gaussianity slows down the convergence to this limit.

\subsection{Using a Larger Sequence Length}

For Transformers, we empirically find that we can better transfer initialization standard deviation from a narrower model (to a wide model) if we use a larger sequence length.
It is not clear why this is the case.
We leave an explanation to future work.

\subsection{Tuning Per-Layer Hyperparameters}
The techniques in this paper allow the transfer across width of (learning rate, initialization, multipliers) simultaneously for all parameter tensors.
Thus, to get the best results, one should ideally tune all such hyperparameters.
In practice, we find that just tuning the global learning rate and initialization, along with input, output, and attention multipliers, yield good results.

\section{Which Hyperparameters Can Be Transferred? (Continued)}

\subsection{Further Discussions on Hyperparameter Categories}
\label{sec:hypdiscussions}

Below, we discuss the reasoning behind each kind, which are supported by our empirical evidence collected in \cref{fig:wikitext2_mega} on Transformers as well as those in \cref{sec:resnetexperiments} on ResNet.

\paragraph{Transferable Hyperparameters}
\label{sec:transferable}

In \cref{tab:transferables}, we summarize which HPs can be transferred across training scale.
The transfer across \emph{width}, as explained in \cref{sec:primer}, is theoretically justified, while we present the transfer across the other dimensions as empirical results.

These cover most of the well-known and important HPs when the need for regularization is not paramount, e.g., during large scale language model pretraining.
{Parameter Multipliers} are not well-known HPs, yet we include them here as they serve a bridge between SP and $\mu$P and can impact model performance in practice.
Concretely, any SP and $\mu$P neural networks of the same width can have their Parameter Multipliers tuned so that their training dynamics become identical.

\paragraph{Hyperparameters That Don't Transfer Well}
Not all HPs transfer well even if we use $\mu$P.
In particular, those whose primary function is to regularize training to mitigate ``overfitting" tend not to transfer well.
Intuitively, regularization needs to be applied more heavily in larger models and when data is scarce, but $\mu$P does not know the data size so cannot adjust the regularization accordingly.

To the best of our knowledge, there is no strict separation between HPs that regularize and those that don't.
However, conventional wisdom tells us that there exists a spectrum of how much regularizing effect a HP has.
For example, dropout probability and weight decay are among those whose primary function is to regularize, whereas batch size and learning rate might regularize training in some cases but affect the dynamics more so in other ways.
Our empirical exploration tells us that the former do not transfer well, while the latter do.
Our subsequent discussion will focus on the latter; we leave to future works the expansion to the former.

\paragraph{Hyperparameters Transfered \emph{Across}}

We have left out a category of HPs that defines the training \emph{scale}, or in practical terms, training cost.
This includes 1) those that define how many operations a model's forward/backward pass takes, such as the model's width, depth, and in the case of language modeling, sequence length; and 2) those that define how many such passes are performed, such as batch size and number of training steps.

As recent works have shown \citep{rosenfeld_constructive_2019,brown2020language,kaplan_scaling_2020}, improvements along any of these \emph{scale} dimensions lead to apparently sustainable gain in performance; as a result, we are primarily interested in transferring other HPs \emph{across} these dimensions that define scale, rather than finding the optimal scale.%
\footnote{
In particular, we are not fixing the total training FLOPs when we scale, which requires understanding the tradeoff of different scale HPs.
For example, when we transfer across batch size, we \emph{fix} the number of steps of training (\emph{not} the number of epochs), so that the total FLOPs scales linearly.}
This category of HPs is particularly crucial as one can speedup training by downsizing in one or multiple such dimensions.
Indeed, it's very common for practitioners to implicitly transfer HPs across the number of training samples by tuning on only a subset of the full training data.

Our insights from the infinite-width limit inspired us to explore HP tranfer across \emph{width}, which does not work under SP as we have shown earlier.
Building upon our success with width, which is well explained theoretically, we hope to push the limit of compute-saving by investigating the other dimensions empirically.
To the best of our knowledge, the transferability of optimal HPs across depth, batch size, sequence length, and training time has not been rigorously investigated previously, with the main exception of the literature on (learning rate, batch size) scaling \citep{smith_dont_2017,shallue_measuring_2018} where our transferability result of learning rate across batch size recapitulates \citep{mccandlish_empirical_2018}.%
\footnote{
There's also a literature on the proper initialization for training deep networks effectively (e.g.\ \cite{schoenholz_deep_2016,yang_mean_2017,yang_layerwise_2018,zhang_residual_2019,bachlechner_rezero_2020,huang_improving_nodate,liu_understanding_2020}), but they do not study the \emph{transferability} per se.
See \cref{sec:previousscalings_appendix}}
See \cref{sec:previousscalings_appendix} on how our results relate to prior works.
We will primarily focus on the Transformer architecture in the main text with evidence for ResNet in \cref{sec:resnetexperiments}.

\subsection{On the Definitions of Width}
\label{sec:width}

\begin{figure}
  \centering
  \includegraphics[width=0.8\textwidth]{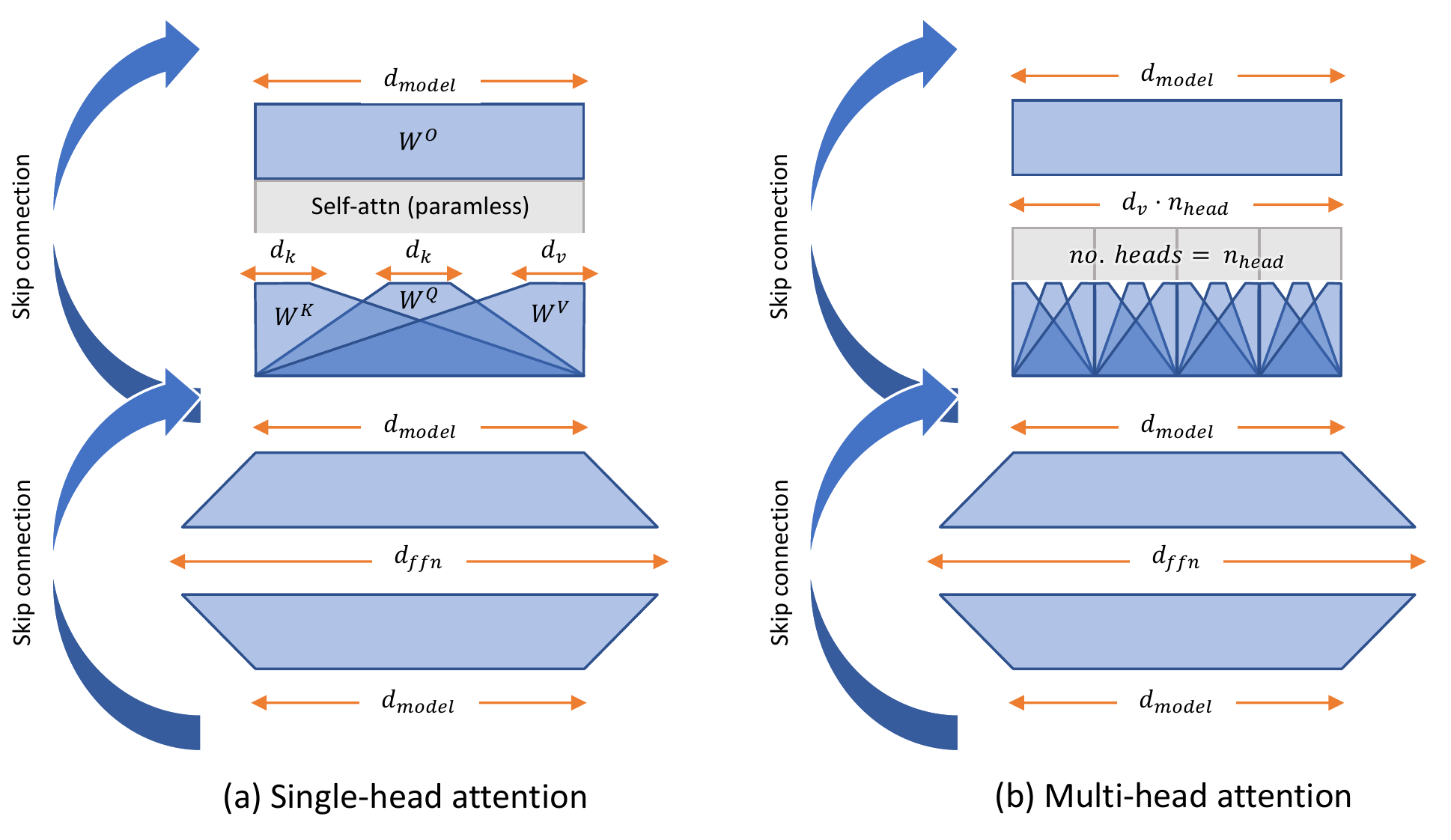}
  \caption[Schematics of each Transformer layer]{\textbf{Schematics of each Transformer layer.} Commonly, the key and value dimensions $d_k$ and $d_v$ are both set to $d_{model} / n_{head}$, and this is referred to as $d_{head}$.}
  \label{fig:transformerschematics}
\end{figure}

Our theory allows more general notions of width.
This is especially relevant in Transformers, where $d_{model}, d_{head} = d_k, d_v, n_{head}, d_{ffn}$ (see \cref{fig:transformerschematics}) can all be construed as measures of width.
We briefly discuss these here, with more theoretical justification in \cref{sec:MUPDerivation} and empirical validation below.

\paragraph{Varying Width Ratio}
So far we have assumed that every hidden layer is widened by the same factor.
But in fact we can widen different hidden layers differently.
This is useful, for example, in a Transformer where we may want to use a smaller $d_{ffn}$ during tuning.
If we are using Adam, as long as the width of every layer still tends to infinity, we still obtain approximately the same limit%
\footnote{
    This also applies for SGD, but we need more involved scaling to keep the limit approximately the same.
}, so the $\mu$Transfer remains theoretically justified.

\begin{figure}
    \centering
    \includegraphics[width=0.32\textwidth]{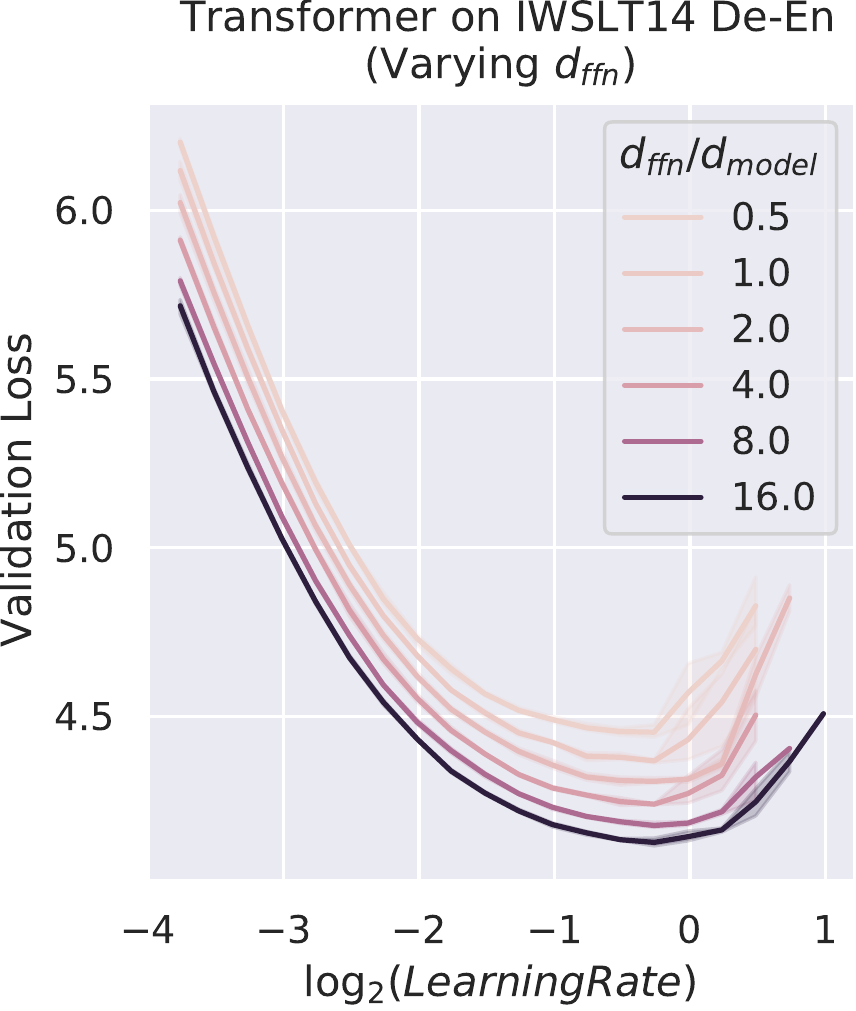}
    \caption[Width ratio can be varied arbitrarily in $\mu$Transfer]{Learning rate landscape in $\mu$P is stable even if we vary $d_{ffn}$ by a factor of 32, fixing $d_{model}$.}
    \label{fig: iwslt_ffn}
\end{figure}

See \cref{fig: iwslt_ffn} for an empirical validation on IWSLT-14 using a Transformer.

\paragraph{Number of Attention Heads}

\begin{figure}
    \centering
    \includegraphics[width=0.85\textwidth]{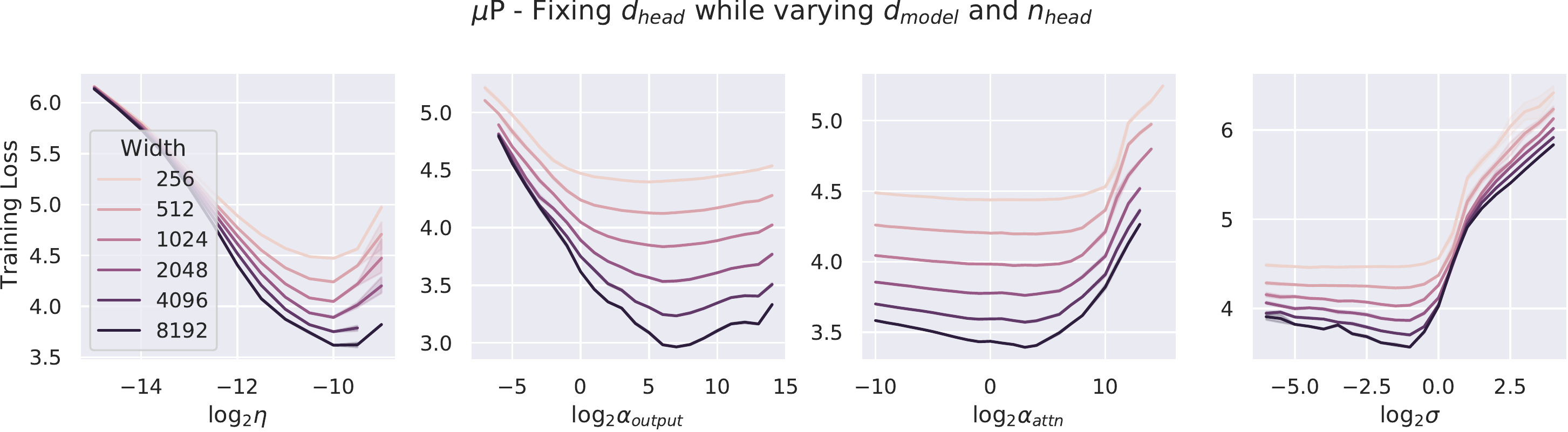}
    \caption[$\mu$Transfer can handle increasing $n_{head}$ while fixing $d_{head}$ as well as increasing $d_{head}$ while fixing $n_{head}$, or a mix of both]{$\mu$Transfer across width when we fix $d_{head}$ and vary $d_{model}$ and $n_{head}$.
    $\alpha_{output}, \alpha_{attn}$ are multipliers for output and key weights, and $\sigma$ is initialization standard deviation.}
    \label{fig: wikitext2_nhead}
\end{figure}

In attention-based models, one typically splits hidden size into multiple attention heads following $d_{model} = d_{head} \times n_{head}$.
So far we have assumed $d_{head}$ and $d_{model}$ to be width, but it's possible and potentially advantageous to fix $d_{head}$ and treat $n_{head}$ as the width, or increasing both simultaneously.
This allows our technique to handle many popular models, including GPT-3~\cite{brown2020language}, which scale up by fixing $d_{head}$ and increasing $n_{head}$.
See \cref{fig: wikitext2_nhead} for an empirical validation on Wikitext-2.

\paragraph{Varying Just the Width of Attention Heads}

A specific useful instance of varying width ratio is decoupling the key and value dimensions $d_k$ and $d_v$ and scaling $d_k$ differently from (typically larger than) $d_{model} / n_{head}$.
This works as long as we use $1/d$ scaled-attention as in \cref{defn:trsfmrMUP} (instead of $1/\sqrt d$ as is done commonly).
When tuning on the small proxy model, if $d_k$ is too small, the HP landscape can be quite noisy.
Keeping $d_k$ relatively large while shrinking all other dimensions solves this problem, while still obtaining significant speedup.

\section{Experimental Details}
\label{sec:experimentaldetails}

\subsection{IWSLT}
\label{sec:IWSLTdetails}
IWSLT14 De-En is a well-known machine translation benchmark.
We use a Transformer implemented in \texttt{fairseq}~\citep{ott2019fairseq} with a default $d_{model} = \nicefrac{1}{4} d_{ffn} = 512$ and $d_{k}=d_{q}=d_{v}=\nicefrac{d_{model}}{n_{head}}=128$ (amounting to 40M parameters), which we denote as the \emph{1x model}.
For transfer, we tune on a proxy model with the same $n_{head}$ 
but with $d_{model}$ and other dimensions 4 times smaller; we will call this the \emph{0.25x model} (but it has 4M parameters).
All models are trained with Adam for 100 epochs and validated at the end of every epoch.
We tune via random search the learning rate $\eta$, the output layer parameter multiplier $\alpha_{output}$, and the attention key-projection weight multiplier $\alpha_{attn}$ following the grid
\begin{itemize}
    \item $\eta$: $5 \times 10^{-4} \times 2^z, \text{where } z\in \{-1.5, -1.25, -1, ..., 1.25\}$ %
    \item $\alpha_{output}$: $2^z, \text{where } z\in \{-8, -7, -6, ..., 7\}$ %
    \item $\alpha_{attn}$: $2^z, \text{where } z\in\{-3, -2, -1, ..., 8\}$ %
\end{itemize}

\subsection{WMT}
\label{sec:WMTdetails}

We scale up to WMT14 En-De using the large Transformer from~\cite{DBLP:journals/corr/VaswaniSPUJGKP17}, with a $d_{model}=\nicefrac{1}{4}d_{ffn}=1024$ and $d_{q}=d_{k}=d_{v}=\nicefrac{d_{model}}{n_{head}}=64$.
We use the exact same setup and reproduce their result as our baseline.
Then, we build the proxy model by shrinking the target model's $d_{model}$ from the original $1024$ to $256$, $d_{ffn}$ from $4096$ to $256$ and $n_{head}$ from $16$ to $4$.
This reduces the total parameter count from 211M to 15M.
We then perform the HP search on the proxy model and take the best according to validation loss, before testing on the target model.
We tune via random search the learning rate $\eta$, the output layer parameter multiplier $\alpha_{output}$, and the attention key-projection weight multiplier $\alpha_{attn}$ following the grid
\begin{itemize}
    \item $\eta$: $6 \times 10^{-4} \times 2^z, \text{where } z\in \{-1.5, -1.25, -1, ..., 1.25\}$
    \item $\alpha_{output}$: $2^z, \text{where } z\in \{-8, -7, -6, ..., 7\}$
    \item $\alpha_{attn}$: $2^z, \text{where } z\in\{-3, -2, -1, ..., 8\}$
\end{itemize}

\subsection{BERT}
\label{app:bert}
\label{app:bert_prototype}

\paragraph{Details of BERT Prototype}
Our proxy model has 10 Transformer layers with $d_{model}=d_{ffn}=256$.
We also reduce the number of attention heads to 8 with a $d_{head}$ of 32.
We call it BERT Prototype since we can increase its width and depth according to our definitions to recover both BERT Base and BERT Large, which enables us to sweep HPs once and use for both models.
Overall, BERT Prototype has 13M trainable parameters, a fraction of the 110M in BERT Base and the 350M in BERT Large.

\paragraph{Hyperparameters Tuned for Pretraining}
We tune the following HPs for pretraining: Adam learning rate $\eta$, embedding learning rate $\eta_{emb}$, output weight multiplier $\alpha_{output}$, attention logits multiplier $\alpha_{attn}$, layernorm gain multiplier $\alpha_{LN_{gain}}$, and bias multiplier $\alpha_{bias}$.

We sample 256 combinations from the follow grid:
\begin{itemize}
    \item $\eta$: $1 \times 10^{-4} \times 2^z, \text{where } z\in \{1.5, 2, 2.5, 3, 3.5\}$ 
    \item $\eta_{emb}$: $1 \times 10^{-4} \times 2^z, \text{where } z\in \{-1, -0.5, 0, 0.5, 1\}$ 
    \item $\alpha_{output}$: $2^z, \text{where } z\in \{2, 4, 6\}$
    \item $\alpha_{attn}$: $2^z, \text{where } z\in\{3, 3.5, 4, ..., 7\}$
    \item $\alpha_{LN_{gain}}$: $2^z, \text{where } z\in\{8.5, 9, 9.5, 10, 10.5\}$
    \item $\alpha_{bias}$: $2^z, \text{where } z\in\{8.5, 9, 9.5, 10, 10.5\}$
\end{itemize}
The ranges are chosen to include the implicit choices of these HPs in SP BERT Large.

\paragraph{Finetuning Procedure and Hyperparameters}
We hand-pick the finetuning HPs after training the full-sized model.
As regularization is an essential ingredient in successful finetuning, we do not transfer such HPs (at least via the suite of techniques presented in this work) (see \cref{tab:whichtransfer}).
We focus on MNLI \cite{mnli2018} and QQP, which are two representative tasks from GLUE \cite{wang2018glue}. Following \cite{liu2019mt-dnn}, we used Adam \cite{kingma2014adam} with a learning rate of $5\times 10^{-5}$ and a batch size of 64.
The maximum number of epochs was set to $5$. 
A linear learning rate decay schedule with warm-up of $0.1$ was used. 
All the texts were tokenized using wordpieces and were chopped to spans no longer than $128$ tokens. 

\subsection{GPT-3}
\label{sec:gpt3-appendix}

\paragraph{Baseline 6.7B GPT-3 Transformer}
As the GPT-3 codebase has evolved since the publication of \citep{brown2020language}, we re-trained the 6.7B model from scratch to remove changes in our codebase as a possible confounder.
The main differences to \citep{brown2020language} 
are 1) a modified learning rate decay schedule, where the learning rate is decayed to zero at the end of training rather than being decayed to 0.1 of the initial value, and 2) use of relative attention in place of absolute attention.
Unfortunately, after all experiments were finished, we found this re-run baseline used absolute attention instead of relative attention, while the $\mu$Transfer model still used relative attention.

\paragraph{Random Search using Reduced-Width Proxy Model}
In order to find a good set of hyperparameters for the $\mu$Transfer version of the 6.7B model, we performed a hyperparameter search over a reduced version of the model (i.e., the proxy model), where the width is set to 256 hidden units.
This proxy model inherits changes from the evolved GPT-3 codebase: it uses relative~\citep{dai2019transformerxl} (instead of absolute) position encoding.
Early on, we noted that on the proxy model, linear learning rate decay outperformed the default cosine schedule, so all subsequent experiments for the proxy models use a linear decay schedule.
By \cref{fig:wikitext2_mega}, $\mu$Transferring this linear decay schedule to the full model should maintain such a performance advantage over the cosine schedule.

The hyperparameter search space consists of the following hyperparameters:

\begin{itemize}
    \item \textbf{learning rate:} Sampled from $10^{\text{Uniform}(-4, -1)}$
    \item \textbf{initialization scale:} All the parameters are multiplied - sampled from $10^{\text{Uniform}(-1, 1)}$
    \item \textbf{attention temperature:} Reciprocal of the multiplier applied to the input to attention softmax. Sampled from $4^{\text{Uniform}(-1, 1)}$.
    \item \textbf{output temperature:} Reciprocal of the multiplier applied to the input to softmax that produces the distribution over output tokens. Sampled from $4^{\text{Uniform}(-1, 1)}$.
    \item \textbf{embedding multiplier:} Scalar by which we multiply the output of the embedding layer. Sampled from $10^{\text{Uniform}(-1, 1)}$.
    \item \textbf{relative position embedding multiplier:} Scalar by which we multiply vectors representing relative position. Sampled from $10^{\text{Uniform}(-1, 1)}$.
\end{itemize}

In order to make the search more efficient we reduced the total number of training tokens.
We hypothesized that tuning hyperparameters on a reduced total number of tokens does not significantly affect optimal hyperparameters.
To verify, we trained two different horizons and compared the results.
While the target model was to be trained on 300 billion tokens, we tuned the proxy model on only subsets consisting of 4 billion and 16 billion tokens.
This impacts both the total training time and the length of the linear learning rate decay schedule.
Other than hyperparameters explicitly listed above and the training horizon, the rest was the same as what we intended to use for the full width 6.7B training run.

\begin{figure}
    \center
    \begin{subfigure}{.5\textwidth}
      \centering
      \includegraphics[width=\linewidth]{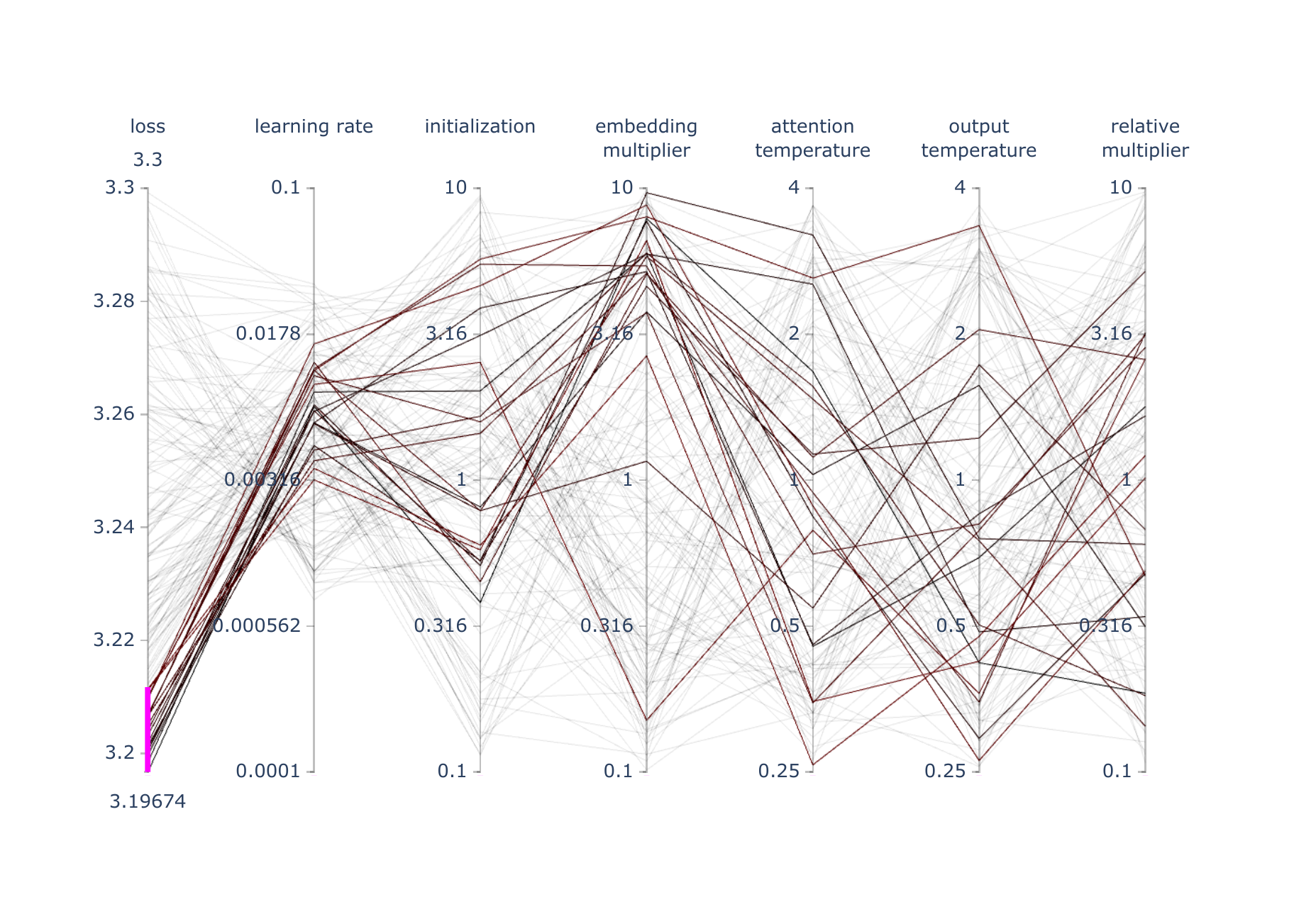}
    \end{subfigure}%
    \begin{subfigure}{.5\textwidth}
      \centering
      \includegraphics[width=\linewidth]{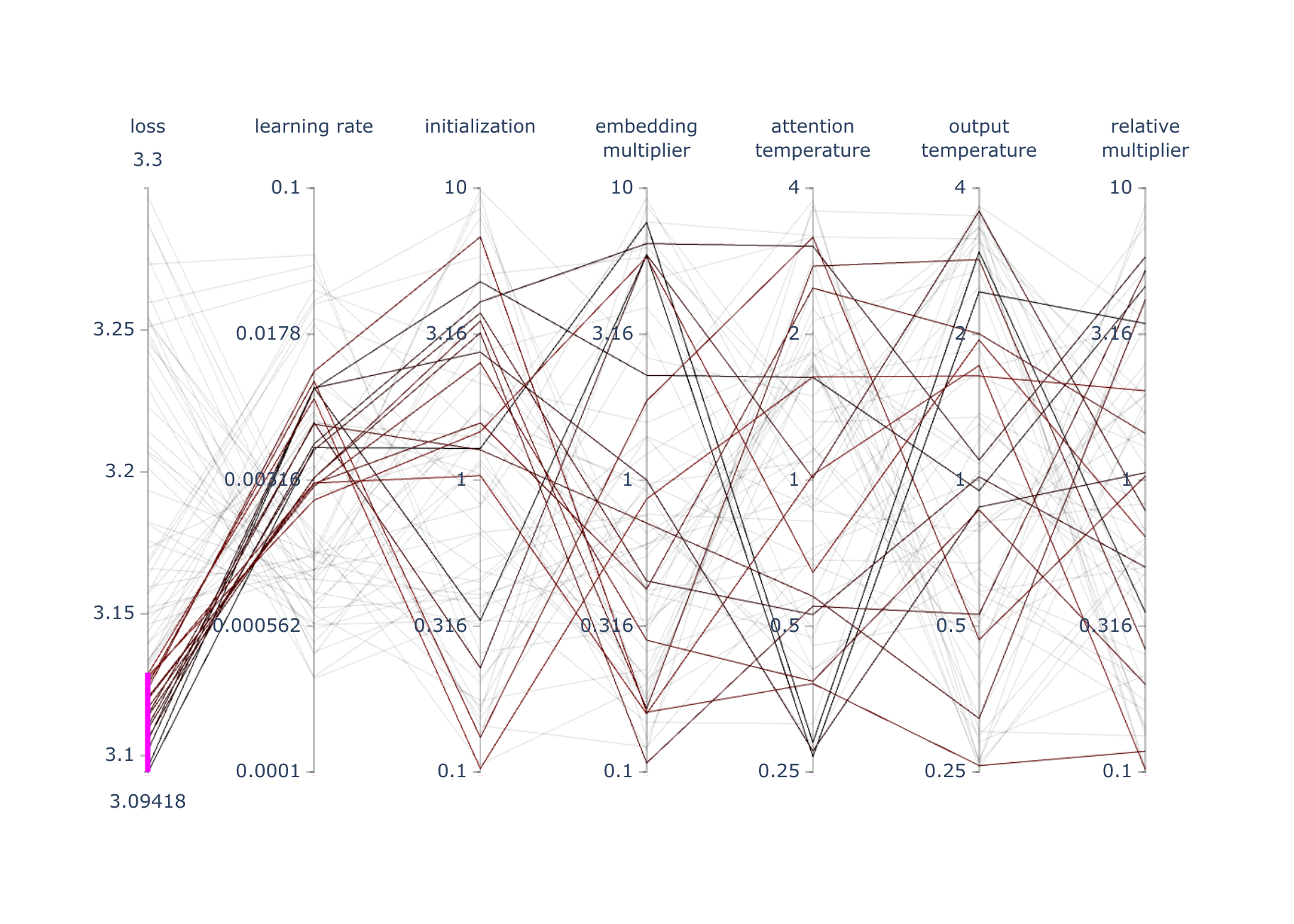}
    \end{subfigure}
    \caption[Results of the random search over reduced-width GPT-3 proxy models]{\textbf{Results of the random search over reduced-width GPT-3 proxy models} trained on 4 (left) and 16 (right) billion tokens. Only the best performing runs are highlighted.}
    \label{fig:4and16bres}
\end{figure}

\paragraph{Analyzing the Results of the Random Search}
We performed 467 training runs of the proxy model, out of which 350 were for 4 billion tokens (286 completed without diverging) and 117 for 16b tokens (80 completed without diverging).
See \cref{fig:4and16bres} for summary of the results.

As suspected, we observed that the results are well-aligned for both 4 and 16 billion tokens versions.
We observe learning rate and initialization scale impact the results the most. Based on the results we chose 0.006 for the former and 2.5 for the latter.
Since most other hyperparameters appear to have negligible effect on performance, they were kept at their default values of 1, the only exception being the embedding scale, where higher values seem to perform better and it was therefore set to 10.

\paragraph{Training the $\mu$Transfer Model}

\begin{figure}
    \center
    \includegraphics[width=0.8\textwidth]{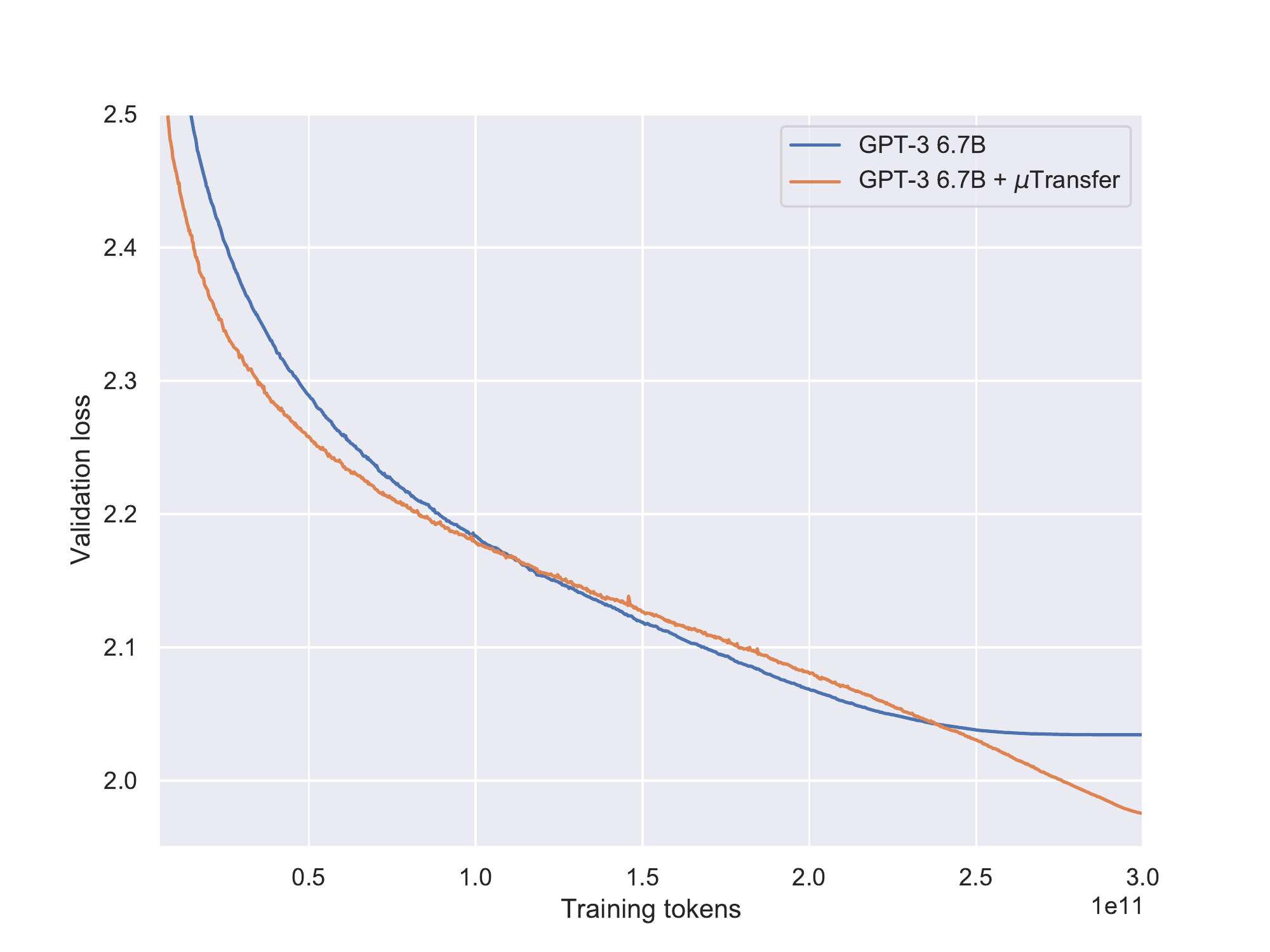}
    \caption[The training curves of the GPT-3 6.7B model with $\mu$Transfer and a re-run with the original settings from \cite{brown2020language}]{
        \textbf{The training curves of the GPT-3 6.7B model with $\mu$Transfer (orange) and a re-run with the original settings from \cite{brown2020language} (blue).}
        The $\mu$Transfer model uses relative attention while the re-run uses absolute attention.
        In addition, the former was trained using FP32 activations and weights after initially encountering stability issues with the hyperparameters computed using $\mu$P, while the re-run used the original FP16 training.
        The $\mu$Transfer model seems to underperform in the middle of training, but achieves a much better final validation loss once the learning rate is fully decayed.
        While the original model uses a cosine schedule, the $\mu$Transfer model uses a linear learning rate decay schedule transferred from the proxy model.
    }
    \label{fig:gpt3-curves}
\end{figure}

We encountered frequent divergences in our initial attempt to train the $\mu$Transfer model.
We traced the issue back to underflow of FP16 tensors in the backwards pass and therefore switched to training the model in FP32.
This allowed us to finish the training run without divergences.
We hypothesize that the divergence issue is related to $\mu$Transfer picking more aggressive hyperparameters, for example a higher learning rate on linear weight tensors compared to the original model.
In order to exclude code differences as a possible confounder, we re-trained GPT-3 6.7B from scratch using the original hyperparameters.
The only difference compared to the version published in \cite{brown2020language} is that the learning rate was decayed fully, whereas the learning rate of the model from \cite{brown2020language} was only decayed to 10\% of its starting value.
The retrained model performs slightly worse than the original published in \cite{brown2020language}.
We suspect that this is because it made less progress during the last phase of training where the learning rate is close to zero.
The training curves of the $\mu$Transfer model and the re-run of the original 6.7B can be seen in \cref{fig:gpt3-curves}.
Detailed evaluation results can be found in \cref{tab:gpt3-evals} and \cref{tab:gpt3-evals-13b-comparison}.

\paragraph{Ratio of Tuning Cost to Pretraining Cost}
in FLOPs can be approximated as
\[
  \frac
  {s (t_1 N_1 + t_2 N_2)}
  {S T} \approx 0.07\]
where
\begin{itemize}
  \item $s = 40$ Million is number of parameters of the proxy model
  \item $S = 6.7$ Billion is number of parameters of the target model
  \item $t_1= 4$ Billion is the number of training tokens for the short horizon HP search, and $N_1 = 350$ is the corresponding number of random HP search trials.
  \item $t_2= 16$ Billion is the number of training tokens for the longer horizon HP search, and $N_1 = 117$ is the corresponding number of random HP search trials.
  \item $T = 300$ Billion is the number of training tokens for the 6.7B target model.
\end{itemize}
Here we are using the fact that the training FLOPs of a Transformer per token is roughly proportional to its number of parameters.

\begin{table}
\caption[Full evaluation results of our GPT-3 6.7B models]{
  \textbf{Full evaluation results of our GPT-3 6.7B models}: The new model tuned with $\mu$Transfer (marked \emph{$\mu$P}), the original model from \cite{brown2020language}, and a re-training of this model from scratch with the original hyperparameter settings (marked \emph{re-run}).
    The sampling-based evaluations shown here are a subset of the ones from \cite{brown2020language}.
    Since the sampling-based evaluations are subject to high variance, Wikitext 103 and the LM1B benchmark have been added to help distinguish the relative performance of the $\mu$P and non-$\mu$P model.
    Note that Wikitext-103 \citep{wikitext103} and the LM1B \citep{1billionword} benchmarks overlap with the training dataset.
    Accuracies and F1 scores have been multiplied by 100.
    The perplexities reported in this table are based on a custom BPE encoding and are not comparable to other results in the literature.
    The number $k$ of examples in the context for each task is identical to \cite{brown2020language}. \\
    \emph{Note:} Zero-shot, One-Shot and Few-Shot refer to the number of additional query and answer pairs passed in the context when performing the sampling-based evaluations, not the ''shots'' involved in hyperparameter transfer.
}
\footnotesize
\addtolength{\tabcolsep}{-2pt}
\centering
\begin{tabular}{lllSSSSSSSSS}
\toprule
    &&& \multicolumn{3}{c}{Zero-shot} & \multicolumn{3}{c}{One-shot} & \multicolumn{3}{c}{Few-shot} \\
\cmidrule(lr){4-6} \cmidrule(lr){7-9} \cmidrule(lr){10-12}
    Task & Split & Metric & {$\mu$P} & {\cite{brown2020language}} & {re-run} & {$\mu$P} & {\cite{brown2020language}} & {re-run} & {$\mu$P} & {\cite{brown2020language}} & {re-run} \\
\midrule
    Validation dataset & valid & ce & \bfseries 1.98 & & 2.03 & & & & & & \\

    PTB & test & ppl & \bfseries 11.3921 & & 12.986696243286133 & & & & & & \\

    Wikitext 103 & test & ppl & \bfseries 8.5612 & & 9.129402160644531 & & & & & & \\

    LM1B & test & ppl & \bfseries 20.5135 & & 21.73866081237793 & & & & & & \\

\midrule

    HellaSwag & dev & acc &
    \bfseries 71.9877 & 67.4 & 66.6899025440216 &
    \bfseries 71.0516 & 66.5 & 65.85341691970825 &
    \bfseries 72.4358 & 67.3 & 66.43099188804626
    \\

    LAMBADA & test & acc &
    \bfseries 73.4912 & 70.3 & 70.8131194114685 &
    \bfseries 69.8816 & 65.4 & 64.75839614868164 &
    74.71376061439514 & \bfseries 79.1000 & 77.12012529373169
    \\

    StoryCloze & test & acc &
    \bfseries 79.4228 & 77.7 & 77.33832001686096 &
    \bfseries 80.5986 & 78.7 & 78.30037474632263 &
    \bfseries 84.2330 & 81.2 & 81.13308548927307
    \\

    NaturalQS & test & acc &
    \bfseries 9.8615 & 5.79 & 7.202216066481995 &
    \bfseries 14.6814 & 9.78 & 10.609418282548475 &
    \bfseries 20.2216 & 17.0 & 15.706371191135734
    \\

    TriviaQA & dev & acc &
    \bfseries 47.0036 & 38.7 & 37.50781934192418 &
    \bfseries 50.4441 & 44.4 & 42.462154385086954 &
    \bfseries 55.5361 & 51.6 & 49.85612410859502
    \\

    WebQS & test & acc &
    \bfseries 11.3189 & 7.73 & 9.793307086614174 &
    \bfseries 20.2264 & 15.1 & 16.190944881889763 &
    \bfseries 32.9724 & 27.7 & 28.198818897637796
    \\

    Ro$\to$En 16 & test & BLEU-sb &
    \bfseries 26.9000 & 8.75 & 13.699999809265137 &
    \bfseries 36.5000 & 34.2 & 33.5 &
    \bfseries 38.2000 & 36.2 & 35.599998474121094
    \\

    En$\to$Ro 16 & test & BLEU-sb &
    \bfseries 18.1000 & 5.31 & 4.400000095367432 &
    \bfseries 21.0000 & 18.2 & 17.299999237060547 &
    \bfseries 22.0000 & 19.6 & 18.799999237060547
    \\

    Fr$\to$En 14 & test & BLEU-sb &
    \bfseries 29.8000 & 15.5 & 19.600000381469727 &
    \bfseries 31.7000 & 31.6 & 30.100000381469727 &
    \bfseries 38.0000 & 36.4 & 36.5
    \\

    En$\to$Fr 14 & test & BLEU-sb &
    \bfseries 29.6000 & 11.4 & 11.600000381469727 &
    \bfseries 28.8000 & 28.3 & 26.0 &
    33.29999923706055 & \bfseries 33.3000 & 31.200000762939453
    \\

    De$\to$En 16 & test & BLEU-sb &
    \bfseries 31.7000 & 18.2 & 21.700000762939453 &
    \bfseries 33.3000 & 31.9 & 31.100000381469727 &
    \bfseries 38.9000 & 36.5 & 36.20000076293945
    \\

    En$\to$De 16 & test & BLEU-sb &
    \bfseries 23.1000 & 9.36 & 9.0 &
    \bfseries 24.6000 & 21.7 & 21.100000381469727 &
    \bfseries 27.6000 & 24.1 & 24.5
    \\

    Winograd & test & acc &
    85.34798622131348 & 85.7 & \bfseries 86.8132 &
    \bfseries 84.6154 & 84.6 & 84.24908518791199 &
    \bfseries 86.4469 & 85.4 & 83.88278484344482
    \\

    Winogrande & dev & acc &
    \bfseries 66.7719 & 64.5 & 62.50986456871033 &
    \bfseries 67.5612 & 65.8 & 64.48302865028381 &
    \bfseries 71.0339 & 67.4 & 67.1665370464325
    \\

    PIQA & dev & acc &
    \bfseries 79.0533 & 78.0 & 77.96518206596375 &
    \bfseries 77.3123 & 76.3 & 76.87703967094421 &
    \bfseries 79.2165 & 77.8 & 77.74755358695984
    \\

    ARC (Challenge) & test & acc &
    42.13286638259888 & 41.4 & \bfseries 42.4825 &
    \bfseries 43.9685 & 41.5 & 42.395105957984924 &
    \bfseries 43.7937 & 43.7 & 42.74475574493408
    \\

    ARC (Easy) & test & acc &
    \bfseries 64.2857 & 60.2 & 61.860668659210205 &
    \bfseries 65.3439 & 62.6 & 63.447970151901245 &
    \bfseries 67.3280 & 65.8 & 65.34391641616821
    \\

    OpenBookQA & test & acc &
    \bfseries 54.4000 & 50.4 & 52.60000228881836 &
    \bfseries 56.4000 & 53.0 & 52.799999713897705 &
    \bfseries 58.4000 & 55.2 & 54.40000295639038
    \\

    Quac & dev & f1 &
    \bfseries 41.7609 & 36.1 & 38.15895247309869 &
    \bfseries 43.1352 & 39.0 & 39.537164749562244 &
    \bfseries 44.0041 & 39.9 & 39.94074638688334
    \\

    RACE-h & test & acc &
    \bfseries 45.0257 & 44.1 & 43.196111929423466 &
    \bfseries 44.8828 & 44.3 & 42.881646725094065 &
    \bfseries 45.1973 & 44.7 & 43.3676390255295
    \\

    RACE-m & test & acc &
    \bfseries 58.3565 & 54.4 & 53.96936026124237 &
    \bfseries 57.9387 & 54.7 & 53.830084611446416 &
    \bfseries 58.5655 & 55.4 & 55.43175382866501
    \\

    SQuADv2 & dev & f1 &
    \bfseries 59.9188 & 52.7 & 50.919942008123755 &
    \bfseries 64.9009 & 57.1 & 54.71815738148722 &
    \bfseries 68.8726 & 62.1 & 58.41277489166473
    \\

    CoQA & dev & f1 &
    \bfseries 78.5000 & 72.8 & 72.9 &
    \bfseries 80.9000 & 75.1 & 74.4 &
    \bfseries 81.3000 & 77.3 & 75.4
    \\

    DROP & dev & f1 &
    17.115981543624162 & 17.0 & \bfseries 17.3645 &
    23.292575503355703 & \bfseries 27.3000 & 25.698301174496642 &
    \bfseries 33.9242 & 29.7 & 28.656564597315437
    \\

    BoolQ & dev & acc &
    \bfseries 69.3578 & 65.4 & 60.948012232415905 &
    \bfseries 74.0673 & 68.7 & 64.95412844036697 &
    \bfseries 73.9144 & 70.0 & 69.72477064220183
    \\

    CB & dev & acc &
    21.428571428571427 & 28.6 & \bfseries 37.5000 &
    \bfseries 60.7143 & 33.9 & 32.142857142857146 &
    62.5 & 60.7 & \bfseries 66.0714
    \\

    Copa & dev & acc &
    \bfseries 82.0000 & 80.0 & 76.99999809265137 &
    81.00000023841858 & \bfseries 82.0000 & 81.00000023841858 &
    \bfseries 88.0000 & 83.0 & 81.99999928474426
    \\

    RTE & dev & acc &
    \bfseries 55.2347 & 55.2 & 46.20938628158845 &
    \bfseries 61.0108 & 54.9 & 58.844765342960294 &
    52.707581227436826 & 49.5 & \bfseries 59.9278
    \\

    WiC & dev & acc &
    \bfseries 0.0000 & \bfseries 0.0000 & \bfseries 0.0000 &
    50.0 & 50.3 & \bfseries 50.3135 &
    50.470219435736674 & \bfseries 53.1000 & 51.2539184952978
    \\

    ANLI R1 & test & acc &
    \bfseries 33.7000 & 32.3 & 33.399999141693115 &
    \bfseries 32.4000 & 31.6 & 31.700000166893005 &
    30.899998545646667 & \bfseries 33.1000 & 30.700001120567322
    \\

    ANLI R2 & test & acc &
    \bfseries 33.8000 & 33.5 & 33.000001311302185 &
    \bfseries 34.8000 & 33.9 & 33.70000123977661 &
    \bfseries 35.0000 & 33.3 & 32.19999969005585
    \\

    ANLI R3 & test & acc &
    32.66666531562805 & \bfseries 34.8000 & 33.41666758060455 &
    \bfseries 34.7500 & 33.1 & 33.250001072883606 &
    \bfseries 36.9167 & 33.9 & 32.33333230018616
    \\

\bottomrule
\end{tabular}
\label{tab:gpt3-evals}
\end{table}

\begin{table}
\caption[Our $\mu$Transferred GPT-3 6.7B model performs comparably to the twice-as-large GPT-3 13B model from \cite{brown2020language}]{
    \textbf{Evaluation results comparing the GPT-3 6.7B model tuned with $\mu$Transfer against the twice-as-large GPT-3 13B model from \cite{brown2020language}.}
    The two models have similar performance on most of the evaluation tasks.
}
\label{tab:gpt3-evals-13b-comparison}
\footnotesize
\addtolength{\tabcolsep}{-2pt}
\centering
\begin{tabular}{lllSSSSSSSSSSSS}
\toprule
    &&& \multicolumn{2}{c}{Zero-shot} & \multicolumn{2}{c}{One-shot} & \multicolumn{2}{c}{Few-shot} \\
\cmidrule(lr){4-5} \cmidrule(lr){6-7} \cmidrule(lr){8-9}
    Task & Split & Metric &
    {6.7B+$\mu$P} & {13B\cite{brown2020language}} &
    {6.7B+$\mu$P} & {13B\cite{brown2020language}} &
    {6.7B+$\mu$P} & {13B\cite{brown2020language}} \\
\midrule

    HellaSwag & dev & acc &
    \bfseries 71.9877 & 70.9 &
    \bfseries 71.0516 & 70.0 &
    \bfseries 72.4358 & 71.3 
    \\

    LAMBADA & test & acc &
    \bfseries 73.4912 & 72.5 &
    \bfseries 69.8816 & 69.0 &
    74.71376061439514 & \bfseries 81.3000 
    \\

    StoryCloze & test & acc &
    79.4227659702301 & \bfseries 79.5000 &
    \bfseries 80.5986 & 79.7 &
    \bfseries 84.2330 & 83.0 
    \\

    NaturalQS & test & acc &
    \bfseries 9.8615 & 7.84 &
    \bfseries 14.6814 & 13.7 &
    20.221606648199447 & \bfseries 21.0000 
    \\

    TriviaQA & dev & acc &
    \bfseries 47.0036 & 41.8 &
    50.44413862129363 & \bfseries 51.3000 &
    55.53609408232203 & \bfseries 57.5000 
    \\

    WebQS & test & acc &
    \bfseries 11.3189 & 8.22 &
    \bfseries 20.2264 & 19.0 &
    32.97244094488189 & \bfseries 33.5000 
    \\

    Ro$\to$En 16 & test & BLEU-sb &
    \bfseries 26.9000 & 20.8 &
    36.5 & \bfseries 36.7000 &
    38.20000076293945 & \bfseries 38.4000 
    \\

    En$\to$Ro 16 & test & BLEU-sb &
    \bfseries 18.1000 & 6.43 &
    \bfseries 21.0000 & 20.8 &
    \bfseries 22.0000 & 21.8 
    \\

    Fr$\to$En 14 & test & BLEU-sb &
    \bfseries 29.8000 & 22.4 &
    \bfseries 31.7000 & 31.4 &
    38.0 & \bfseries 38.3000 
    \\

    En$\to$Fr 14 & test & BLEU-sb &
    \bfseries 29.6000 & 15.3 &
    28.799999237060547 & \bfseries 30.1000 &
    33.29999923706055 & \bfseries 35.5000 
    \\

    De$\to$En 16 & test & BLEU-sb &
    \bfseries 31.7000 & 24.4 &
    33.29999923706055 & \bfseries 34.5000 &
    38.900001525878906 & \bfseries 39.1000 
    \\

    En$\to$De 16 & test & BLEU-sb &
    \bfseries 23.1000 & 11.0 &
    \bfseries 24.6000 & 23.3 &
    27.600000381469727 & \bfseries 27.7000 
    \\

    Winograd & test & acc &
    85.34798622131348 & \bfseries 87.9000 &
    84.61538553237915 & \bfseries 86.1000 &
    \bfseries 86.4469 & 82.4 
    \\

    Winogrande & dev & acc &
    66.77190065383911 & \bfseries 67.9000 &
    \bfseries 67.5612 & 66.9 &
    \bfseries 71.0339 & 70.0 
    \\

    PIQA & dev & acc &
    \bfseries 79.0533 & 78.5 &
    77.31229662895203 & \bfseries 77.8000 &
    79.21653985977173 & \bfseries 79.9000 
    \\

    ARC (Challenge) & test & acc &
    42.13286638259888 & \bfseries 43.7000 &
    \bfseries 43.9685 & 43.1 &
    43.793705105781555 & \bfseries 44.8000 
    \\

    ARC (Easy) & test & acc &
    \bfseries 64.2857 & 63.8 &
    65.34391641616821 & \bfseries 66.8000 &
    67.3280417919159 & \bfseries 69.1000 
    \\

    OpenBookQA & test & acc &
    54.40000295639038 & \bfseries 55.6000 &
    \bfseries 56.4000 & 55.8 &
    58.399999141693115 & \bfseries 60.8000 
    \\

    Quac & dev & f1 &
    \bfseries 41.7609 & 38.4 &
    \bfseries 43.1352 & 40.6 &
    \bfseries 44.0041 & 40.9 
    \\

    RACE-h & test & acc &
    \bfseries 45.0257 & 44.6 &
    \bfseries 44.8828 & 44.6 &
    \bfseries 45.1973 & 45.1 
    \\

    RACE-m & test & acc &
    \bfseries 58.3565 & 56.7 &
    \bfseries 57.9387 & 56.9 &
    \bfseries 58.5655 & 58.1 
    \\

    SQuADv2 & dev & f1 &
    \bfseries 59.9188 & 56.3 &
    \bfseries 64.9009 & 61.8 &
    \bfseries 68.8726 & 67.7 
    \\

    CoQA & dev & f1 &
    \bfseries 78.5000 & 76.3 &
    \bfseries 80.9000 & 77.9 &
    \bfseries 81.3000 & 79.9 
    \\

    DROP & dev & f1 &
    17.115981543624162 & \bfseries 24.0000 &
    23.292575503355703 & \bfseries 29.2000 &
    \bfseries 33.9242 & 32.3 
    \\

    BoolQ & dev & acc &
    \bfseries 69.3578 & 66.2 &
    \bfseries 74.0673 & 69.0 &
    \bfseries 73.9144 & 70.2 
    \\

    CB & dev & acc &
    \bfseries 21.4286 & 19.6 &
    \bfseries 60.7143 & 55.4 &
    62.5 & \bfseries 66.1000 
    \\

    Copa & dev & acc &
    81.99999928474426 & \bfseries 84.0000 &
    81.00000023841858 & \bfseries 86.0000 &
    \bfseries 88.0000 & 86.0 
    \\

    RTE & dev & acc &
    55.23465703971119 & \bfseries 62.8000 &
    \bfseries 61.0108 & 56.3 &
    52.707581227436826 & \bfseries 60.6000 
    \\

    WiC & dev & acc &
    \bfseries 0.0000 & \bfseries 0.0000 &
    \bfseries 50.0000 & \bfseries 50.0000 &
    50.470219435736674 & \bfseries 51.1000 
    \\

    ANLI R1 & test & acc &
    \bfseries 33.7000 & 33.2 &
    32.40000009536743 & \bfseries 32.7000 &
    30.899998545646667 & \bfseries 33.3000 
    \\

    ANLI R2 & test & acc &
    \bfseries 33.8000 & 33.5 &
    \bfseries 34.8000 & 33.9 &
    \bfseries 35.0000 & 32.6 
    \\

    ANLI R3 & test & acc &
    32.66666531562805 & \bfseries 34.4000 &
    \bfseries 34.7500 & 32.5 &
    \bfseries 36.9167 & 34.5 
    \\

\bottomrule
\end{tabular}
\end{table}

\section{Additional Experiments}

\subsection{Experiments on ResNets}
\label{sec:resnetexperiments}

\subsubsection{ResNet on CIFAR-10}
\label{sec:resnetcifar10experiments}

\begin{figure}
    \centering
    \includegraphics[width=0.85\textwidth]{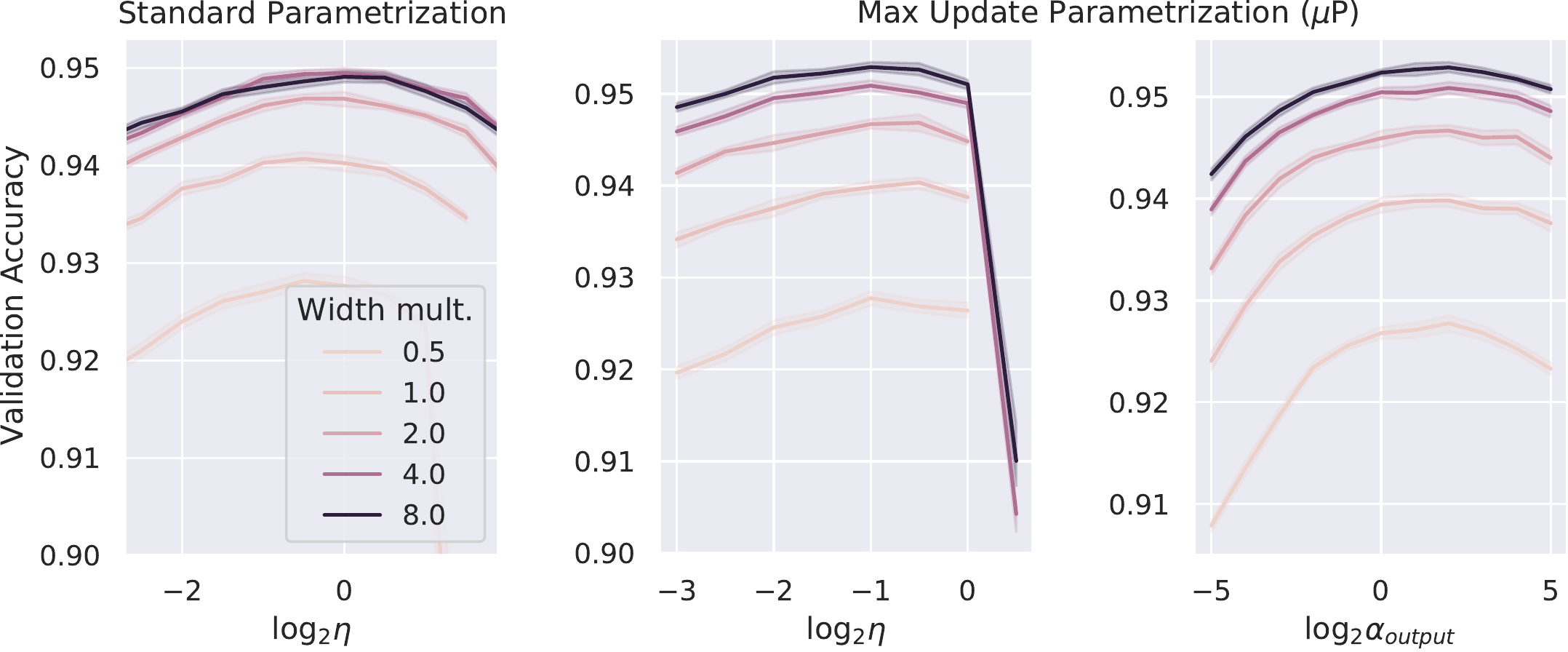}
    \caption[Verifying $\mu$P hyperparameter stability on ResNet]{ResNet on CIFAR-10 for different widths (compared to a base network). On the \textbf{left}, the widest network SP underperforms; on the \textbf{right}, the $\mu$P network has a more consistent HP landscape and performs better. Both networks are tuned at the smallest width for the HP ($\eta$ or $\alpha_{output}$) not in the x-axis.}
    \label{fig: fast_cifar_SP_MUP}
\end{figure}

\paragraph{Setup}
For this case we use Davidnet~\cite{davidnet}, a ResNet variant that trains quickly on CIFAR-10, so as to efficiently investigate its HP landscape.
We train with SGD on CIFAR-10 for 10 epochs; all results are averaged over 15 random seeds.
We use a width multiplier to identify models of different width, and a multiplier of 1 corresponds to the original model in~\cite{davidnet}.
We look at validation accuracy here as the model barely overfits, and our observations will hold for the training accuracy as well.
We first conduct a learning rate sweep for models of different widths using SP; the result is shown in \cref{fig: fast_cifar_SP_MUP}, on the left.

\paragraph{Hyperparameter Stability}
Note that the best model with a width multiplier of 8 under-performs that with a multiplier of 4.
We run the same sweep with $\mu$P, along with a sweep of the output multiplier ($\alpha_{output}$); the result is shown in \cref{fig: fast_cifar_SP_MUP}, on the right.
We notice that wider models always perform better under $\mu$P and that the optimal learning rate $\eta$ and $\alpha_{output}$ are stable across width.

\paragraph{Hyperparameter Transfer}
Next, we perform a grid search for learning rate ($\eta$) and $\alpha_{output}$ on the 0.5x model for both SP and $\mu$P.%
\footnote{Here we tune the 0.5x model instead of the 1x model to simulate the situation that one does ``exploratory work'' on the 1x model but, when scaling up, would like to tune faster by using a smaller proxy model.
\label{foot:tuneon0.5}}
Then, we take the best combination and test on the 8x model, simulating how a practitioner might use $\mu$Transfer.
The result is shown in \cref{tab: fast_cifar_transfer}, where $\mu$P outperforms SP by $0.43\%\pm.001\%$.

\begin{table}[tph]
    \centering
    \small
    \caption[$\mu$Transfer results for ResNet on CIFAR10]{ResNet on CIFAR10: Transferring the best learning rate ($\eta$) and $\alpha_{output}$ from widening factor $0.5$ to $8$; $\mu$P significantly outperforms SP given the same search grid. The best HPs are different as the models are parametrized to be identical at 1x width.\cref{foot:tuneon0.5}}
    \label{tab: fast_cifar_transfer}
    \begin{tabular}{ccccc} \toprule
         Transfer Setup      & Best $\eta$   & Best $\alpha_{output}$ & Valid. Acc. (0.5x) & Valid. Acc. (8x)  \\ \midrule
         SP         & 0.707    & 4                 & 92.82\%    & 94.86\%            \\
         $\mu$P        & 0.5      & 4                 & 92.78\%    & \textbf{95.29\%}   \\ \bottomrule
    \end{tabular}
\end{table}

\subsubsection{Wide ResNet on ImageNet}

\paragraph{Setup}
For this case we use Wide-Resnet, or WRN \citep{zagoruyko2017wide}, a ResNet variant with more channels per layer, to further showcase $\mu$Transfer across width, i.e., number of channels.
We train with SGD on ImageNet for 50 epochs following standard data augmentation procedures.
We use a width multiplier to identify models of different width, and a multiplier of 1 corresponds to the original WRN-50-2-bottleneck in \citep{zagoruyko2017wide}.

\paragraph{Hyperparameter Transfer}
We start with a proxy model with a width multiplier of 0.125 and tune several HPs using the following grid:
\begin{itemize}
    \item $\eta$: $1 \times 2.048 \times 2^z, \text{where } z\in \{-5, -4, -3, ..., 4\}$
    \item $\alpha_{output}$: $10 \times 2^z, \text{where } z\in \{-5, -4, -3, ..., 4\}$
    \item weight decay co-efficient $\gamma$: $3.05 \times 10^{-5} \times 2^z, \text{where } z\in\{-2, -1.5, -1, ..., 1.5\}$
    \item SGD momentum $\beta$: $0.875 \times 2^z, \text{where } z\in\{-2, -1.5, -1, ..., 1.5\}$
\end{itemize}

The grid is centered around the default HPs used by \citep{nvidia_resnet50} for ResNet-50; while not expected to be competitive for WRN, they represent a reasonable starting point for our experiment.

We randomly sample 64 HP combinations from the grid and train for 50 epochs, before selecting the one with the highest top-1 validation accuracy.
Then, we scale up the model following both $\mu$P and SP and run with the same HPs we just selected.
The result is shown in \cref{tab: imagenet_transfer}, where $\mu$P outperforms SP by $0.41\%$ in terms of top-1 validation accuracy.

\begin{table}[tph]
    \centering
    \small
    \caption[$\mu$Transfer results for Wide ResNet on ImageNet]{ResNet on ImageNet: Transferring the best learning rate ($\eta$), $\alpha_{output}$, $\gamma$, and $\beta$ from widening factor $0.125$ to $1$; $\mu$P significantly outperforms SP given the same search grid.}
    \label{tab: imagenet_transfer}
    \begin{tabular}{ccccccc} \toprule
         Transfer Setup   & Best $\eta$   & Best $\alpha_{output}$   & Best $\gamma$   & Best $\beta$   & Valid. Acc. (0.125x) & Valid. Acc. (1x)  \\ \midrule
         SP               & 32.768        & .625                    & .000015        & .4375         & 58.12\%              & 76.75\%            \\
         $\mu$P           & 32.768        & .625                    & .000015        & .4375         & 58.12\%              & \textbf{77.16\%}   \\ \bottomrule
    \end{tabular}
\end{table}

\subsection{Experiments on Transformers}

\begin{figure}
  \centering
  \includegraphics[width=0.85\textwidth]{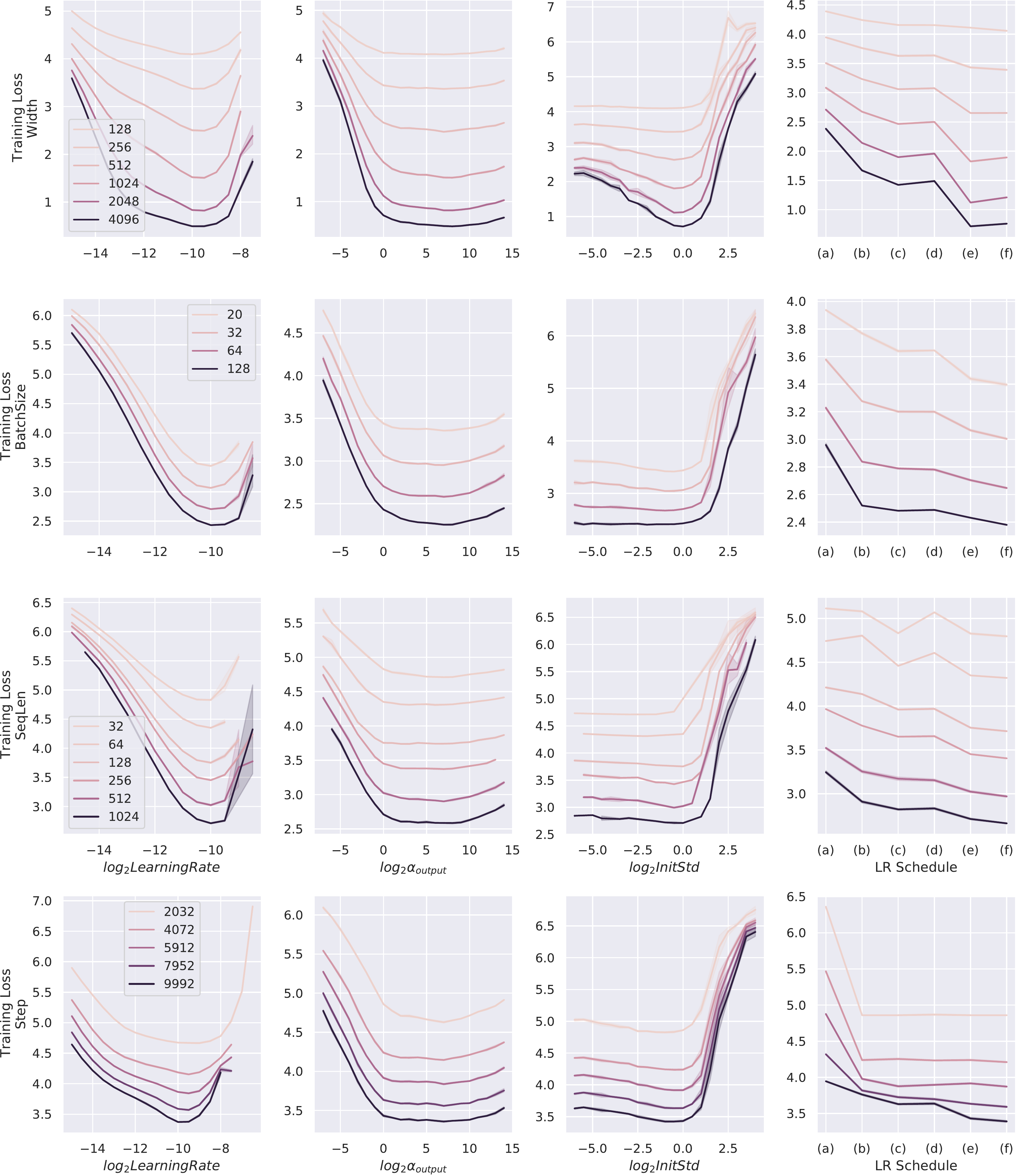}
  \caption[Verifying hyperparameter stability under $\mu$P for Post-LN Transformers]{\textbf{Empirical validation of $\mu$Transfer for Post-LN Transformers.}
  Same setting as \cref{fig:wikitext2_mega}.
  }
  \label{fig:wikitext2_postln}
\end{figure}

\begin{figure}
  \centering
  \includegraphics[width=\textwidth]{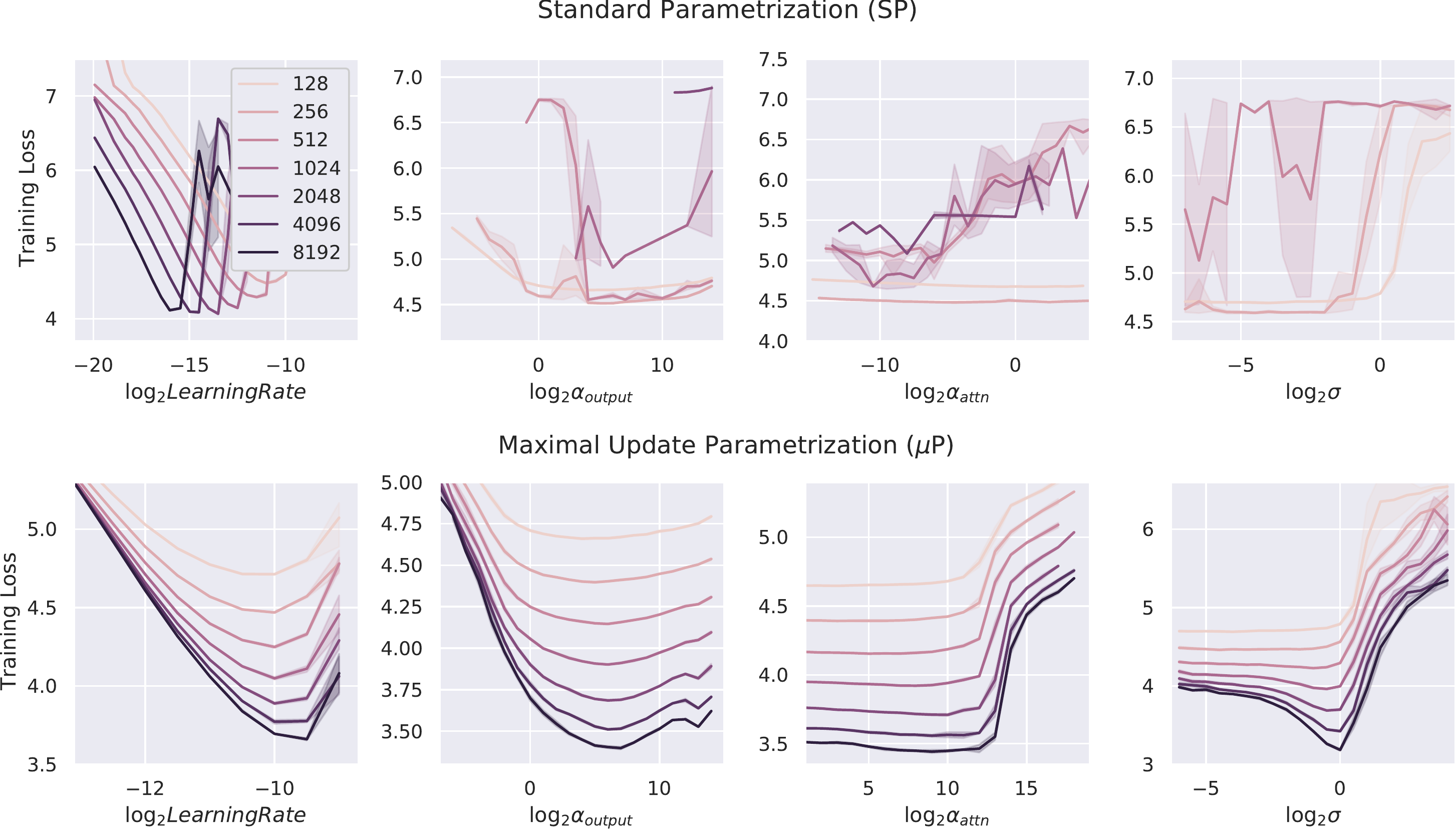}
  \caption[$\mu$Transfer vs naive transfer for post-layernorm Transformers on Wikitext-2]{Post-layernorm Transformer with SP and $\mu$P on Wikitext-2. 
  We sweep one HP across width ($d_{model}$) at a time while keeping the rest fixed; we also scale $d_{head}$ linearly with $d_{model}$ and fixing $n_{head}$. 
  $\alpha_{output}, \alpha_{attn}$ are multipliers for output and key weights, and $\sigma$ is initialization standard deviation.
  This yields unstable result for SP, as expected, where missing points/curves represent divergence; in $\mu$P, the optimal HP choices stabilize as width increases.}
  \label{fig: wikitext2_SP_MUP}
\end{figure}

\begin{figure}
  \centering
  \includegraphics[width=0.85\textwidth]{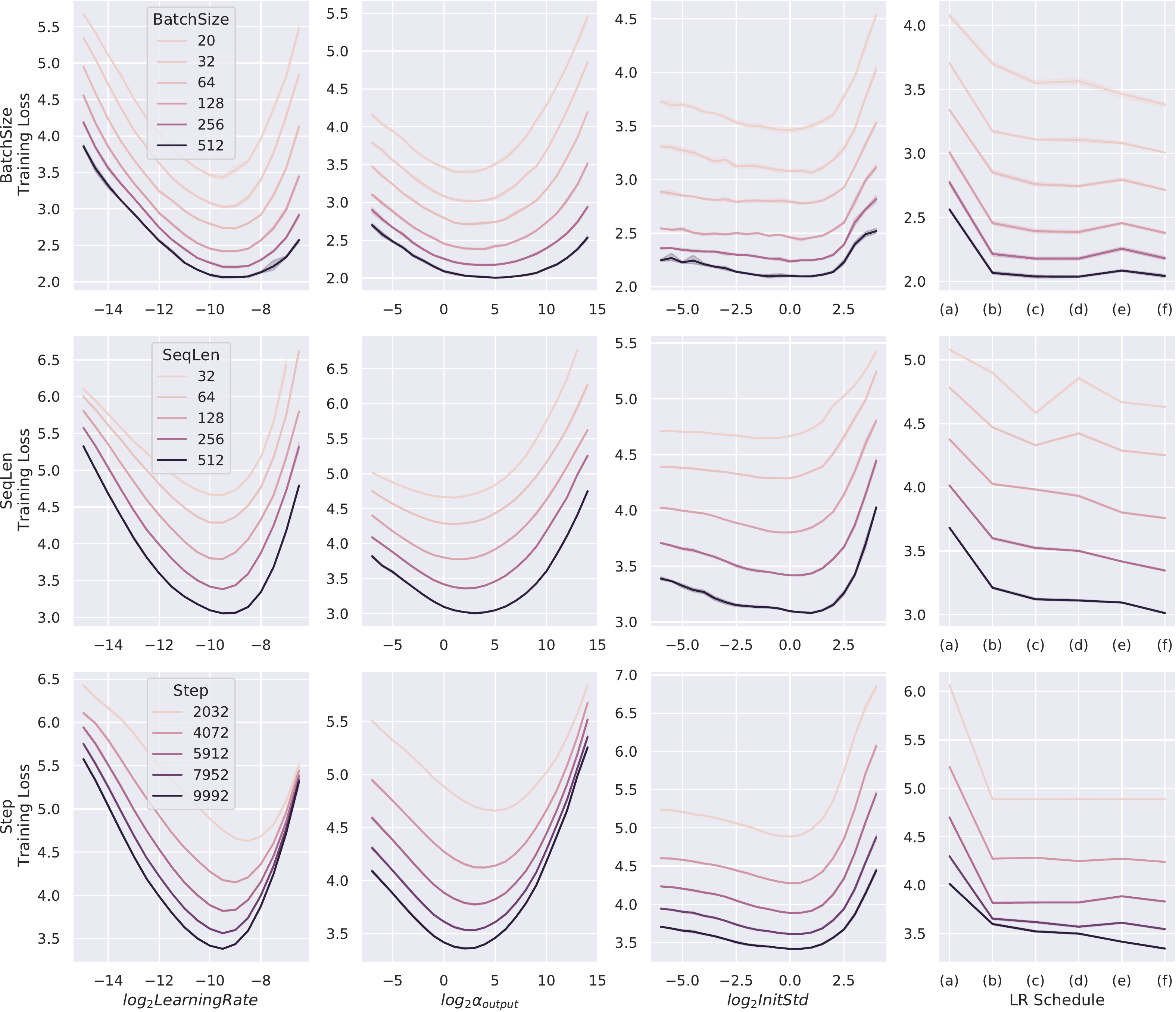}
  \caption[Empirical validation of $\mu$Transfer across Batch Size, Sequence Length, and Training Time on pre-LN Transformers]{\textbf{Empirical validation of $\mu$Transfer across Batch Size, Sequence Length, and Training Time on pre-LN Transformers.}
  Same setting as \cref{fig:wikitext2_mega}.
  Despite some shift, the optimal HPs are roughly stable when transferring from batch size 32, sequence length 128, and 5000 training steps.
  }
  \label{fig:wikitext2_bsz_seqlen_step}
\end{figure}

\subsubsection{Verifying Transfer across Batch Size, Sequence Length, and Training Time on Wikitext-2}
\label{app:bsz_seqlen_time}
See \cref{fig:wikitext2_bsz_seqlen_step}.

\subsubsection{Post-Layernorm Transformers}

\cref{fig:wikitext2_postln} shows the transferability of learning rate, $\alpha_{output}$, initialization standard deviation, and Adam $\beta_2$ across width, batch size, sequence length, and training steps for post-layernorm Transformers.
However, in general, we find transfer across depth to be fragile.

\subsubsection{Hyperparameter Instability of SP Transformers}

\begin{figure}[t]
    \centering
    \includegraphics[width=0.32\textwidth]{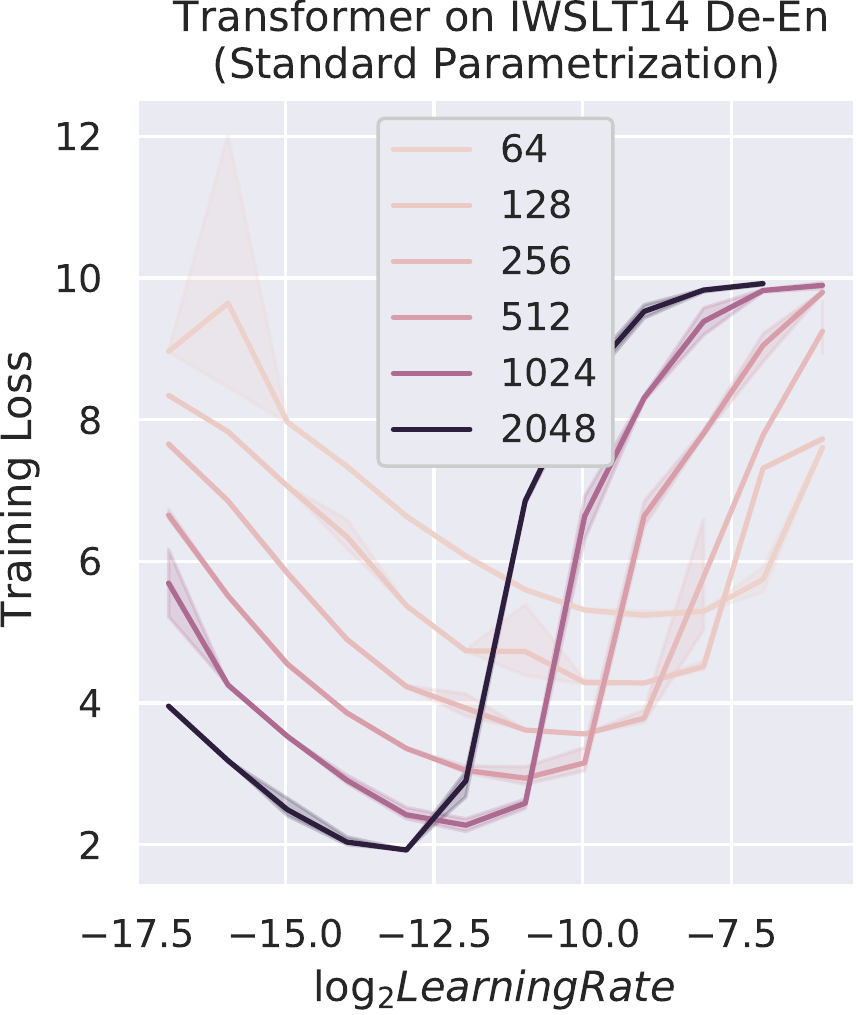}
    \caption[Learning rate landscape is highly unstable under standard parametrization in IWSLT]{Learning rate landscape is highly unstable under standard parametrization in IWSLT.
    \label{fig: iwslt14_SP}
}
\end{figure}

\cref{fig: wikitext2_SP_MUP} and \cref{fig: iwslt14_SP} show the HP instability inherent in SP Transformers.

\section{Implementing \texorpdfstring{$\mu$}{mu}Transfer in a Jiffy}
\label{sec:packageimpl}

As we have shown, one can enable $\mu$Transfer by just reparametrizing the desired model in Maximal Update Parametrization ($\mu$P).
While conceptually simple, switching from Standard Parametrization (SP) to $\mu$P can be error-prone, as popular deep learning frameworks are built around SP.
We strive to build a tool that fulfills two goals:
\newcommand{\basewidth}{\texttt{base\_width}}
\begin{enumerate}
    \item Minimize code changes when switching to $\mu$P;
    \item Keep model behavior invariant, under this switch, at a given \emph{base model shape}.
\end{enumerate}

By \emph{model shape}, we mean the collection of dimensions of all parameters of the model.
The latter goal, which we call \emph{parametrization backward compatibility}, ensures that any code base works exactly as before at the base model shape, similar to \cref{eqn:MUPMLP_basewidth}, e.g.\ the loss at any time step remains exactly the same before and after the switch to $\mu$P.
Of course, when widths start to differ from the base model shape, the model behavior necessarily changes so that HPs can be transferred.

There are two common approaches to setting the base model shape:
1) If one intends to tune a large target model, then the user can set the base model shape to be the shape of the target model (e.g.\ BERT-large or T5-large), so that the target model itself is in standard parametrization.
Then one can tune a proxy model with e.g.\ $width = 124$ to obtain the optimal HPs for the target model.
In addition, if one wishes to scale up further e.g.\ $width = 1024$, then these HPs remain optimal.
2) If one has done exploration on a new idea with a small model and now wishes to scale up, reusing the HP found during this exploration,
then one can set the base model shape to be the shape of the exploratory small model.
Of course, in both scenarios, depth, batch size, and sequence lengths can be scaled up and down as well according to \cref{fig:wikitext2_bsz_seqlen_step} (though note that currently we require users to recreate the base model shape at new depths, since the number of parameters now change with depth).

\paragraph{The \texttt{mup} Package}
We provide our tool as a Python package called \texttt{mup} designed to work with PyTorch.
The following example illustrates the usage of our package.
\begin{center}
  \includegraphics[width=0.7\textwidth]{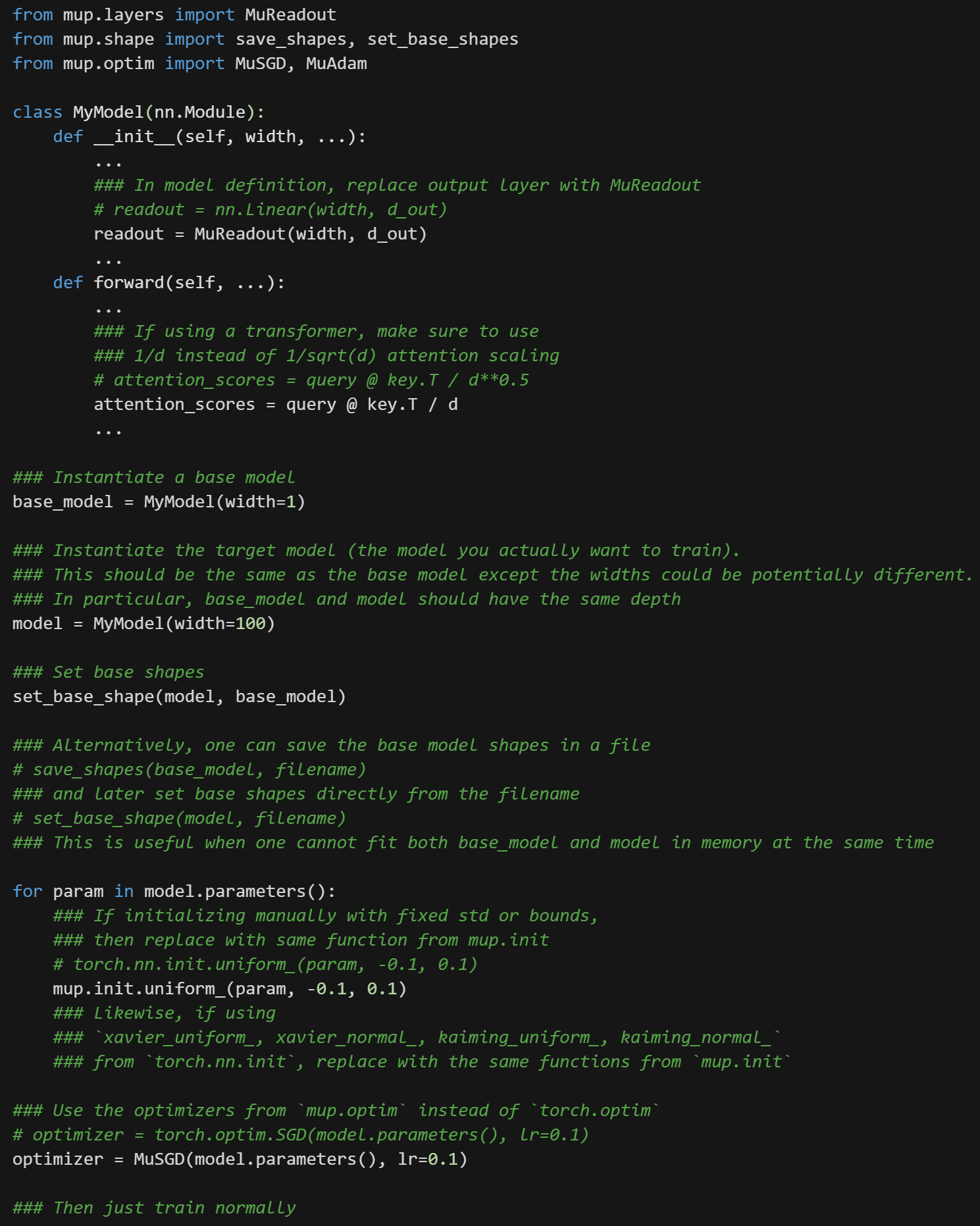}
\end{center}

\paragraph{What Happens in the \texttt{mup} Package}

Under the hood, \texttt{mup} implements the $\mu$P formulation in \cref{tab:MUPalt}.
By invoking \texttt{set\_base\_shape(model, base\_model)}, each parameter tensor \texttt{p} of \texttt{model} gets a \texttt{p.infshape} attribute that stores, for each of its dimensions, the corresponding base dimension and whether that dimension should be considered ``infinite'' (i.e. will be scaled up/down, e.g., $d_{model}$ of a Transformer) or ``finite'' (i.e. will be fixed, e.g., vocabulary size).
This information is used in the initializers and optimizers to automatically scale the parameters or learning rates to be compliant with $\mu$P.
For example, by \cref{tab:MUPalt}, the Adam learning rate of hidden weights \texttt{p} is calculated as $\eta / \texttt{p.infshape.width\_mult()}$, where \texttt{p.infshape.width\_mult()} essentially calculates $\frac{\texttt{fan\_in}}{\texttt{base\_fan\_in}}.$

\section{Reverse-\texorpdfstring{$\mu$}{mu}Transfer for Diagnosing Training Instability in Large Models}
\label{sec:trainingstability}

\begin{figure}[H]
    \centering
    \includegraphics[width=0.85\textwidth]{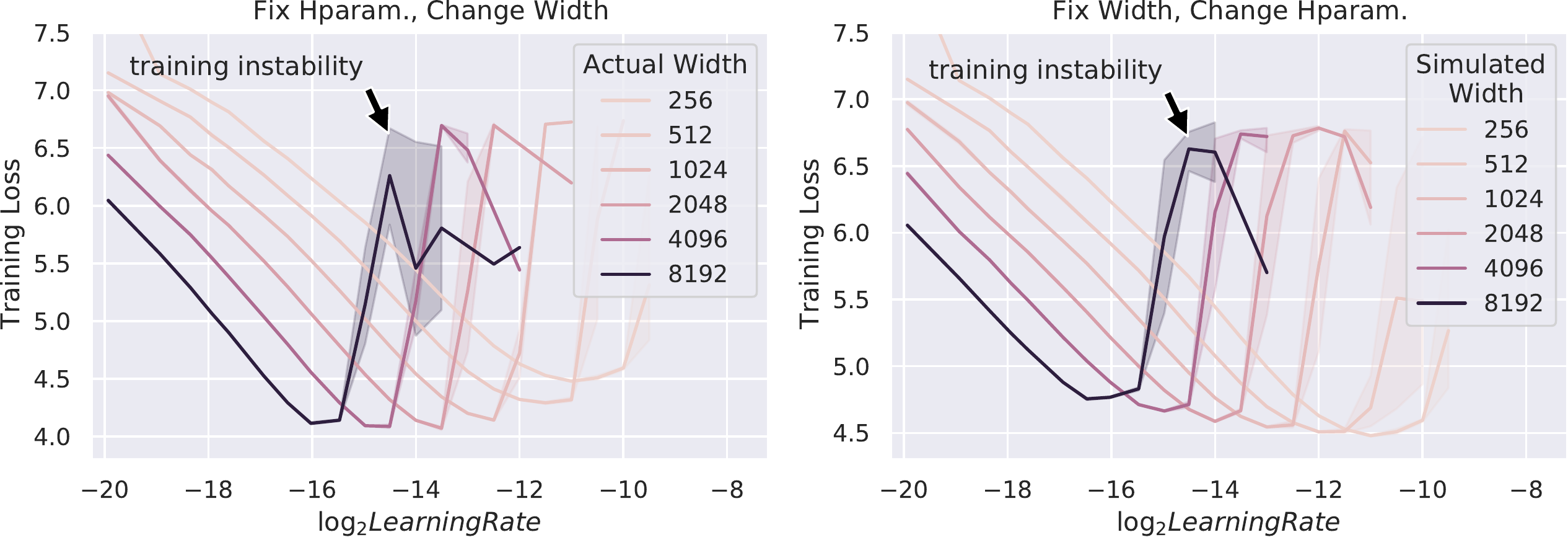}
    \caption[Replicating training instability issue on a small Transformer by \emph{reverse-$\mu$transferring} hyperparameters]{
    \textbf{Replicating training instability on a small Transformer by \emph{reverse-$\mu$Transferring} hyperparameters.}
    These experiments concern 2-layer Transformers in Standard Parametrization (SP) on Wikitext-2, trained with Adam, where width is defined as $d_{model}=d_{ffn}$.
    (Left) LR-vs-loss for wider and wider Transformers.
    (Right) Likewise for \emph{simulated width}:
    Here each point $(\log_2 \eta, loss)$ for simulated width $n$ indicates the loss from training a width-256 $\mu$P Transformer with base width $n$ and LR $\eta$ (i.e.\ loosely speaking, it's using LR transferred from $\eta$ in a width-$n$ SP Transformer).
    \textbf{Takeaway:}
    The overall shapes of the curves are identical between the left and right plots%
    \protect\footnotemark{};
    in particular, a learning rate leads to instability in a wide model iff it does so when transferred back to a narrow model.
    }
    \label{fig: wikitext2_Stability}
\end{figure}

Large Transformers are famously fickle to train \citep{popel_training_2018,liu-etal-2020-understanding}.
We note that a possible source of this instability for larger transformers is the failure of naive hyperparameter transfer via the standard parametrization.
This is certainly consistent with \cref{fig: figure1}, which shows that the optimal learning rate for small Transformers can lead to trivial performance in large Transformers.
We support this hypothesis further by \emph{reverse-$\mu$Transferring the instability-inducing HPs from a large Transformer to a small one and replicating the training instability.}
This is shown in \cref{fig: wikitext2_Stability}.

Practically, this reverse-$\mu$Transfer technique can be used to diagnose or debug training instability problems of large models.
We offer two case studies toward this claim.

1) When training transformers of width 8192 on Wikitext-2, we found certain HP combinations caused divergence in the middle of training.
We reverse-$\mu$Transferred one such HP combination to a model of width 256 and replicated this divergence.
By analyzing this small model's activations right before this divergence, we found that the cause is due to attention logits blowing up.
Note this debugging session proceeded much more quickly than if we directly worked with the large model.
Later we confirmed this is indeed the same cause of the width-8192 model's divergence.

2) A 6B-parameter language model (in standard parametrization) in a separate project experienced repeated blow-up in the middle of training.
We reverse-$\mu$Transferred its hyperparameters to a smaller, 100M-parameter model and replicated the training instability.
This was solved by a retuning of the small model via random search.

\footnotetext{{
    Note that the curves on the left are ``lower'' than curves on the right.
    This just reflects the increasing capacity of wider models able to fit the training data better, so is orthogonal to our point.}}

    \section{An Intuitive Introduction to the Theory of Maximal Update Parametrization}
    \label{sec:intuitiveintro2muP}

    In what follows, we seek to describe useful intuitions and rules of thumb that would be helpful to practitioners and empirical researchers alike in figuring out what is the right neural network parametrization.
    The intuitions we shall describe regarding SGD can be made rigorous as in \citep{TP3,TP4}; those regarding Adam are new, and their formalization will be done in an upcoming paper.

    First, we write down the most basic intuition regarding sums of many random elements, which will underlie all of the calculations that follow.

    \paragraph{Law of Large Numbers (LLN)}
    If $x_1, \ldots, x_n, \ldots$ ``look like'' random independent samples of a random variable $X$, then
    \[
      \frac 1 n \sum_{i=1}^n x_i \to \EV[X],\quad \text{as $n\to\infty$}.\]

    \paragraph{Central Limit Theorem (CLT)}
    In the same scenario as above,
    \[
      \frac 1 {\sqrt{n}} \sum_{i=1}^n (x_i-\EV[X]) \to \Gaus(0, \sigma(X)),\quad \text{as $n\to\infty$},\]
    where $\sigma(X)$ is the standard deviation of the random variable $X$.

    Of course, there are many subtleties one must resolve to make the statements above truly rigorous (e.g., what is the meaning of ``look like''?), but as rules of thumb, they typically give the correct prediction.

    In particular, here we want to note the following basic intuition regarding the size of a sum of $x_i$:
    \[
      \text{when $n$ is large, }
      \sum_{i=1}^n x_i \text{ has typical size }
      \begin{cases}
        \Theta(n) & \text{if $\EV[X] \ne 0$}\\
        \Theta(\sqrt n) & \text{otherwise}
      \end{cases}
    \]
    Here, ``typical size'' can be taken to mean the size $99\%$ of time.
    Again, we stress that this is a good rule of thumb that yields the correct prediction in the cases we are concerned with here; the rigorous versions of this will come from the Tensor Programs framework (e.g., \citep{TP3}).
    
    \subsection{Behaviors of Gaussian Matrices vs Tensor Product Matrices}
    \label{sec:GausVsTensorProducts}
    
    Central to the derivation of $\mu$P for any architecture are key insights on the behaviors of two kinds of random matrices: 1) iid Gaussian random matrix and 2) tensor product matrix (by which we mean a sum of outer products) and more generally what we call \emph{nonlinear} tensor product matrix (see \cref{eq:nonlineartensor}). For example, a neural network, randomly initialized in the typical way, will have each weight matrix look like the former. However, every step of training by gradient descent adds a sum of outer products to this initial matrix, so that the \emph{change in weights} constitute a tensor product matrix. For Adam, the change in weights is not a tensor product but a more general \emph{nonlinear tensor product matrix (see \cref{eq:nonlineartensor}).} In this section, we will particularly focus on the \emph{right scaling} for the entries of such matrices, leading to a discussion of \emph{the right neural network parametrization} in the next section. 
    We concentrate on the key heuristics but eschew burdensome rigor.
    
    \subparagraph*{Key Insights}
    
    Consider a random vector $v\in\R^{n}$ with approximately iid entries and a random matrix $A$ of either size $n\times n$ or $1\times n$, both having entries of size $\Theta(1)$.\footnote{in the sense that the the variance of the entries are $\Theta(1)$} In the context of deep learning, $v$ for example can be an activation vector in an MLP, a Gaussian $A$ the hidden weights at initialization, a (nonlinear) tensor product $A$ the change in hidden weights due to training, and a vector $A$ the readout layer weights. Then $Av$ corresponds to a part of the next layer preactivation or the network output. To make sure the preactivations and the output don't blow up, we thus need to understand the scale of $Av$, especially in the general case where $A$ is correlated with $v$.\footnote{Here ``correlated'' formally means $v$ depends on $W^{\trsp}$ in a Tensor Program. This essentially captures all scenarios of ``$v$ correlated with $W$'' that occurs in deep learning.} This is summarized in \cref{tab:mat}, with the derivations below. Intuitively, a (nonlinear) tensor product or vector $A$ will interact with a correlated $v$ via Law of Large Numbers, hence the $n$-scaling, while a Gaussian $A$ interacts with $v$ via Central Limit Theorem, hence the $\sqrt{n}$-scaling.
    
    \begin{table}
    \centering
    \caption[Expected output size of matrix multiplication between different types of random matrices and a random vector, as preparation for deriving $\mu$P]{Expected entry size of $Av$ for different matrices $A$ and vector $v$ correlated with each other, both having entries of size $\Theta(1)$.}
    \label{tab:mat}
        \begin{tabular}{cccc}
        \toprule
        & Standard Gaussian & (Nonlinear) Tensor Product & Vector\tabularnewline
        & $A\in\R^{n\times n}$ & $A\in\R^{n\times n}$ & $A\in\R^{1\times n}$\tabularnewline
        \midrule
        Entry size of $Av$ & $\Theta(\sqrt{n})$ & $\Theta(n)$ & $\Theta(n)$\tabularnewline
        \bottomrule
        \end{tabular}
    \end{table}
    In the derivations below, we answer a slightly different but equivalent question of ``how to scale $A$ such that $Av$ has entry size $\Theta(1)$?''
    
    \subsubsection{Preparation for the Derivations}
    
    By the results of \citep{TP4}, each (pre-)activation vector and its gradient vector in a multi-layer perceptron, at any time during training, have approximately iid coordinates in the large width limit,\footnote{Our intuition here is derived from the assumption that width is much larger than training time; of course, as illustrated by our myriad experiments, these intuition are very useful even when this is not the case, such as when training to convergence.} and something similar can be said for more advanced networks such as ResNet and Transformers \footnote{E.g. in a convnet, the (pre-)activations are iid across channels, but correlated across pixels}.
    
    \begin{defn}\label{defn:ThetaSize}
      We say any such vector $v \in \R^n$ \emph{has $\Theta(n^a)$-sized coordinates}, or just \emph{$\Theta(n^a)$-coordinates} for short, if $\|v\|^2/n = \Theta(n^{2a})$ as $n \to \infty$.
      Because, by the above discussion, the coordinates are roughly iid when $n$ is large, this intuitively means that each entry of $v$ has ``typical size'' $\Theta(n^a)$.
      We make similar definitions with $\Theta$ replaced by $O$ and $\Omega$.
      
      Furthermore, to each such vector $v$ with $\Theta(1)$-sized coordinates, we can associate a random variable $Z^{v}$, independent of $n$, that represents the coordinate distribution of $v$, in such a way that: If vector $u$ is correlated with $v$, then $Z^{u}$ will also be correlated with $Z^{v}$, and $\lim_{n\to\infty}v^{\trsp}u/n=\EV Z^{u}Z^{v}$.        
    \end{defn}
    \subsubsection{Linear Tensor Product Matrix (e.g. SGD Updates)}
    
    The case of (linear) tensor product matrix can be reduced to the outer product case by linearity. Given $u,v,x\in\R^{n}$ having approximately iid coordinates (of size $\Theta(1)$) like discussed above, we can form the outer product 
    \begin{equation}
    A\defeq u\otimes v/n=uv^{\trsp}/n,\label{eq:linearTensorA}
    \end{equation}
    which is the form of a single (batch size 1) gradient update to a weight matrix. Then, by Law of Large Numbers,
    \[
    Ax=u\frac{v^{\trsp}x}{n}\approx cu,\quad\text{where}\quad c=\EV Z^{v}Z^{x}.
    \]
    
    So $Ax$ also has approximately iid coordinates, distributed like $Z^{Ax}\defeq Z^{u}\EV Z^{v}Z^{x}$. Likewise, if $A$ is a sum of outer products $A=\sum_{i=1}^{k}u^{i}\otimes v^{i}/n$, then 
    \[
    Ax=\sum_{i=1}^{k}u^{i}\frac{v^{i\trsp}x}{n},\quad\text{with coordinates distributed as}\quad Z^{Ax}=\sum_{i=1}^{k}Z^{u^{i}}\EV Z^{v^{i}}Z^{x}.
    \]
    
    Notice that each coordinate of $A$ has size $\Theta(1/n)$. The above reasoning shows that, in order for $Ax$ to have coordinate size $\Theta(1)$ (assuming $x$ does), then $\Theta(1/n)$ is the right coordinate size for $A$, in the general case that $v^{i}$ and $x$ are correlated (as is generically the case during gradient descent, with $A=\Delta W$ for some weights $W$ and $x$ being the previous activations).\footnote{In some corner cases when $x$ is uncorrelated with $v$, then $v^{\trsp}x=\Theta(\sqrt{n})$ by Central Limit, so actually $Ax$ has $\Theta(1/\sqrt{n})$-coordinates. However, this case does not come up much in the context of training neural networks.}
    
    \subsubsection{Nonlinear Tensor Product Matrix (e.g. Adam Updates)}
    
    When using Adam or another adaptive optimizer that normalizes the gradient coordinatewise before applying them, we need to modify our argument slightly to obtain the right coordinate size scaling of the matrix. The gradient update $A$, after such normalization, will take the form of
    \begin{equation}
    A_{\alpha\beta}=\psi(u_{\alpha}^{1},\ldots,u_{\alpha}^{k},v_{\beta}^{1},\ldots,v_{\beta}^{k}),\quad\text{for some \ensuremath{\psi:\R^{2k}\to\R} and vectors \ensuremath{u^{i},v^{j}}\ensuremath{\in\R^{n}}.}\label{eq:nonlineartensor}
    \end{equation}
    
    We say a matrix of this form is a \emph{nonlinear tensor product matrix}.
    
    First, note the tensor product matrices (e.g. the form of SGD update) discussed previously (\cref{eq:linearTensorA}) already takes this form, with $\psi(u_{\alpha}^{1},\ldots,u_{\alpha}^{k},v_{\beta}^{1},\ldots,v_{\beta}^{k})=n^{-1}(u_{\alpha}^{1}v_{\beta}^{1}+\cdots+u_{\alpha}^{k}v_{\beta}^{k})$, so \cref{eq:nonlineartensor} is a strict generalization of linear tensor products. Next, for the example of Adam, each gradient update is $\mu/\sigma$ where $\mu$ (resp. $\sigma^{2}$) is the moving average of previous (unnormalized) gradients (resp. the coordinatewise square of the same).\footnote{Adam also has bias correction for the moving averages which can be accomodated easily, but for simplicity we omit them here.} If these unnormalized gradients are the outer products $u^{1}\otimes v^{1},\ldots,u^{k}\otimes v^{k}$, then the update has coordinates
    \begin{equation}
    (\mu/\sigma)_{\alpha\beta}=\psi(u_{\alpha}^{1},\ldots,u_{\alpha}^{k},v_{\beta}^{1},\ldots,v_{\beta}^{k})\defeq\sum_{i}\gamma_{i}u_{\alpha}^{i}v_{\beta}^{i}/\sqrt{\sum_{i}\omega_{i}(u_{\alpha}^{i}v_{\beta}^{i})^{2}},\label{eq:Adamform}
    \end{equation}
    where $\gamma_{i}$ and $\omega_{i}$ are the weights involved in the moving averages.
    
    Now suppose we have some $A\in\R^{n\times n}$ of the form \cref{eq:nonlineartensor}, where $u^{i},v^{i}\in\R^{n}$ have approximately iid coordinates (of size $\Theta(1)$), and $\psi=n^{-1}\bar{\psi}$ where $\bar{\psi}$ doesn't depend on $n$ (in terms of Adam where $\bar{\psi}$ corresponds to the $\psi$ of \cref{eq:Adamform}, this corresponds to using a learning rate of $1/n$). Then for $x\in\R^{n}$ having approximately iid coordinates of size $\Theta(1)$, by Law of Large Numbers,
    \[
    (Ax)_{\alpha}=\frac{1}{n}\sum_{\beta=1}^{n}\bar{\psi}(u_{\alpha}^{1},\ldots,u_{\alpha}^{k},v_{\beta}^{1},\ldots,v_{\beta}^{k})x_{\beta}\approx\EV\bar{\psi}(u_{\alpha}^{1},\ldots,u_{\alpha}^{k},Z^{v^{1}},\ldots,Z^{v^{k}})Z^{x}\defeq\Psi(u_{\alpha}^{1},\ldots,u_{\alpha}^{k}).
    \]
    Here we made the obvious definition 
    \[\Psi:\R^{k}\to\R,\qquad 
    \Psi(r_{1},\ldots,r_{k})\defeq\EV\bar{\psi}(r_{1},\ldots,r_{k},Z^{v^{1}},\ldots,Z^{v^{k}})Z^{x}.\]
    Thus $Ax$ also has approximately iid coordinates (of size $\Theta(1)$),
    \[
    Z^{Ax}\defeq\Psi(Z^{u^{1}},\ldots,Z^{u^{k}}).
    \]
    For example, in the SGD example with $A=u\otimes v/n$ and $\bar{\psi}(u_{\alpha},v_{\beta})=u_{\alpha}v_{\beta}$, this formula gives $Z^{Ax}=\Psi(Z^{u})$ where $\Psi(z)=z\EV Z^{v}Z^{x}$, recovering the earlier derivation.
    
    In any case, the point here is that $A$ has coordinate size $\Theta(1/n)$, and this is the unique scaling that leads to $Ax$ having coordinate size $\Theta(1)$.
    
    \subsubsection{Vector Case (e.g. Readout Layer)}
    
    The vector $A$ case is similar to the tensor product cases above.
    
    \subsubsection{Gaussian Matrix (e.g.\ Hidden Weights Initialization)}
    
    Now consider the case where $A\in\R^{n\times n}$ is random Gaussian matrix with $A_{\alpha\beta}\sim\Gaus(0,1/n)$ and $x\in\R^{n}$ has approximately iid coordinates distributed like $Z^{x}$. In the context of neural network training, $A$ should be thought of as a randomly initialized weight matrix, and $x$ for example can be taken to be an activation vector in the first forward pass.

    \paragraph{A Quick Intuition}
    By standard random matrix theory, $A$ has $\Theta(1)$ operator norm with high probability.
    Thus, with high probability, for any ``typical'' vector $x$, we expect $\|Ax\|= \Theta(\|x\|)$, even if $x$ is correlated with $A$.
    If $Ax$'s coordinates are ``evenly distributed'', then this would imply $Ax$ has $\Theta(1)$-coordinates if $x$ does.
    However, this is not so clear.
    Below we provide intuitions for why this would be the case.
    
    \paragraph{Intuition for Evenness of Coordinate Distribution}
    If $x$ is independent from $A$ (or sufficiently uncorrelated), then each coordinate $(Ax)_{\alpha}$ has variance $\EV(Z^{x})^{2}=\Theta(1)$ (so by definition has size $\Theta(1)$). Thus, here $A$ having $\Theta(1/\sqrt{n})$-coordinates leads to $Ax$ having $\Theta(1)$-coordinates, in contrast to the tensor product case above.
    
    When $x$ is correlated with $A$, it turns out the same scaling applies ($\Theta(1/\sqrt{n})$ is the unique scaling for $A$'s entries such so that $Ax$ has $\Theta(1)$ entries), but the reasoning is much more subtle: In the context of neural network training, it turns out all scenario where $x$ is correlated with $A$ can be reduced to the case where $x=\phi(A^{\trsp}y,\ldots)$ for some coordinatewise nonlinearity $\phi$ and some other vector $\R^{n}$.\footnote{This is because every ``reasonable'' deep learning computation can be expressed in a Tensor Program.} Let's consider a very simple example with $x=A^{\trsp}\onev$ for the all 1s vector $\onev\in\R^{n}$ (which has coordinate size $\Theta(1)$ as can be checked easily). Then, for each index $\alpha\in[n]$, we can calculate
    \[
    (AA^{\trsp}\onev)_{\alpha}=\sum_{\beta,\gamma}A_{\alpha\beta}A_{\gamma\beta}=\sum_{\beta}A_{\alpha\beta}^{2}+\sum_{\beta}\sum_{\gamma\ne\alpha}A_{\alpha\beta}A_{\gamma\beta}.
    \]
    Since $\EV A_{\alpha\beta}^{2}=1/n$, by the Law of Large Number, the first sum $\sum_{\beta}A_{\alpha\beta}^{2}\approx1$. On the other hand, there are $n$ summands of the form $\sum_{\gamma\ne\alpha}A_{\alpha\beta}A_{\gamma\beta}$, all iid with variance $\frac{n-1}{n^{2}}=\Theta(1/n)$. Thus by the Central Limit Theorem, we expect $\sum_{\beta}\sum_{\gamma\ne\alpha}A_{\alpha\beta}A_{\gamma\beta}\approx\Gaus(0,1)$. Therefore, each coordinate of $(AA^{\trsp}\onev)_{\alpha}$ looks like $1+\Gaus(0,1)=\Gaus(1,1)$ and thus has size $\Theta(1)$; again this is caused by $A$ having $\Theta(1/\sqrt{n})$-coordinates.
    
    This example can be generalized to more general $x$ that is correlated with $A$, but the mathematics is quite involved. See \citep{TP3} for more details.
    
    \subsection{Deriving \texorpdfstring{$\mu$P}{MUP} for Any Architecture}
    \label{sec:DeriveMUP}
    
    Armed with the insight from the last section, we now outline the key steps to derive $\mu$P in \cref{tab:MUP} for any architecture. In practice, $\mu$P implies the following desiderata
    \begin{des}
    At any time during training
    \begin{enumerate}
    \item Every (pre)activation vector in a network should have $\Theta(1)$-sized coordinates\label{enu:act1}\footnote{In a convnet, a (pre-)activation vector corresponds to a single pixel across all channels; in general , we expect (pre-)activations are iid across channels, but correlated across pixels}
    \item Neural network output should be $O(1)$.\label{enu:out1}
    \item All parameters should be updated as much as possible (in terms of scaling in width) without leading to divergence.\label{enu:MUP}
    \end{enumerate}
    \end{des}
    
    Let's briefly justify these desiderata. For the desideratum \ref{enu:act1}, if the coordinates are $\omega(1)$ or $o(1)$, then for sufficiently wide networks their values will go out of floating point range. This problem is particularly acute for low-precision formats that are essential for training large models such as BERT or GPT. Moreover, a general nonlinearity is only well-behaved if its input is in a fixed range (although this is not a problem for homogeneous nonlinearities like relu). For example, for tanh nonlinearity, if the preactivation is vanishing $o(1)$, then $\tanh$ is essentially linear; if the preactivation is exploding $\omega(1)$, then the tanh gradient vanishes.
    
    For the desideratum \ref{enu:out1}, a similar justification applies to the numerical fidelity of the loss function and loss derivative.
    Note that, with desideratum \ref{enu:MUP}, this means the network output should be $\Theta(1)$ after training (but it can go to zero at initialization).
    
    Finally, desideratum \ref{enu:MUP} means that 1) we are doing ``maximal feature learning'' \citep{TP4} and 2) every parameter contribute meaningfully in the infinite-width limit. This ensures that learning rate ``plays the same role'' in the finite-width case as in the infinite-width limit. For example, it prevents the scenario where a weight matrix gets stuck at initialization in the limit for any learning rate (so learning rate does not matter) but evolves nontrivially in any finite-width network (so learning rate does matter).
    
    These desiderata will essentially uniquely single out $\mu$P.
    More formally, $\mu$P is the unique parametrization that admits feature learning in all parameters of the neural network \citep{TP4}, and this property theoretically guarantees HP transfer across width (for sufficiently large width).
    However, for the sake of reaching a broader audience, we will focus more on the intuitive derivations from the desiderata rather than on this formal aspect.
    
    Below, we first assume for simplicity that the width of every layer is $n$, and we focus only on dense weights. Later, we will discuss convolutions and varying the widths between layers.
    
    \subsubsection{\texorpdfstring{$\mu$P}{MUP} Derivation From the Desiderata}
    \label{sec:MUPDerivation}

    Below, we will derive the $\mu$P formulation in \cref{tab:MUP}.
    \cref{{tab:MUPalt},{tab:MUPorig}} can be derived from \cref{tab:MUP} via the following equivalences, which can be easily derived via some simple calculations.

    \begin{lem}\label{lem:scalingEquivalences}
      Let $f_t(\xi)$ denote the neural network function after $t$ steps of training (using any fixed sequence of batches), evaluated on input $\xi$.
      Consider a parameter tensor $W$ with learning rate $C$, initialized as $W \sim \Gaus(0, B^2)$, and with a multiplier $A$. Then for any $\theta > 0$, $f_t(\xi)$ stays fixed for all $t$ and $\xi$ if we set 
      \begin{itemize}
        \item when the optimizer is SGD
        \[
          A \gets A \theta,\ B \gets B / \theta,\ C \gets C / \theta^2
        \]
        \item when the optimizer is Adam, 
        \[
          A \gets A \theta,\ B \gets B / \theta,\ C \gets C / \theta;
        \]
      \end{itemize}
    \end{lem}

    For example, for output weights, \cref{tab:MUP} has $A = 1$, $B = 1/\fanin$, $C = \eta/\fanin$ for SGD and Adam.
    Then taking $\theta = 1/\fanin$, we get the entries in \cref{tab:MUPalt}, with $A=1/\fanin$, $B=1$, $C = \eta \cdot \fanin$ for SGD and $C = \eta$ for Adam.
    Taking $\theta = 1/\sqrt{\fanin}$ instead, we get the entries in \cref{tab:MUPorig}, with $A = 1/\sqrt{\fanin}$, $B = 1/\fanin$, $C = \eta$ for SGD and $\eta/\sqrt{\fanin}$ for Adam.
    Similar calculations hold for the input weights scaling in those tables, after taking into consideration that $\fanin$ is considered a constant in terms of width for the input layer.

    We proceed with the derivation of \cref{tab:MUP} below.
    Recall the definitions of \emph{$\Theta(n^a)$-sized coordinates} or \emph{$\Theta(n^a)$-coordinates} from \cref{defn:ThetaSize}.
    
    \subparagraph*{Output Weights}
    
    Suppose $W\in\R^{1\times n}$ is an output weight. By desideratum \ref{enu:act1}, the input $x$ to $W$ has $\Theta(1)$-sized coordinates. Thus $W$ should have $\Theta(1/n)$-coordinates so that $|Wx|=O(1)$. We can initialize $W$ with $\Theta(1/n)$-coordinates and scale its (per-layer) LR so that $\Delta W$ has $\Theta(1/n)$-coordinates as well.
    This means initializing $W_{\alpha\beta} \sim \Gaus(0, \Theta(1/n^2))$ and use $\Theta(1/n)$ learning rate for both SGD and Adam.
    
    \subparagraph*{Hidden Weights}
    
    Consider a square weight matrix $W\in\R^{n\times n}$. Desiderata \ref{enu:act1} guarantees that the input $x$ to $W$ has $\Theta(1)$-sized coordinates. Generally, $x$ will be correlated with $W$. By \cref{tab:mat}, we can immediately derive
    \begin{description}[font=\it]
    \item [{Initialization}] $W$ should be randomly initialized with coordinate size $\Theta(1/\sqrt{n})$
    \item [{LR}] The learning rate should be scaled so that $\Delta W$ has coordinate size $\Theta(1/n)$
    \end{description}
    so that $(W_{0}+\Delta W)x$ is $\Theta(1)$ if $x$ is, inductively satisfying desideratum \ref{enu:act1}. With Adam, this just means the per-layer LR is $\Theta(1/n)$. With SGD and the scaling of output layers above, we can calculate that the gradient of $W$ has $\Theta(1/n)$-coordinates, so the $\Theta(1)$ SGD LR derived above suffices as well.
    
    \subparagraph*{Input Weights}
    
    Suppose $W\in\R^{n\times d}$ is an input weight. To satisfy desideratum \ref{enu:act1} (i.e. for any input $\xi$, $W\xi$ should have $\Theta(1)$-coordinates), we want $W$ to have $\Theta(1)$-coordinates. We can initialize $W$ with $\Theta(1)$-coordinates and scale its (per-layer) LR so that $\Delta W$ has $\Theta(1)$-coordinates as well.
    This implies initialization variance of $\Theta(1)$ (or $\Theta(1/\fanin)$ since $\fanin = \Theta(1)$ here) and Adam learning rate $\Theta(n)$.
    As above, we can calculate that the gradient of $W$ has $\Theta(1/n)$-coordinates, so we want SGD learning rate $\Theta(n)$.
    
    \subparagraph*{Biases}
    
    Biases follow the same reasoning as input weights (just think of it as an input weight with input 1).
    
    \subparagraph*{Attention}
    
    Suppose the key dimension $d_{k}$ is tending to infinity with width with number of heads $n_{head}$ fixed. Then the key-query contraction $q^{\trsp}k\in\R$ scales like $\Theta(d_{k})$ by Law of Large Numbers (instead of Central Limit Theorem because $q$ and $k$ are generally correlated) and desideratum \ref{enu:act1}, hence the $1/d_{k}$ we propose rather than $1/\sqrt{d_{k}}$.
    
    Now suppose instead that $n_{head}$ tends to infinity with width with $d_{k}$ fixed. Let $K,Q\in\R^{N\times d_{k}\times n_{head}},V\in\R^{N\times d_{v}\times n_{head}}$ be keys, queries, and values across all heads and tokens. Thinking of $N\times d_{k}$ as constants, we may view attention as a nonlinearity coordinatewise in the $n_{head}$ dimension. Then it's clear that our parametrization described above already works.
    
    Finally, we may freely let $d_{k}$ and $n_{head}$ both tend to infinity, and the above reasoning shows that our parametrization still works.
    
    \subparagraph*{Changing Width Ratios}
    
    As noted above, at any time in training, every (pre-)activation vector will have approximately iid coordinates (of order $\Theta(1)$ by desideratum \ref{enu:act1}). Another desideratum for $\mu$P is to ensure that this coordinate distribution (at any particular time) stays roughly invariant as widths increases. When all layer widths are tied, this is automatic if the other desiderata are satisfied, hence why we did not list this above.
    
    When width ratios vary, this is not automatic. In this case, we need to choose whether to replace each $n$ with fan-in or fan-out (or some function of them). Making the wrong choices will let the coordinate distributions vary with width ratios.
    
    Obviously, we should replace $n$ with fan-in for the output layers and with fan-out for the input layers since they are the only dimension scaling with $n$. 
    For the hidden weights, we replace $n$ with fan-in so that the forward pass is preserved. When using Adam (and assuming the initialization of $W$ is quickly dominated by the change in $W$), this ensures that the (pre-)activation coordinate distributions are preserved at any time during training even if we vary widths in different layers differently. (For SGD this doesn't quite work in general because the varying width ratios change the gradient sizes of different layers differently, whereas Adam always normalizes the gradient coordinatewise).
    
    \subparagraph*{Convolution}
    
    A convolution weight tensor $W\in\R^{\fanout\times\fanin\times s_{1}\times s_{2}}$ with kernel size $s_{1}\times s_{2}$ can be thought of just as a $s_{1}s_{2}=\Theta(1)$-sized collection of $\fanout\times\fanin$ dense weights. Then all of our discussions above apply accordingly.
    
    \subsection{Why Other Parametrizations Cannot Admit Hyperparameter Transfer}
    \label{sec:otherparamdontwork}

    \subparagraph*{Standard Parametrization (SP)}
    
    SP doesn't work essentially because it leads to blow-up in the infinite-width limit.
    \begin{enumerate}
    \item For Adam with LR $\Theta(1)$, $\Delta W$ would have $\Theta(1)$-coordinates, causing preactivations to blow up like $\Theta(n)$ by Desideratum \ref{enu:act1} and \cref{tab:mat}.
    We can avoid this blowup with LR $\Theta(1/n)$, but this induces a non-maximal feature learning limit, which, as we argue below, cannot transfer hyperparameters in all situations.
    \item For SGD, the gradient of $\R^{n\times n}$ weight has $\Theta(1/\sqrt{n})$-coordinates, so $\Theta(1)$ learning rate would make preactivation scale like $\Theta(\sqrt{n})$ and hence blow up.
    If we use $\Theta(1/width)$ learning rate, then blow-up does not occur.
    However, this infinite-width limit is in the kernel regime \citep{TP4} and thus does not allow HP transfer for the same reason that NTP below does not.
    
    \end{enumerate}

    \subparagraph*{Neural Tangent Parametrization (NTP)}
    
    We have concrete examples, e.g. Word2Vec in \citep{TP4}, where the NTK limit has trivial performance --- so HPs have no effect at all --- vastly outperformed by finite-width networks --- where HPs matter. More importantly, wider does not always do better in NTP, especially in tasks where feature learning is crucial \citep{TP4,yang2022efficient}.
    So in the context of modern deep learning e.g.\ large language model pretraining, NTP (or SP with $\Theta(1/width)$ LR) does not make sense for wide neural networks.
    
    \subparagraph*{Other Parametrizations}
    
    Recall the \emph{Dynamical Dichotomy Theorem} proven in \cite{TP4}, which says that any nontrivial stable ``natural parametrization'' (formally, ``\emph{abc-parametrization},'' \citep{TP4}) either admits a feature learning limit or a kernel limit, but not both.
    
    Our argument above against SP and NTP will also work against any parametrization inducing a kernel limit.
    Therefore, it remains to ask, can other \emph{feature learning} parametrizations transfer HPs?
    
    We argue no. As shown in \cite{TP4}, any other feature learning parametrization differs from $\mu$P essentially only in that some parameters are not updated maximally.
    By \citep[Sec 6.4]{TP4}, in the infinite-width limit, such parameters can be thought of as being fixed at initialization. Therefore, in such infinite-width limits, the learning rate of such parameters becomes useless.
    As such, we cannot hope for the HP landscape of the limit to reflect the HP landscape of finite-width neural networks.
    
    $\mu$P is the unique feature learning parametrization that updates all parameters maximally, so that the learning rate of each parameter plays approximately the same role in finite-width neural networks as in the infinite-width limit. Consequently, the HP landscape of the $\mu$P limit should reflect the HP landscape of finite-width neural networks.

\end{document}